\documentclass{article} % For LaTeX2e
\usepackage{iclr2025_conference,times}

\usepackage{graphicx}
\usepackage{booktabs} 
\usepackage{multirow} 
\usepackage{subcaption}
\usepackage[utf8]{inputenc}
\usepackage{newunicodechar}
\usepackage[percent]{overpic}
\usepackage{xcolor} 
\newunicodechar{↔}{\ensuremath{\leftrightarrow}}
\usepackage{float}

\usepackage{pifont}
\newcommand{\cmark}{\ding{51}} % check mark

\usepackage{amsmath,amsfonts,bm}

\def\eqref#1{equation~\ref{#1}}
\def\1{\bm{1}}

\def\vc{{\bm{c}}}

\def\ve{{\bm{e}}}

\def\vl{{\bm{l}}}

\def\vn{{\bm{n}}}

\def\vp{{\bm{p}}}

\def\vv{{\bm{v}}}

\def\mD{{\bm{D}}}

\def\mK{{\bm{K}}}

\def\mS{{\bm{S}}}

\DeclareMathAlphabet{\mathsfit}{\encodingdefault}{\sfdefault}{m}{sl}
\SetMathAlphabet{\mathsfit}{bold}{\encodingdefault}{\sfdefault}{bx}{n}

\def\emS{{S}}

\usepackage{hyperref}
\usepackage{url}

\title{Joint Shadow Generation and Relighting via Light-Geometry Interaction Maps}

\author{
Shan Wang $^{1,2}$
\quad \quad   Peixia Li$^{1}$ \quad \quad   Chenchen Xu$^{1}$ \quad  \quad  Ziang Cheng$^{1}$ \quad \quad
Jiayu Yang$^{1}$ \\ 
\makebox[\linewidth][c]{\textbf{Hongdong Li}$^{1,2}$ \quad \quad  \textbf{Pulak Purkait}$^{1}$} \\
\makebox[\linewidth][c]{ $^{1}$Amazon \quad \quad $^{2}$Australian National University }
}
\iclrfinalcopy % Uncomment for camera-ready version, but NOT for submission.
\begin{document}

\maketitle

\begin{abstract}
We propose Light–Geometry Interaction (LGI) maps, a novel representation that encodes light-aware occlusion from monocular depth. Unlike ray tracing, which requires full 3D reconstruction, LGI captures essential light–shadow interactions reliably and accurately, computed from off-the-shelf 2.5D depth map predictions. LGI explicitly ties illumination direction to geometry, providing a physics-inspired prior that constrains generative models. Without such prior, 
these models often produce floating shadows, inconsistent illumination, and implausible shadow geometry.
Building on this representation, we propose a unified pipeline for joint shadow generation and relighting-unlike prior methods that treat them as disjoint tasks-capturing the intrinsic coupling of illumination and shadowing essential for modeling indirect effects.
By embedding LGI into a bridge-matching generative backbone, we reduce ambiguity and enforce physically consistent light–shadow reasoning.
To enable effective training, we curated the first large-scale benchmark dataset for joint shadow and relighting, covering reflections, transparency, and complex interreflections.
Experiments show significant gains in realism and consistency across synthetic and real images.
LGI thus bridges geometry-inspired rendering with generative modeling, enabling efficient, physically consistent shadow generation and relighting. 
%Code and data will be publicly released. 

% [{\bf I would suggest moving Appendix B and C to the main body of the paper, or even used as a Teaser picture. Done}]  

\end{abstract}    
\section{Introduction}
\label{sec:intro}

\begin{figure}[htbp]
\centering
    \begin{subfigure}{0.20\textwidth}
        \includegraphics[width=\linewidth]{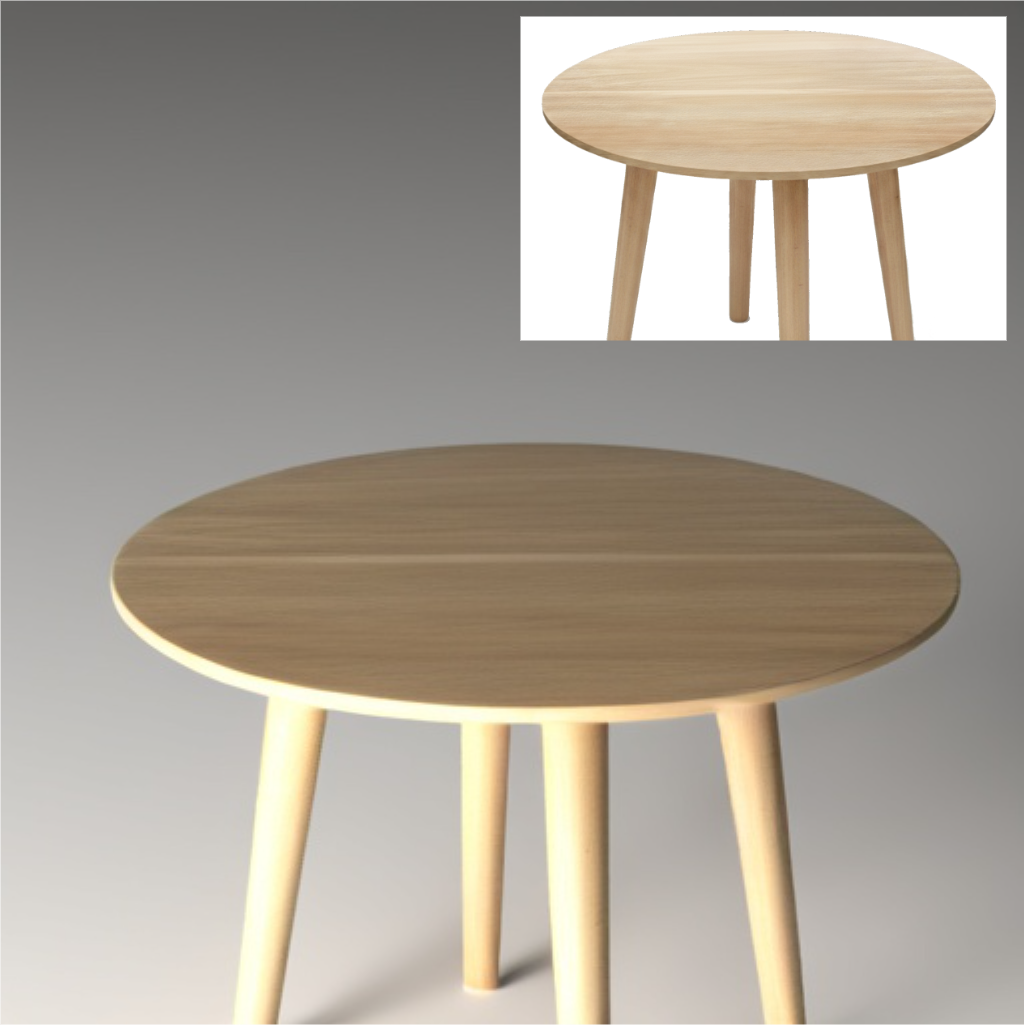}
        \caption{Table insertion}
    \end{subfigure}
    \begin{subfigure}{0.20\textwidth}
        \includegraphics[width=\linewidth]{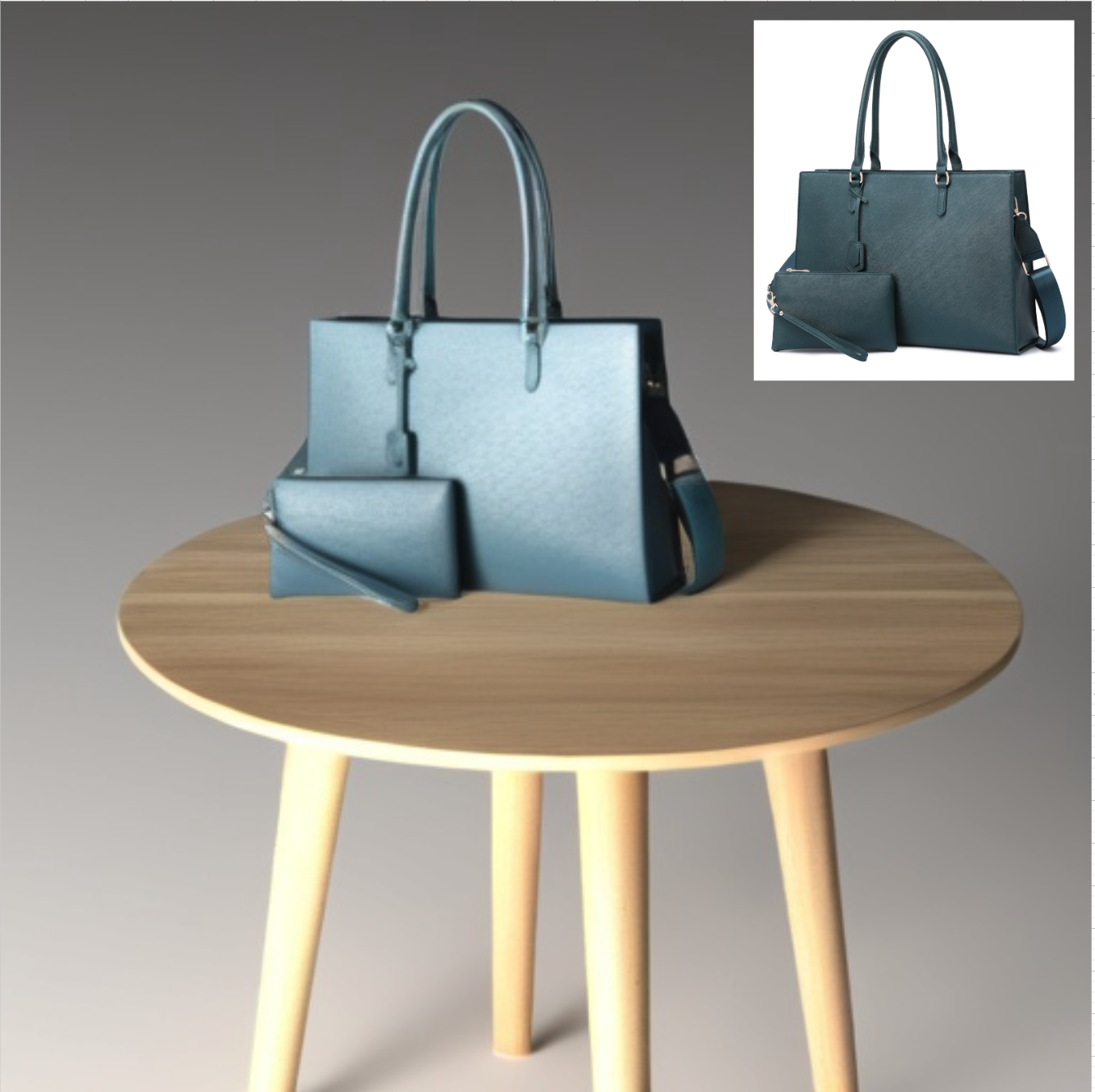}
        \caption{Bag insertion}
    \end{subfigure}
    \begin{subfigure}{0.20\textwidth}
        \includegraphics[width=\linewidth]{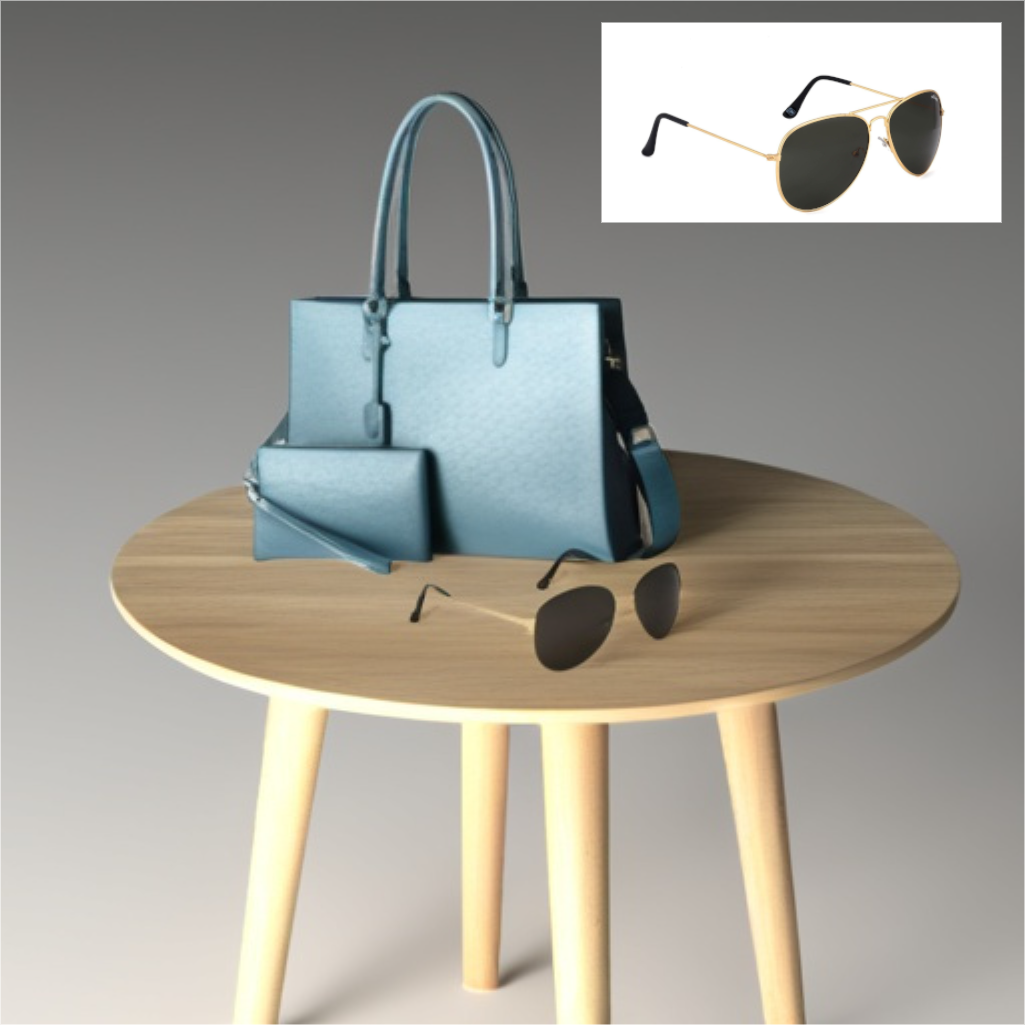}
        \caption{Glass insertion}
    \end{subfigure}
    \begin{subfigure}{0.20\textwidth}
        \includegraphics[width=\linewidth]{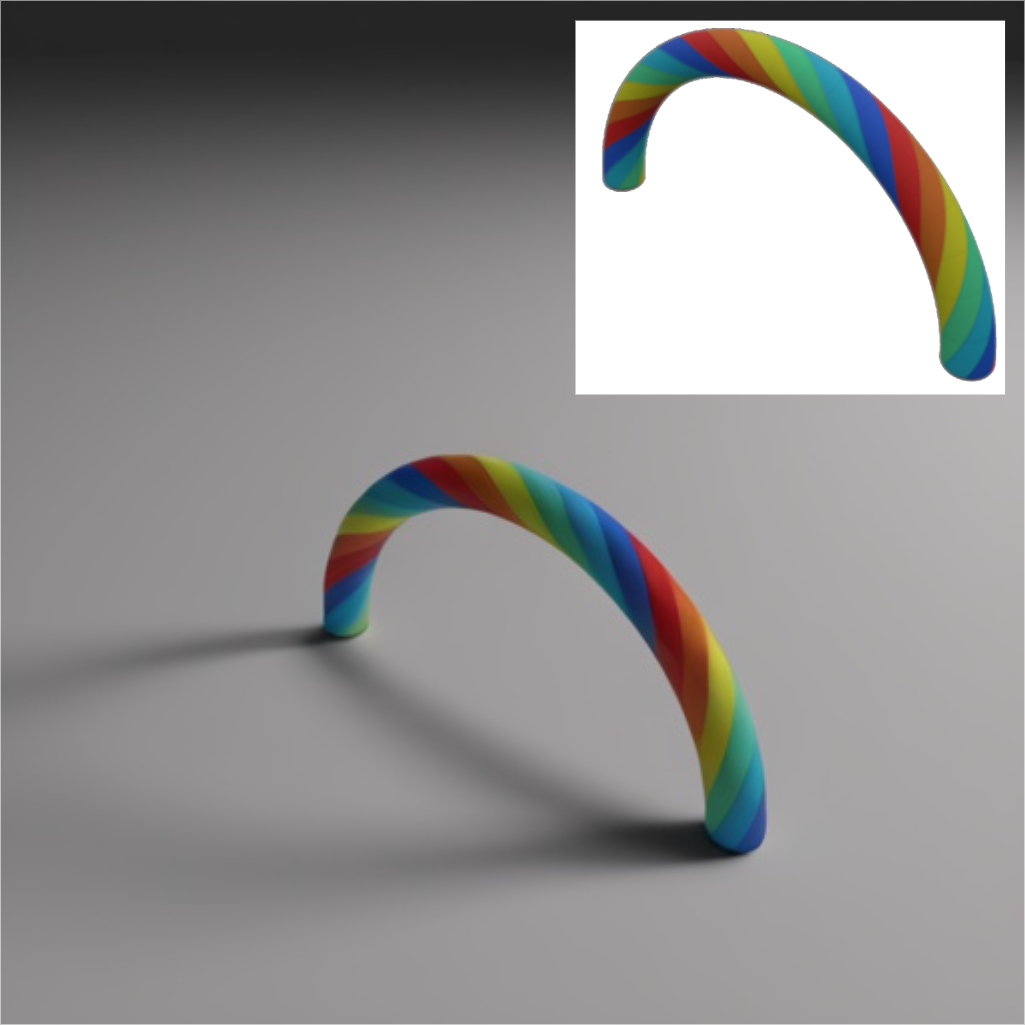}
        \caption{Two lights}
    \end{subfigure}
    \begin{subfigure}{0.1505\textwidth}
        \includegraphics[width=\linewidth]{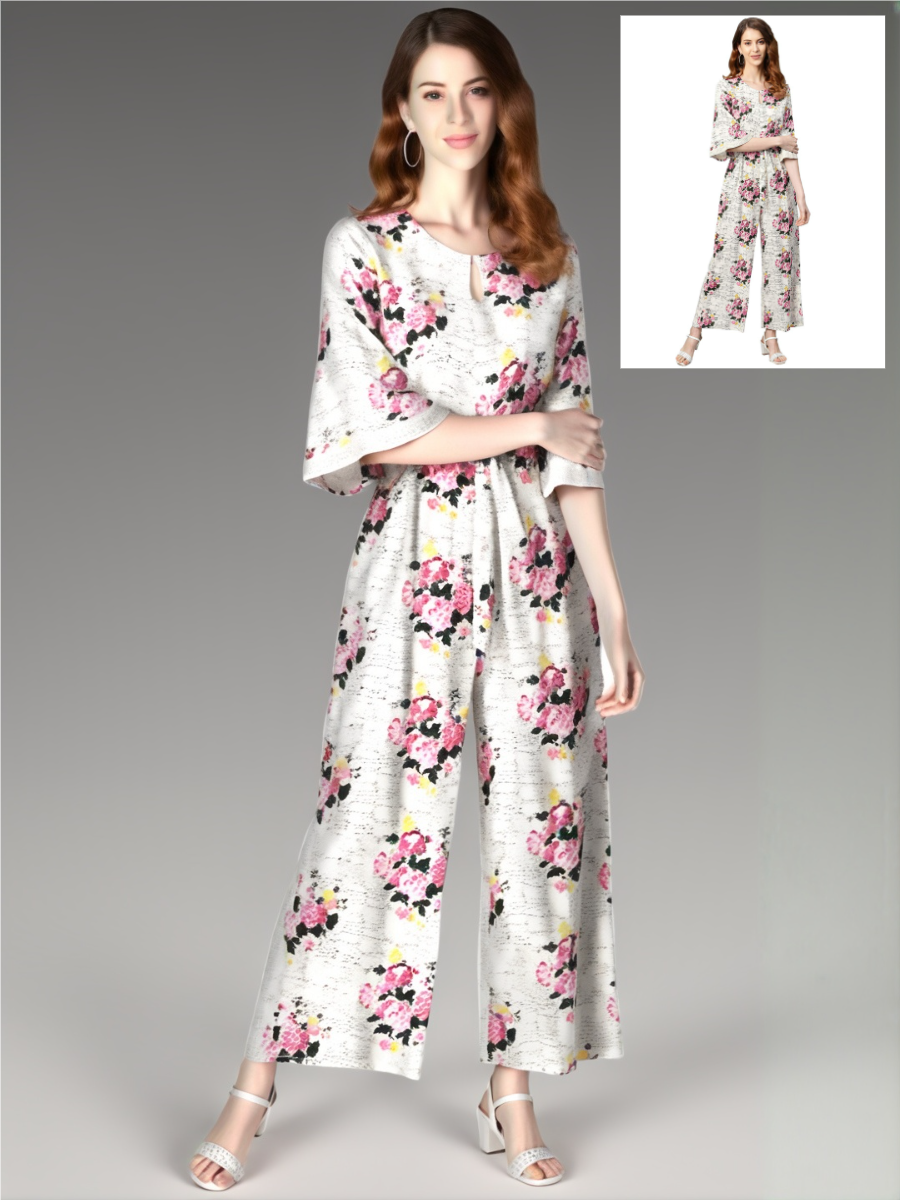}
        \caption{Two lights}
    \end{subfigure}
\caption{Effectiveness of our joint shadow generation and relighting pipeline. Our method produces realistic, texture-aware shadows consistent with object and scene geometry, while preserving faithful relighting across diverse materials such as wood, leather, metal, and glass. (a-c) Multiple object interactions. (d-e) Generalization to multiple light sources.
}
\label{fig:teeser}
\end{figure}

Shadow generation and relighting are important tasks in a wide range of visual computing applications, including virtual product placement, post-capture image editing, augmented reality, and digital content creation. 
Realistic shadow generation and relighting require reasoning about how light interacts with scene geometry. Traditional physically based rendering methods achieve this through explicit 3D reconstruction and ray tracing ~\cite{pharr2023physically,keller1997instant}, but they are computationally expensive and impractical in single-view settings. On the other hand, recent generative approaches, such as bridge matching \cite{tasar2024controllable, chadebec2025lbm} or diffusion \cite{liu2020arshadowgan,hong2022shadow,Liu_2024_CVPR,winter2024objectdrop}, can synthesize shadows and illumination from RGB inputs, but in the absence of physical constraints they often produce floating shadows, inconsistent illumination, or implausible geometry, particularly under complex lighting.

We introduce \emph{Light–Geometry Interaction} (LGI) maps, a novel 2.5D representation that directly encodes occlusion relationships between light and geometry from monocular depth. LGI maps differ from traditional ray tracing, which requires full 3D geometry, by providing a compact, differentiable approximation of light transport suitable for end-to-end learning. They also improve upon prior depth-based conditioning by explicitly coupling depth with light direction, serving as a physics-inspired prior that constrains generative models while remaining computationally efficient.

Building on LGI maps, we design a unified pipeline for joint shadow generation and object-level relighting in scene-aware settings, where a newly inserted object must cast shadows, produce reflections, and be illuminated consistently with the background light.
% Building on LGI map, we design a unified pipeline for joint shadow generation and image relighting. 
Prior works typically treat these tasks independently: shadow models \cite{Liu_2024_CVPR,zhao2025shadow,tasar2024controllable,chadebec2025lbm} generate planar or template-based shadows, while relighting methods \cite{zhang2025scaling,kim2024switchlight} focus solely on object reflectance. However, light and shadows are intrinsically coupled — accurate modeling requires reasoning about direct illumination, secondary reflections, and inter-reflections simultaneously. Our joint formulation enforces this coupling, yielding coherent shadow–light interactions that cannot be achieved when the two tasks are separated.

To support this task, we construct a large-scale synthetic dataset, \emph{ShadRel}, for joint modeling of shadows and relighting. Unlike existing datasets that emphasize either hard shadows ~\cite{Liu_2024_CVPR,tasar2024controllable} or object-only relighting ~\cite{DBLP:journals/corr/abs-2005-05460,kim2024switchlight}, ShadRel dataset includes soft shadows, reflective and transparent materials, and inter-reflections, providing a comprehensive resource for training and evaluation.
% . This provides the first large-scale resource for training and evaluating models that require consistency across both shadow and relighting dimensions.

Through extensive evaluation, we demonstrate that our framework achieves state-of-the-art (SOTA) performance across a wide range of challenging visual scenes, bridging the gap between geometry-free neural rendering and computationally intensive physically-based approaches. 
Though designed at the single-object and single-light level, it naturally extends to multi-object editing and multiple light sources (Fig.~\ref{fig:teeser}).
Our framework thus offers an efficient yet physically inspired alternative suitable for practical shadow-aware image editing and realistic relighting.

{Our contributions} are summarized as follows: 
\begin{itemize}
\item Light–Geometry Interaction Map: a novel light-aware occlusion representation that bridges the gap between geometry-inspired rendering and unconstrained generative models.

\item Joint shadow–relighting pipeline: a unified framework to couple shadow generation and relighting, enabling physically consistent reasoning about direct lighting, secondary reflections, and inter-reflections.

\item ShadRel dataset for coupled light transport: a large-scale dataset designed to capture challenging illumination effects, supporting rigorous training and evaluation.

\end{itemize}

\section{Related Work}
\label{sec:relatedwork}

% \subsection{Shadow Generation}
\textbf{Shadow Generation.}
Classical rendering techniques such as ray tracing~\cite{wald2001interactive,purcell2005ray} and path tracing~\cite{christensen2018renderman,lafortune1996rendering} accurately simulate light transport but require detailed 3D geometry and are computationally expensive. Neural methods attempt to reconstruct geometry from multi-view inputs~\cite{zhao2024illuminerf,lin2023urbanir}, making them unsuitable for single-view settings.
Alternative approaches bypass full 3D reconstruction. SSN~\cite{sheng2021ssn} estimates ambient occlusion from object masks, and pixel height maps approximate geometry for shadow simulation~\cite{sheng2022controllable,sheng2023pixht}. GAN-based models~\cite{zhang2019shadowgan,liu2020arshadowgan} use adversarial training and spatial attention to synthesize shadows, while diffusion-based~\cite{hong2022shadow,Liu_2024_CVPR,winter2024objectdrop} and bridge-based methods~\cite{tasar2024controllable,chadebec2025lbm} generate shadow regions with adjustable placement.  
Yet most of these methods are fundamentally 2D, relying on templates or bounding boxes. \cite{hong2022shadow} decompose synthesis into shape estimation and filling, while \cite{zhao2025shadow} use rotated boxes and templates. Such heuristics break down in complex scenes with ambiguous occlusion. Depth-conditioned works~\cite{griffiths2022outcast,kocsis2024lightit} leverage predicted depth to approximate geometry but require multi-stage pipelines and shading annotations. In contrast, we treat predicted depth as a 2.5D structural cue, embedding it directly into LGI maps to provide a differentiable prior for light–shadow interactions in an end-to-end setting. 

% \subsection{Image Relighting} 
\textbf{Image Relighting.}
% Traditional r
Relighting methods based on inverse rendering and reflectance models~\cite{wenger2005performance,wang2020single,zeng2024dilightnet,zhu2022learning},
primarily focus on regressing radiance on object surfaces, while largely ignoring cast shadows.
More recent deep learning approaches~\cite{bhattad2024stylitgan,xing2024retinex,kim2024switchlight,liang2025diffusion,zeng2024rgb,zhang2025zerocomp,lin2023urbanir,liang2025diffusion,zhu2022learning} estimate or disentangle image intrinsics (e.g., albedo, normals, roughness, metallic) to constrain relighting, but these typically require strong supervision. 
\cite{zhang2025scaling} instead enforces linear consistency between appearances under different illuminations and their mixture, but requires training scenes with at least two distinct light sources.
Some recent work highlights the role of shadows in achieving consistent relighting. \cite{fortier2024spotlight} shows that coarse shadow modeling improves illumination quality, underscoring the need for light–shadow coherence. Still, most prior methods focus on object illumination alone, overlooking the coupled effects of shadows.
% and secondary light transport. 

% Traditional relighting methods rely on inverse rendering and reflectance models~\cite{wenger2005performance,wang2020single}. More recent deep learning approaches~\cite{zhang2025scaling,bhattad2024stylitgan,xing2024retinex,kim2024switchlight,liang2025diffusion,zeng2024rgb,zhang2025zerocomp,lin2023urbanir} estimate or disentangle image intrinsics (e.g., albedo, normals) to constrain relighting, but these typically require strong supervision. 
% Some recent work highlights the role of shadows in achieving consistent relighting. \cite{fortier2024spotlight} shows that coarse shadow modeling improves illumination quality, underscoring the need for light–shadow coherence. Still, most prior methods focus on object illumination alone, overlooking the coupled effects of shadows.

% \subsection{Toward Joint Approaches}
\textbf{Toward Joint Approaches.}
Although shadow synthesis and relighting are tightly coupled in real scenes, very few methods tackle them together. Most follow a sequential design: for example, \cite{griffiths2022outcast} predicts shadow shape before relighting, and \cite{fortier2024spotlight} uses coarse shadows to guide relighting. These pipelines remain disjoint and rely on handcrafted intermediate steps.
Recent generative frameworks such as bridge matching~\cite{tasar2024controllable,chadebec2025lbm} and diffusion-based harmonization~\cite{winter2024objectdrop} offer flexible image-to-image translation, but without explicit light–geometry modeling they often produce inconsistent illumination and floating shadows.  
In contrast, our method introduces Light–Geometry Interaction maps as a structured prior and integrates them into a single pipeline for joint shadow generation and relighting, capturing secondary reflections and inter-reflections that independent pipelines cannot.

\section{Baseline Model: Latent Bridge Matching}

Recent works~\cite{winter2024objectdrop,tasar2024controllable} demonstrate that diffusion models and bridge matching techniques can effectively generate realistic shadows and enable photorealistic image relighting. We adopt latent bridge matching~\cite{chadebec2025lbm} as our baseline model.

Latent bridge matching learns to transform samples from a source distribution $\pi_0$ to a target distribution $\pi_1$, given paired samples $(x_0, x_1) \sim \pi_0 \times \pi_1$. For efficiency, the method operates in latent space using an encoder-decoder pipeline: $z = \mathcal{E}(x)$ and $x = \mathcal{D}(z)$, following ~\cite{rombach2022high}.
The method defines a Brownian bridge~\cite{revuz2013continuous} between latent codes $z_0$ and $z_1$ sampled from their latent distributions, respectively:
\begin{equation}
z(t) = (1-t)z_0 + tz_1 + \sigma \sqrt{t(1 - t)}\, \epsilon, \quad \epsilon \sim \mathcal{N}(0, \mathit{I}), t \in [0,1], \sigma \geq 0.
\label{eq:trajectory}
\end{equation}
where $t$ denotes the time step and $\sigma$ controls noise level.
%A special case is reached when $\sigma=0$, where Eq.~\ref{eq:trajectory} reduces to standard bridge matching with linear interpolant.
%
A neural network is then trained to regress the corresponding SDE drift $v_\theta$ along this bridge by minimizing a mean squared error loss, defined as the expectation $\mathbb{E}[\cdot]$ over paired samples ($z_0, z_1$) and time steps $t$:
\begin{equation}
    \mathcal{L}_{z} = \mathbb{E} \left[ \left\| v(z(t), t)  - v_\theta(z(t), t) \right\|^2 \right], \quad v(z(t), t) = \frac{(z_1 - z(t))}{1 - t}.
\label{eq:matching}
\end{equation}

The framework naturally extends to conditional generation by introducing conditioning variables $c$, in which case the SDE drift becomes $v_\theta(z(t), t, c)$. During training, paired samples are encoded to latents, a timestep $t$ is sampled, and noisy samples $z(t)$ are created based on Eq.~\ref{eq:trajectory}. The predicted target latent is retrieved as:
\begin{equation}
\hat{z}_1 = (1 - t) \cdot v_\theta(z(t), t, c) + z(t),
\label{eq:target_z}
\end{equation}
which is then decoded to image space as $\hat{x}_1 = \mathcal{D}(\hat{z}_1)$. The final loss combines latent matching (Eq.~\ref{eq:matching}) with a pixel-level loss $\mathcal{L}_{x}(\cdot)$: 
\begin{equation}
\mathcal{L} = \mathcal{L}_{z}(\mathcal{E}(x_0), \mathcal{E}(x_1)) + \lambda \cdot \mathcal{L}_{x}(\hat{x}_1, x_1),
\label{eq:final_loss}
\end{equation}
where the baseline model adopts LPIPS~(\cite{zhang2018lpips}) as its pixel-level loss.
While effective for general image-to-image translation tasks, this baseline framework has limitations for shadow generation and relighting applications. By solely relying on 2D image information, the method lacks access to crucial geometric properties such as surface normals, depth, and spatial relationships. This absence of 3D geometric understanding limits its ability to accurately model the complex interactions between objects, lighting conditions, and shadow formation, ultimately affecting the physical realism and consistency of the generated results.

\section{Method} 
\label{sec:method}

\begin{figure}[h]
\begin{center}
\includegraphics[width=0.95\textwidth]{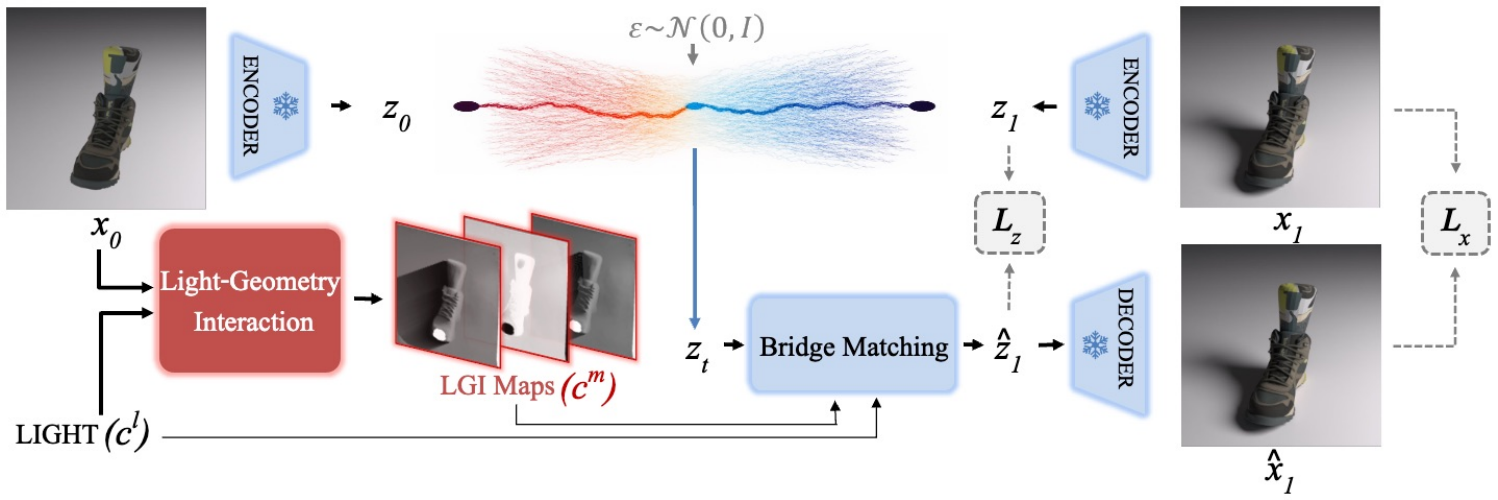}
% \fbox{\rule[-.5cm]{0cm}{4cm} \rule[-.5cm]{4cm}{0cm}}
\end{center}
\vspace{-0.4cm}
\caption{Overview of the proposed method. Our approach uses a bridge-matching strategy to transform shadow-free latent codes ($z_0$) into shadowed counterparts ($z_1$), conditioned on global light cues (e.g., light color, radius) and image-derived light–geometry interaction maps. The key novelty lies in generating these interaction maps from image ($x_0$) and light (see \ref{subsec:occlusion} for details).}
% Overview of the Proposed Method. Our approach leverages a bridge-matching strategy to transform shadow-free latent ($z_0$) to their shadowed counterparts ($z_1$), conditioned on lighting information. In addition to global lighting parameters, we incorporate LGI maps (detailed in \ref{subsec:occlusion}) derived from the estimated depth and light. }
\label{fig:overview}
\vspace{-0.2cm}
\end{figure}

Our framework enables accurate and physically informed shadow generation and relighting by introducing LGI information calculated only from monocular depth images. An overview of our approach is shown in Fig.~\ref{fig:overview}. The system transforms shadow-free images $x_0 \in \pi_0$ into shadowed counterparts $x_1 \in \pi_1$. 
We initialize from a pretrained diffusion model, and keep the encoder $\mathcal{E}(\cdot)$ and decoder $\mathcal{D}(\cdot)$ frozen.
% (SDXL)
Training then focuses on bridge matching conditioned on LGI maps.
% derived from light–geometry interaction.
% by leveraging LGI maps approximated through an LGI module. 

%off-the-shelf monocular depth estimation~\cite{chadebec2025lbm}

%The predicted target latent is retrieved as $\hat{z}_1 = (1 - t) \cdot v_\theta(z(t), t, c^l, c^m) + z(t)$, 
Following Eq.~\ref{eq:trajectory}, we obtain $z(t)$ and pass it to the network to predict target latent $\hat{z}_1$ as in Eq.~\ref{eq:target_z}.
The drift network $v_\theta$ is conditioned on $c = \{c^l, c^m\}$, where $c^l$ denotes the global lighting parameters and $c^m$ denotes the LGI maps. The global parameters $c^l$ contain information about light color, radius, distance, intensity, and direction (azimuth and elevation). The azimuth is defined in the camera plane, while elevation is measured relative to it, consistent with our LGI maps construction. The LGI maps $c^m$ provide image-derived lighting-geometry interactions, as detailed in Section~\ref{subsec:occlusion}.

To focus computation on the most important regions, we replace the pixel-level loss term $\mathcal{L}_{x}(\cdot)$ in Eq.~\ref{eq:final_loss} with a weighted L1 loss in image space that emphasizes areas of brightness changes:
\begin{equation}
\mathcal{L}_x(\hat{x}_1, x_1) = \frac{1}{M}\sum_{m=1}^{M} w^{(m)} \cdot |x_1^{(m)} - \hat{x}_1^{(m)}|, \quad w^{(m)} = (|x_1^{(m)} - x_0^{(m)}| > \tau) \oplus \mathcal{K},
\label{eq:lx}
\end{equation}
where $\hat{x}_{1}$ denotes the predicted shadowed image, $m$ indexes over the $M$ training samples, $\tau$ is a threshold used to identify regions with significant brightness changes, and $\oplus \mathcal{K}$ denotes a dilation operation to expand these regions. 
%The final training objective combines both losses as $\mathcal{L} = \mathcal{L}_z + \lambda\mathcal{L}_x$, where $\lambda$ is a weighting factor balancing the contributions of latent and image-space losses.

\subsection{Light-Geometry Interaction Maps Generation}
\label{subsec:occlusion}

In this section, we describe how LGI maps $\vc^m$ are derived from input images $x_0$ and lighting $\vl$. 
% We model light sources as point lights and compute the LGI maps through a five-stage process involving depth estimation, 3D lifting, ray sampling, elevation difference calculation and LGI maps generation.
Light sources are modeled as point lights, and LGI maps are generated through five steps: depth estimation, 3D lifting, ray sampling, elevation difference calculation, and final map construction.

\textbf{Depth Estimation.} We first estimate depth $\mD$ using an off-the-shelf monocular method~\cite{chadebec2025lbm}. Since monocular depth estimation typically produces only normalized depth up to an unknown scale, we rescale the predictions to match the light coordinates. This does not require metric scale—only consistency between light and scene geometry. In ShadRel dataset, light is given in camera coordinates with meter units, while we also provide an image-harmonized variant (see supplementary material) with normalized depth $(0,1]$ and spatial extent $[-0.5,0.5]$.
% ~\footnote{Our method supports any coordinate system as long as the relative position of the light with respect to the scene is known. In the proposed dataset, light coordinates are specified in camera coordinates with meter units. We also provide an image-harmonized method, detailed in the supplementary material, where the light position is estimated in normalized coordinates with scene depth values in $(0,1]$, and width and height in $[-0.5,0.5]$.}

\begin{figure}[htbp]
\centering
    \begin{subfigure}{0.24\textwidth}
        \includegraphics[width=\linewidth]{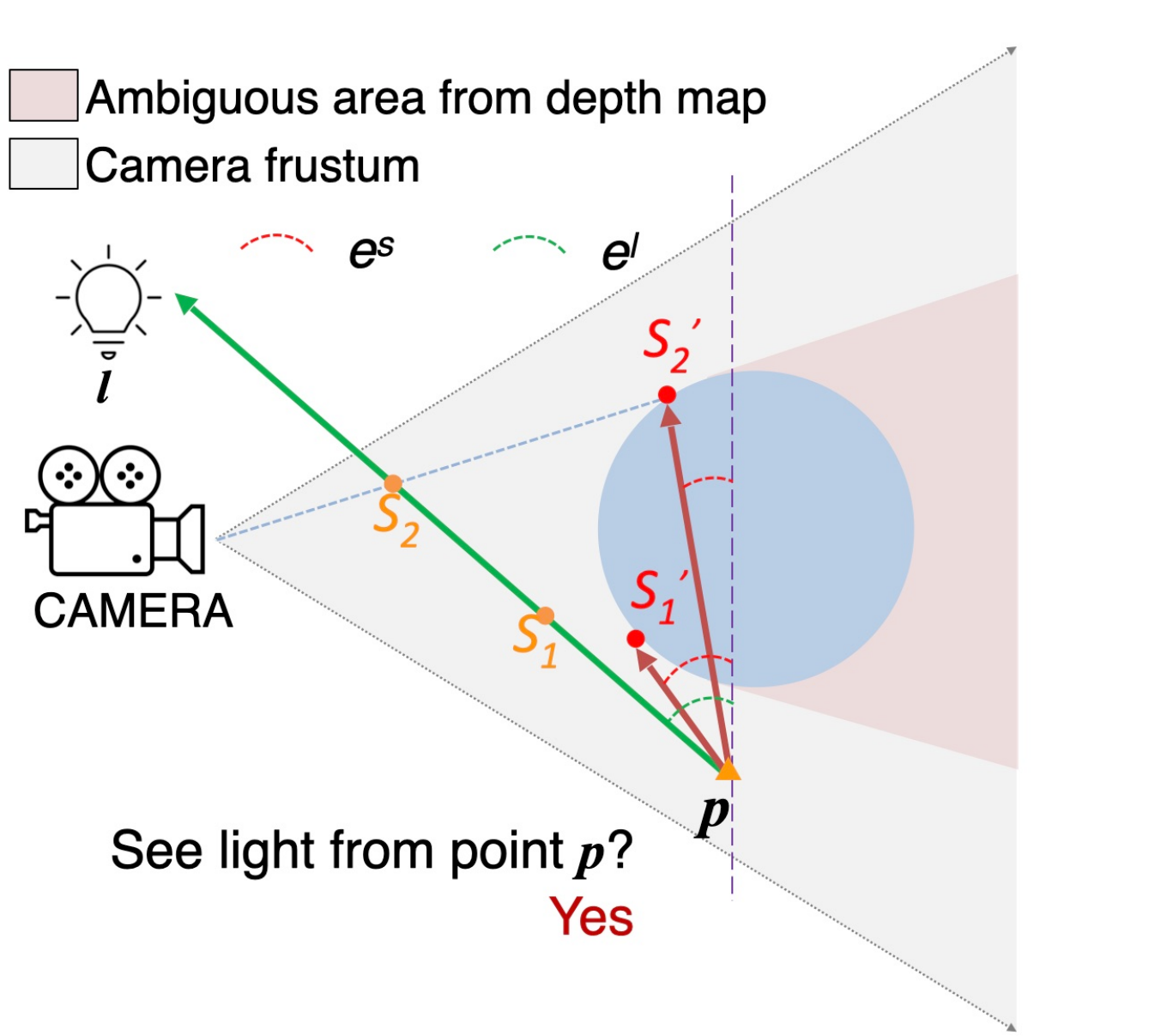}
        \caption{Front lighting}
    \end{subfigure}
    \begin{subfigure}{0.24\textwidth}
        \includegraphics[width=\linewidth]{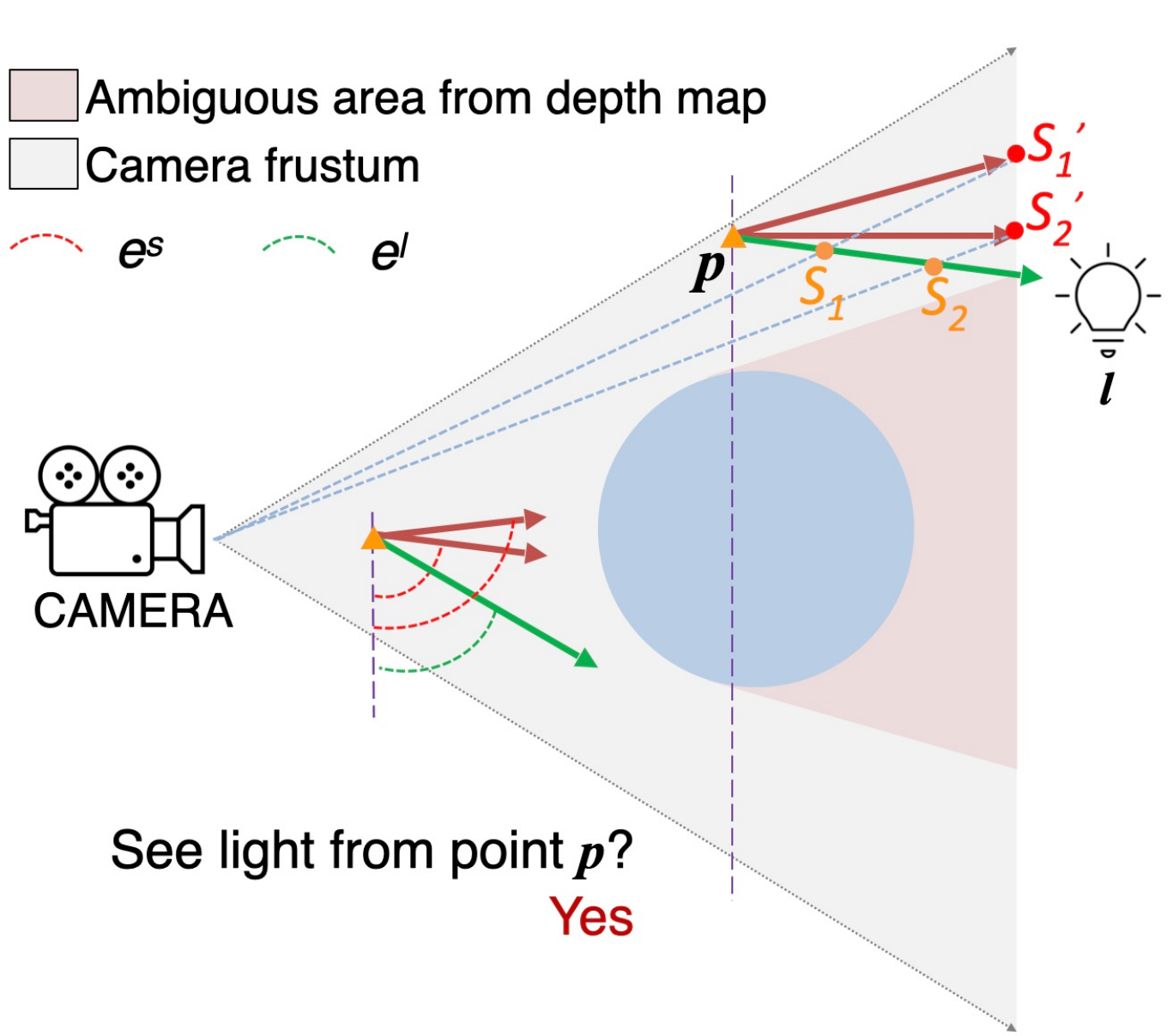}
        \caption{Back lighting}
    \end{subfigure}
    \begin{subfigure}{0.24\textwidth}
        \includegraphics[width=\linewidth]{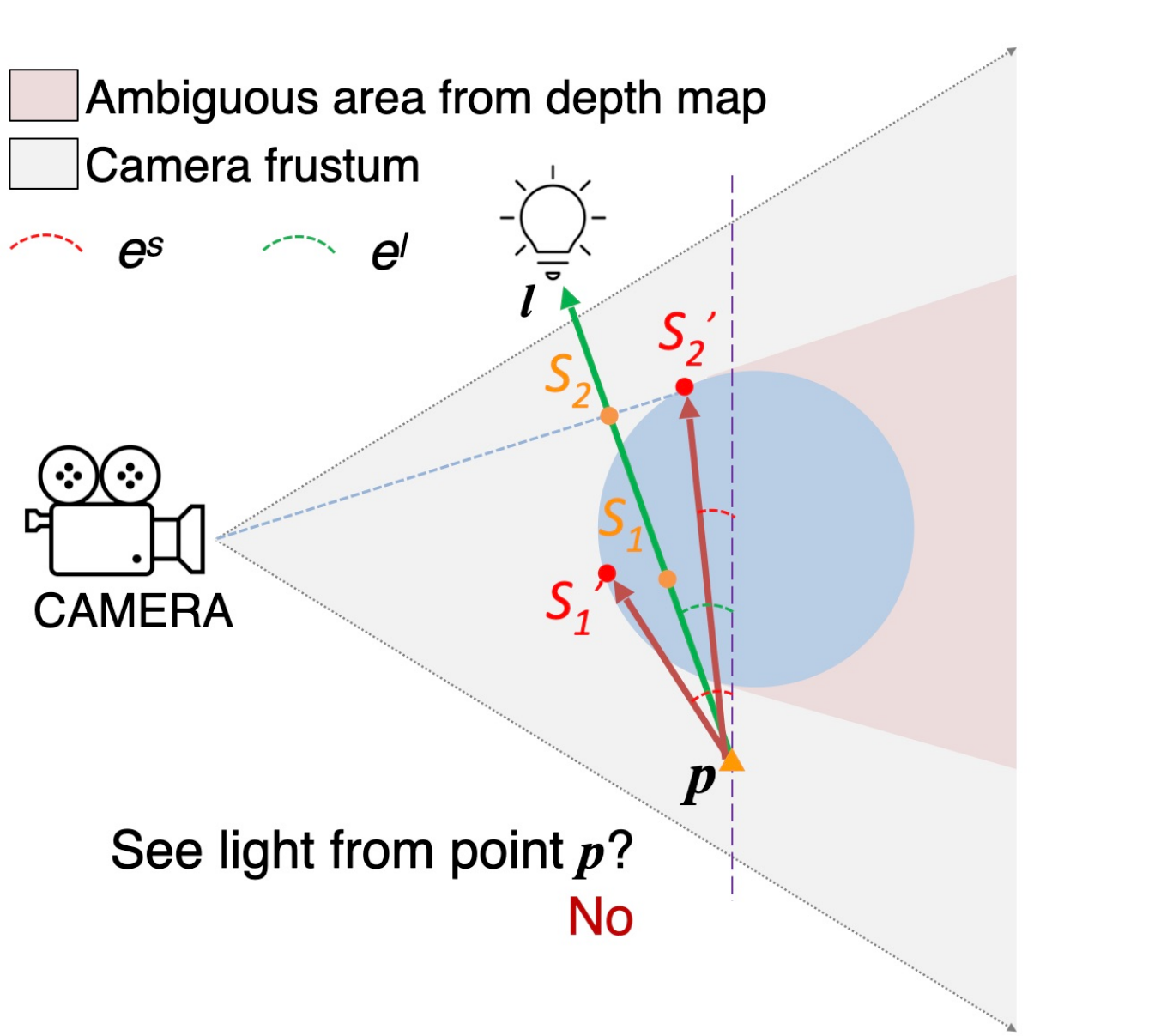}
        \caption{Shadow}
    \end{subfigure}
    \begin{subfigure}{0.24\textwidth}
        \includegraphics[width=\linewidth]{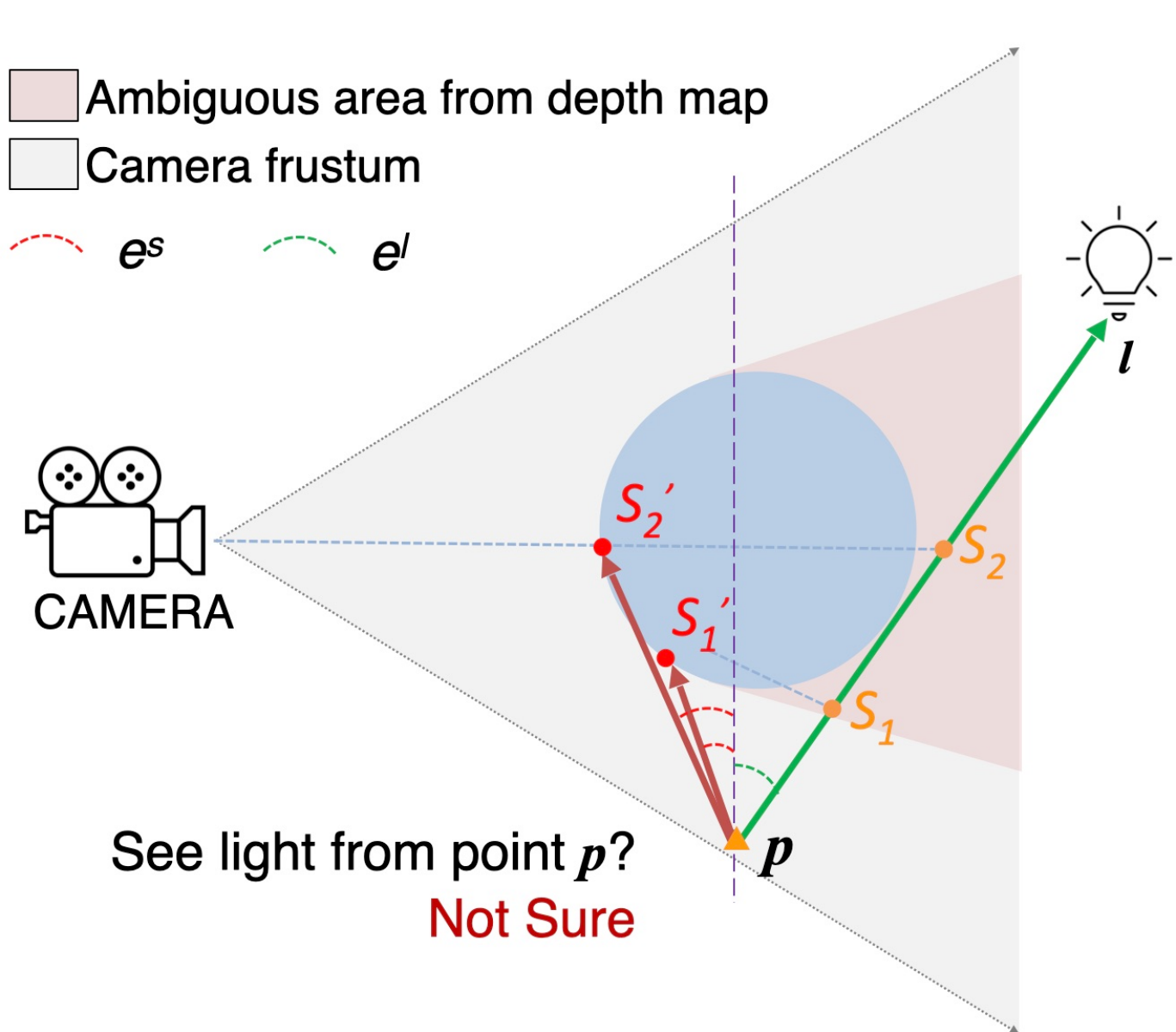}
        \caption{Ambiguous shadow}
        \label{fig:ambiguous_shadow}
    \end{subfigure}
\caption{ Elevation difference calculation. Scene objects are shown in \textcolor{blue}{blue}. Each 2D pixel is lifted to 3D at point $\vp$ , from which a ray is cast toward the light source $\vl$. Along this ray, we uniformly sample $n$ \textcolor{orange}{points $\mS$} within the valid front-facing camera frustum. Each sampled point is reprojected onto the image plane to retrieve its depth from the predicted depth map, yielding a set of \textcolor{red}{reprojected points $\mS'$}. The elevation angles $\ve^s$ of these reprojected points are then compared with the elevation angle $\ve^l$ of the light ray to compute elevation difference $\ve^d$. 
% Light occlusion at $\vp$ indicates potential shadow.
If the light is occluded when viewed from point $\vp$, that point is likely to lie in shadow.
}
\label{fig:sample}
\vspace{-0.3cm}
\end{figure}

\textbf{Lifting 2D to 3D.}  Each pixel in the 2D image is lifted into 3D space by using the predicted depth map. Specifically, given the pixel coordinates and the estimated depth, we compute the 3D position in the camera coordinate frame via the inverse camera projection:
\begin{equation}
\vp = \mD(u, v) \cdot \mK^{-1} \begin{bmatrix} u & v & 1 \end{bmatrix}^{\top},
\label{eq:lift}
\end{equation}
where $\mD(u,v)$ is the depth value at pixel $(u, v)$, $\mK$ is the camera intrinsic matrix and $\vp \in \mathbb{R}^3$ is the resulting 3D point in camera coordinates.
To provide intuition, we use a simple example—a blue ball—in Fig.~\ref{fig:sample}. The diagram shows a cross-sectional view of the 3D space. Because depth maps capture only 2.5D geometry, they fail to provide depth for occluded surfaces. Such ambiguous or occluded regions are marked in pink in Fig.~\ref{fig:sample}. 

\textbf{Ray Sampling.} %\textcolor{magenta}{We adopt standard ray sampling}. 
As shown in the Fig.~\ref{fig:sample}, \textcolor{green}{a ray} is cast from each lifted point $\vp$ toward the light source $\vl$. We uniformly sample $N$ points along this ray, forming the set $\textcolor{orange}{S} = \{ \vp + \delta_n (\vl - \vp) \}_{n=1}^N$, where $\delta_n$ are evenly spaced scalar distances constrained within the front-facing camera frustum. Each sampled point in $\mS$ is projected back to the image plane $(u_n', v_n')$ to retrieve its corresponding depth from the predicted depth map, and following Eq.~\ref{eq:lift}, resulting in reprojected points $\textcolor{red}{S'}= \{ \emS_n' \}_{n=1}^N$.
% \begin{equation}
% \emS_i' = \mD(u_i', v_i') \cdot \frac{\emS_{i,1:3}}{\emS_{i,3}}.
% \end{equation}
Points with infinite depth are marked as invalid and excluded from further computation.

\textbf{Elevation Difference Calculation.} For each valid point in $\mS'$, we compute the elevation angle $\ve^s_n$ and compare it with the elevation angle of the light ray $\ve^l$ to determine potential light occlusion:
\begin{equation}
\ve^d_n = \ve^s_n - \ve^l, \quad
\ve^s_n = \arcsin \left( \frac{\vv^{s}_n \cdot \vn }{\left\| \vv^{s}_n \right\|_2} \right),  \quad
\ve^l = \arcsin \left( \frac{\vv^{l}  \cdot \vn}{\left\| \vv^{l} \right\|_2} \right), 
\label{eq:surface_a}
\end{equation}
where $\vv^{s}_n = \mS'_{n}- \vp$ is the vector from point $\vp$ to surface point $\mS'_{n}$, $\vv^{l} = \vl- \vp$ is the vector from point $\vp$ to light point $\vl$, and $\vn = \begin{bmatrix} 0 & 0 & 1 \end{bmatrix}^{\top}$ is the normal vector of camera plane. Here, $\cdot$ represents the dot product.
The elevation angle is measured relative to the camera plane and lies within the range ($-\pi/2$, $\pi/2$). Consequently, the LGI values $\ve^d$ are naturally bounded within ($-\pi$, $\pi$), which is favorable for deep network input stability.

\begin{figure}[htbp]
\centering
    \begin{subfigure}{0.19\textwidth}
        \includegraphics[width=\linewidth]{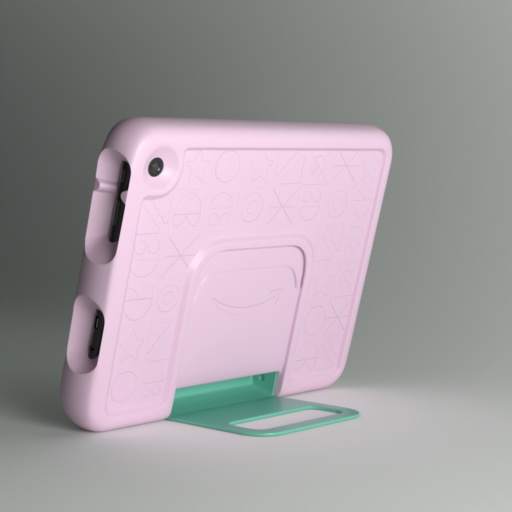}
        \caption{Shadowed Image}
    \end{subfigure}
    \begin{subfigure}{0.19\textwidth}
        \includegraphics[width=\linewidth]{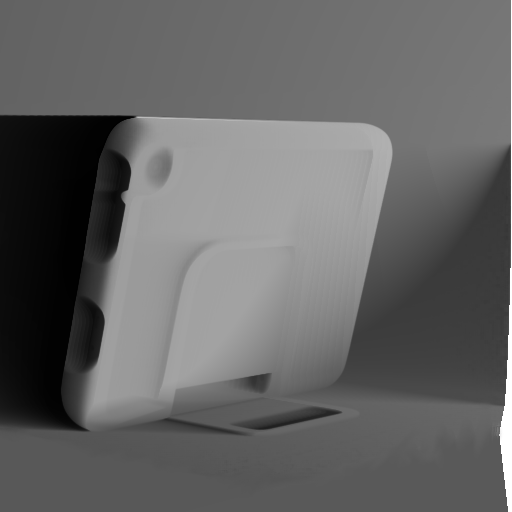}
        \caption{$\vc^m_1$}
    \end{subfigure}
    \begin{subfigure}{0.19\textwidth}
        \includegraphics[width=\linewidth]{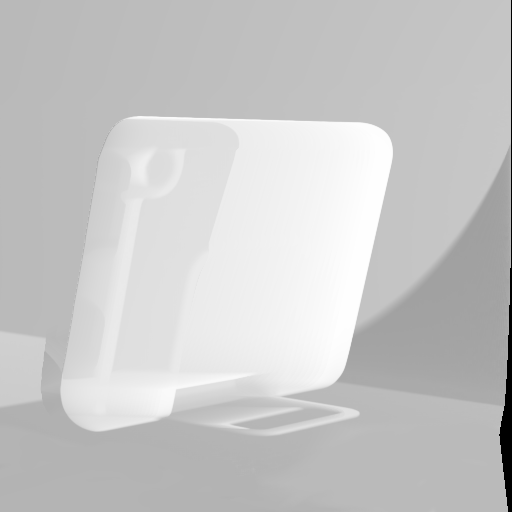}
        \caption{$\vc^m_2$}
    \end{subfigure}
    \begin{subfigure}{0.19\textwidth}
        \includegraphics[width=\linewidth]{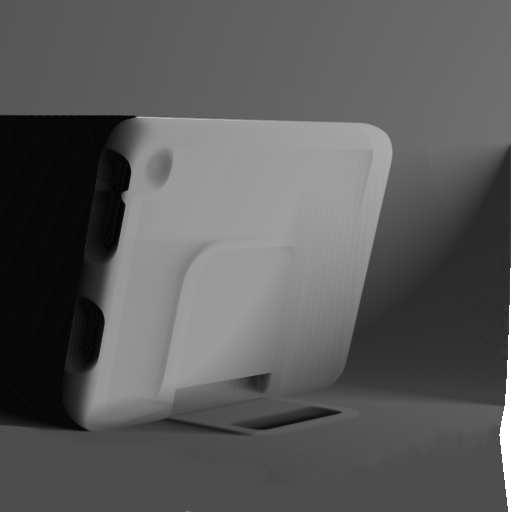}
        \caption{$\vc^m_3$}
    \end{subfigure}
    \begin{subfigure}{0.19\textwidth}
        \includegraphics[width=\linewidth]{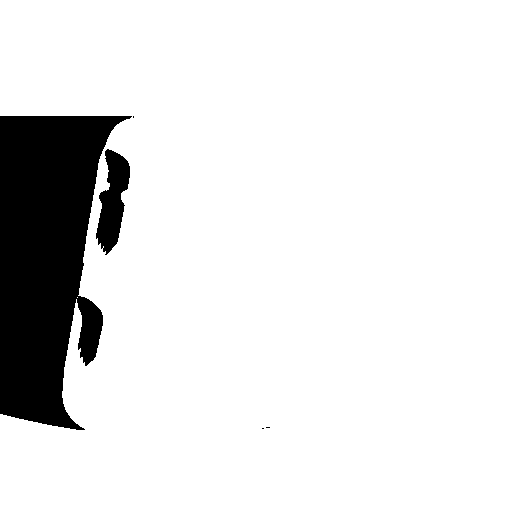}
        \caption{Hard Shadow}
        \label{fig:hard_s}
    \end{subfigure}
\caption{Example of LGI maps. They indicate shadows cast in the environment and also reveal self-shadowing and shading effects, which can be useful for relighting.
}
\label{fig:lgi_visualization}
\end{figure}

\textbf{LGI Maps Generation.} Since the depth map provides only 2.5D information, it cannot capture occluded regions behind foreground objects. An example of ambiguous shadows is shown in Fig.~\ref{fig:ambiguous_shadow}. If the 2.5D representation sufficiently captures the scene geometry (e.g., piece-like objects), a hard shadow mask (Fig.~\ref{fig:hard_s}) can be derived by checking whether surface elevation matches light elevation using the condition $\min \left| \ve^d \right| < \eta$, where $\eta$ is a small threshold (set to $5^{\circ}$ in our visualization).

In cases where geometric information is insufficient, we defer occlusion reasoning to the model by embedding three-channel LGI maps derived from the elevation difference: ($\vc^m_1$) the minimum elevation difference, indicating the potential start of occlusion; ($\vc^m_2$) the maximum elevation difference, indicating the potential end of occlusion; and ($\vc^m_3$) the value with the smallest absolute difference, representing the most likely point of direct occlusion:
\begin{equation}
\vc^m_1 = \min_{n=1}^N \ve^d_n,  \quad \vc^m_2 = \max_{n=1}^N \ve^d_n, \quad 
\vc^m_3 = \ve^d_{\text{i}^{\star}}, \quad \text{i}^{\star} = \arg\min_{1 \leq n \leq N} \left| \ve^d_n \right|.
\label{eq:embed_e}
\end{equation}
An example visualization of these embeddings is shown in Fig.~\ref{fig:lgi_visualization}. They capture not only the cast shadows within the environment but also reveal self-shadowing and shading effects.
We use the LGI maps ($\vc^m$) as conditioning signals by concatenating them with image features to guide the bridge-matching in Eq.~\ref{eq:target_z}, providing light-aware occlusion cues.

\subsection{Extension to image harmonization} 
In addition to explicit light condition setting, we also extend our method to the task of image harmonization, where the light sources are implied by the image,
demonstrating its robustness and generalization to outdoor and realistic images. 
The core architecture of our method remains unchanged. We introduce an additional light estimation network
to infer lighting conditions directly from the composited image, removing the need for explicit lighting input.
Since our LGI maps are fully differentiable, they enable self-supervised light estimation. Specifically, we use the shadow mask to supervise the predicted lighting. Further details are provided in the supplementary material.

% \[
% K = 
% \begin{bmatrix}
% W & 0 & W/2 \\
% 0 & H & H/2 \\
% 0 & 0 & 1
% \end{bmatrix}
% \]

\subsection{ShadRel Dateset} %{Synthetic Dataset}
As no dataset currently exists to support training for coupled light transport, we construct the first large-scale 
% To maximize generality, we have assembled an extensive 
dataset consisting of 817K virtual 3D objects, all crafted by professional 3D artists. The textures of these 3D assets assume the physically-accurate principled Bidirectional Scattering Distribution Function (BSDF) formulation by \cite{burley2012_principled_bsdf}, capable of simulating a variety of materials including those of glossy, metallic, or transparent appearances.
This dataset not only enables effective training but also provides a benchmark for evaluating future methods, and research in coupled illumination and shadow modeling.
% establishing a foundation for research in coupled illumination and shadow modeling.

To simulate real-world lighting and shadow effects, we render images in Blender using a photorealistic path tracer \cite{blender_cycles_44}. For each object, we generate following types of images for training:
\begin{itemize}
\item \textbf{Input images}: the object lit by a random High Dynamic Range Imaging (HDRI) environment map drawn from a dataset of panoramas collected from the Internet.
\item \textbf{Background images}: a planar floor and/or vertical wall of random RGB color, illuminated by a single target point light, where the object is to be inserted.
\item \textbf{Target images}: the composite of object and background under the same point light --- this is the supervision signal for output image.
Additionally, a corresponding depth map is also rendered for generating light‑occlusion maps.
\end{itemize}

Each object is augmented by sampling 4 random camera poses and focal lengths; for every camera we further sample 5 point-light-background configurations, totaling 20 target images per object. Point lights vary in position, color, and radius, producing shadows of diverse direction and softness.

Above images are stored in physical units of scene radiance. During training, we apply real-world camera response functions provided by \cite{colour_science_046}, and tone-map results to the $[0,1]$ standard intensity range. The reader is referred to supplementary material for details about dataset composition, rendering configuration and costs.

\section{Experiments}
\label{sec:experiments}
We evaluate our approach on three benchmark datasets for light-controllable generation and image harmonization (implicit lighting), and present qualitative visualizations on real-world images to demonstrate generalization capability. 
% We also provide ablation studies of the components. 
We further conduct ablation studies, including component analysis, depth-estimation variants, and efficiency evaluation, to validate our design choices.
% to analyze the contribution of each component.

% \subsection{Datasets and Evaluation Metrics}
\noindent\textbf{Datasets and Evaluation Metrics.}
Our method jointly addresses controllable shadow generation and image relighting. As no published dataset supports this combined task, we build a new ShadRel dataset for training and primary evaluation. 
We also assess the controllable shadow generation component on the public benchmark of~\cite{tasar2024controllable}, which targets clean-background shadow generation.
For the joint task, we compare with LBM~\cite{chadebec2025lbm}, measuring overall image quality (RMSE, SSIM), shadow quality (overall: BER/IoU, shadow: RMSE/SSIM/BER), and object relighting quality (object RMSE/SSIM). 
% following~\cite{tasar2024controllable,zhao2025shadow}. 
We further provide qualitative comparison with SwitchLight~\cite{kim2024switchlight}.
% For the joint task, we compare with LBM~\cite{chadebec2025lbm}, measuring overall image quality, shadow quality, and object relighting quality. Following~\cite{tasar2024controllable,zhao2025shadow}, overall image quality is measured using Root Mean Squared Error (RMSE) and Structural Similarity Index Measure (SSIM). Shadow quality is evaluated with Intersection over Union (IoU) and Balanced Error Rate (BER). In addition, we compute region-specific metrics: shadow RMSE, shadow SSIM, and shadow BER within shadow regions, and object RMSE and object SSIM within object regions.
% Additionally, we provide a qualitative comparison with the state-of-the-art human relighting method SwitchLight~\cite{kim2024switchlight}.
%
For controllable shadow generation, we compare against CSG~\cite{tasar2024controllable} on their benchmark and metrics (IoU and RMSE). 
Finally, we extend our method to image harmonization, following ~\cite{liu2024shadow} on the DESOBAv2 dataset~\cite{liu2023desobav2}, reporting global/local RMSE (GR, LR), SSIM (GS, LS), and BER (GB, LB).

% \subsection{Implementation Details}
\noindent\textbf{Implementation Details}. 
We initialize our framework from Stable Diffusion XL v1.0~\cite{podell2023sdxl}.
All experiments use $512 \times 512$ images with a batch size of 5.
% across 8 NVIDIA V100-SXM2 GPUs. 
Training employs AdamW optimizer~\cite{loshchilov2017decoupled} with a learning rate of $3 \times 10^{-5}$.
The threshold $\tau$ in Eq.~\ref{eq:lx} is empirically set to 0.01, and a dilation operation with a kernel size of 17 is applied. In the final loss function, we set the objective weighting factor $\lambda = 10$. The number of sample points $N$ is set to 16.
All models are implemented in PyTorch and trained end-to-end.

\subsection{Experimental Results}
\begin{figure}[htbp]
\begin{center}
\includegraphics[width=\textwidth]{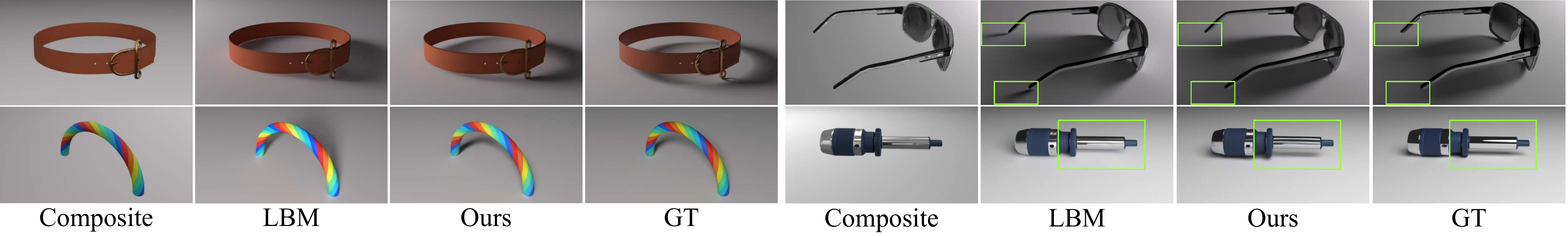}
% \fbox{\rule[-.5cm]{0cm}{4cm} \rule[-.5cm]{4cm}{0cm}}
\end{center}
\vspace{-0.4cm}
\caption{Qualitative comparison with LBM on synthetic object images. Our method produces shadows and relighting effects consistent with geometry and lighting conditions.
% , and accurately models inter-reflections between objects and the environment.
}
\label{fig:compare_lbm}
\vspace{-0.1cm}
\end{figure}

\noindent\textbf{Evaluation on joint shadow synthesis and relighting}. Tab.~\ref{tab:ours_comparison} presents quantitative results on the ShadRel dataset. For fairness, we retrain LBM with the same settings. Our method outperforms LBM in controllable shadow generation and object relighting, demonstrating the effectiveness of our light-geometry interaction formulation.
Fig.~\ref{fig:compare_lbm} shows qualitative comparisons, where our method responds more accurately to light and preserves object geometry, producing realistic shadows, precise relighting, and capturing complex object–environment light transport.
%benefiting from our high-quality dataset that supports detailed supervision of these phenomena.

\begin{table*}[htbp]
\centering
\footnotesize
\caption{Joint shadow synthesis and relighting results on ShadRel dataset.}
\renewcommand{\arraystretch}{1.1}
\setlength{\tabcolsep}{5pt}
\begin{tabular}{l cccc ccc cc}
\toprule
\multirow{2}{*}{\textbf{Method}} & \multicolumn{4}{c}{\textbf{Overall}} & \multicolumn{3}{c}{\textbf{Shadow region}} & \multicolumn{2}{c}{\textbf{Object region}} \\
 \cmidrule(lr){2-5} \cmidrule(lr){6-8} \cmidrule(lr){9-10}
& RMSE $\downarrow$ & SSIM $\uparrow$ & BER $\downarrow$ & IoU $\uparrow$ & RMSE $\downarrow$ & SSIM  $\uparrow$ & BER $\downarrow$ & RMSE $\downarrow$ & SSIM $\uparrow$ \\
\midrule
LBM & 0.0417 & 0.7148 & 0.0847 & 0.7166 & 0.1543 & 0.5690 & 0.1549 & 0.0298 & 0.6797\\
Ours & \textbf{0.0334} & \textbf{0.7227} & \textbf{0.0588} & \textbf{0.8096} & \textbf{0.0898} & \textbf{0.6195} & \textbf{0.1103} & \textbf{0.0282} & \textbf{0.6875}\\
\bottomrule
\end{tabular}
\label{tab:ours_comparison}
\vspace{-0.4cm}
\end{table*}

% \begin{figure}[htbp]
% \begin{center}
% \includegraphics[width=\textwidth]{image/jasper_vis.pdf}
% % \fbox{\rule[-.5cm]{0cm}{4cm} \rule[-.5cm]{4cm}{0cm}}
% \end{center}
% \caption{Visual example of clean-background shadow generation, demonstrating that our light-geometry interaction approach provides accurate control over shadow shapes.}
% \label{fig:jasper_vis}
% \end{figure}

\begin{figure}[htbp]
\begin{center}
\includegraphics[width=\textwidth]{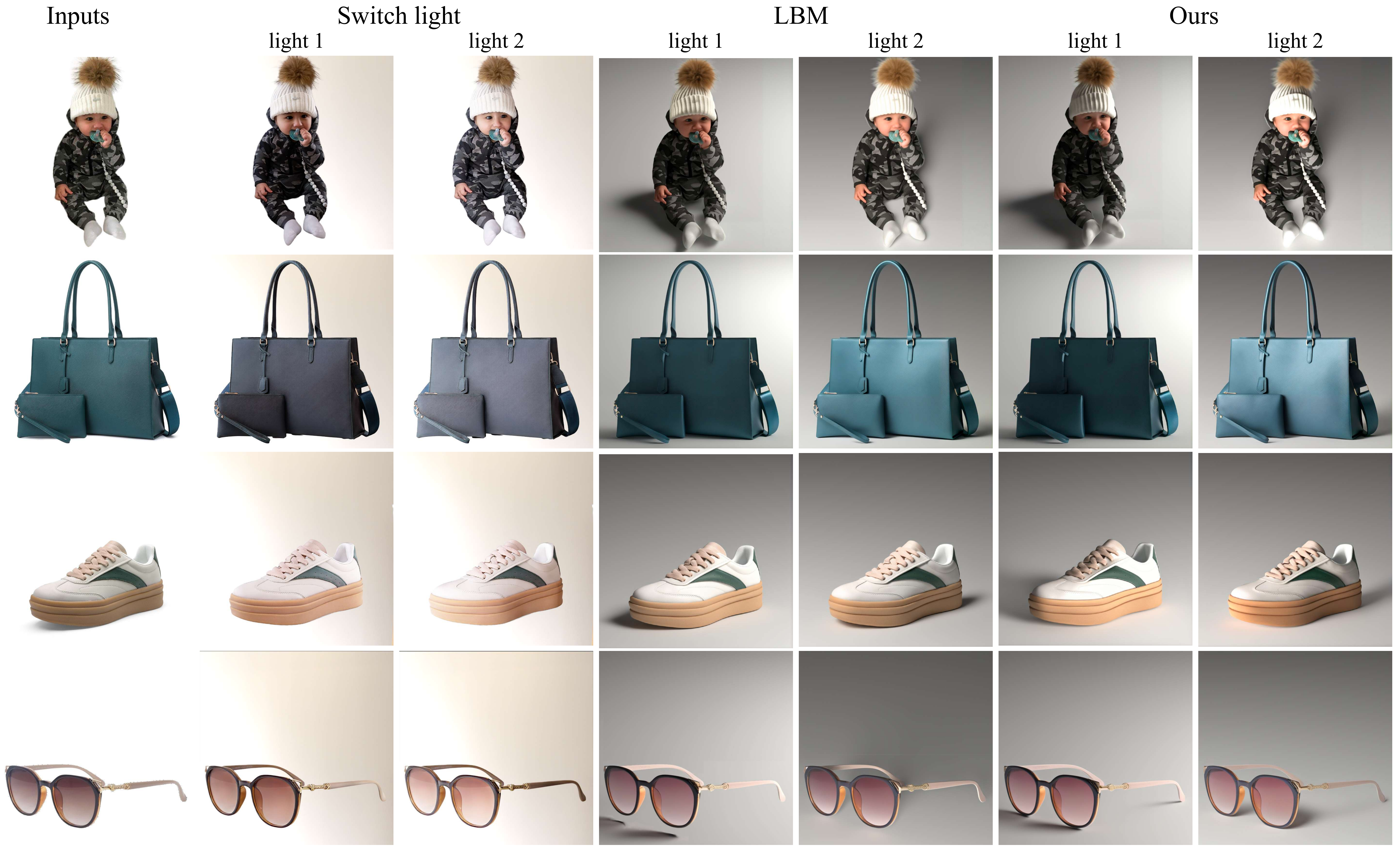}
% \fbox{\rule[-.5cm]{0cm}{4cm} \rule[-.5cm]{4cm}{0cm}}
\end{center}
\vspace{-0.2cm}
\caption{Qualitative comparison with SwitchLight (object relighting without shadow generation) and LBM on real object insertions. Despite being trained solely on synthetic object images, our method outperforms SOTA relighting approaches on real-world object images.}
\label{fig:compare_vis}
\vspace{-0.5cm}
\end{figure}

\noindent\textbf{Qualitative analysis on real images}. We further provide qualitative comparisons with SwitchLight and LBM on real-word object images in Fig.~\ref{fig:compare_vis}. SwitchLight is designed for human relighting with point light sources and does not support shadow generation. We compare their method with ours on both human relighting and other real-world objects.
%\footnote{Additional qualitative results are provided in the supplementary video.}
Our method produces realistic relighting effects, including for humans, while faithfully adhering to the light direction and preserving the original object colors. In addition, it generates more accurate shadows that align with both the light direction and object geometry, avoiding floating artifacts or inconsistencies in shadow orientation.
Despite being trained solely on synthetic data, without any real-world or human-specific samples, our method generalizes well to to real images, including human portraits and complex objects.
% The results show that our method not only achieves better performance on human relighting but also generates more realistic shadows. Despite being trained on a synthetic dataset, our method generalizes well to real-world data, enabled by our physically inspired approach and high-quality dataset.

\begin{figure}[htbp]
\begin{center}
\includegraphics[width=\textwidth]{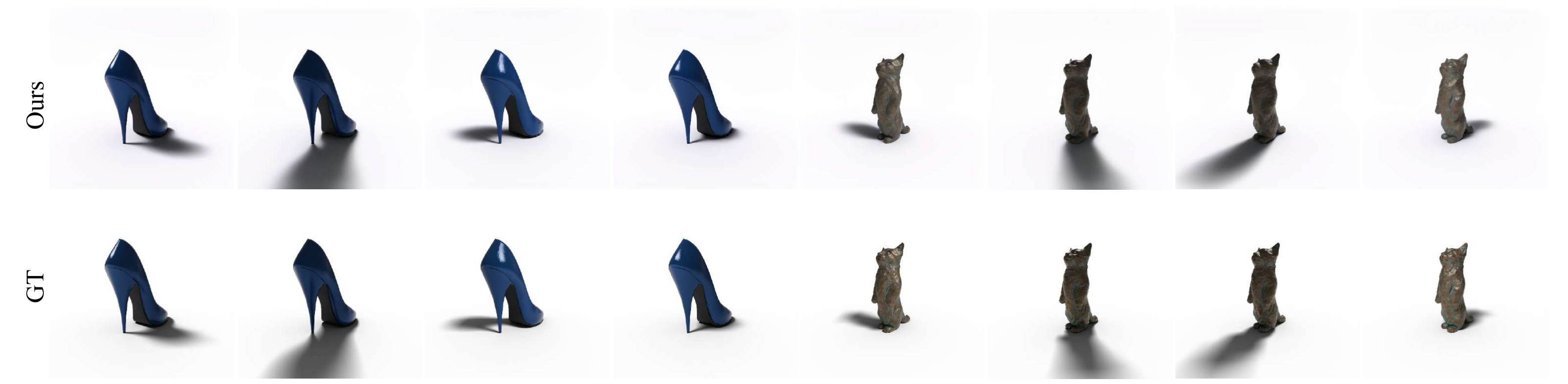}
% \fbox{\rule[-.5cm]{0cm}{4cm} \rule[-.5cm]{4cm}{0cm}}
\end{center}
\caption{Visual example of clean-background shadow generation, demonstrating that our light-geometry interaction approach provides accurate control over shadow shapes.}
\label{fig:jasper_vis}
\end{figure}

\noindent\textbf{Evaluation on clean-background shadow generation}.
Tab.~\ref{tab:jasper_b} reports quantitative comparisons with CSG~\cite{tasar2024controllable} on their benchmark. As CSG did not release training data or model weights,
we adapt our approach to their clean-background setting by training on ShadRel with pure white backgrounds. Results show that our light–geometry interaction formulation enables more accurate control of shadow shapes. 
% Additional qualitative results are provided in the Supplementary.
Qualitative results in Fig.~\ref{fig:jasper_vis} further confirm that our method produces shadows with accurate shapes and realistic density.

\begin{table*}[!htbp]
\centering
\footnotesize
\caption{Comparison with CSG~\cite{tasar2024controllable} on their benchmark.}
\vspace{-0.2cm}
\renewcommand{\arraystretch}{1.1}
\setlength{\tabcolsep}{4pt}
\begin{tabular}{l cc cc cc}
\toprule
\multirow{2}{*}{\textbf{Method}} & \multicolumn{2}{c}{Track 1 (Softness Control)} & \multicolumn{2}{c}{Track 2 (Horz. D. Control)} & \multicolumn{2}{c}{Track 3 (Vert. D. Control)} \\
 \cmidrule(lr){2-3} \cmidrule(lr){4-5} \cmidrule(lr){6-7}
& IoU $\uparrow$ & RMSE $\downarrow$ 
& IoU $\uparrow$ & RMSE $\downarrow$
& IoU $\uparrow$ & RMSE $\downarrow$ \\
\midrule
CSG & 0.818 & \textbf{0.021} & 0.780 & \textbf{0.030} & 0.776 & 0.028 \\
Ours & \textbf{0.821} & \textbf{0.021} & \textbf{0.798} & \textbf{0.030} & \textbf{0.785} & \textbf{0.027} \\
\bottomrule
\end{tabular}
% Arrows indicate whether higher or lower values are better.}
\label{tab:jasper_b}
\vspace{-0.4cm}
\end{table*}

% We present qualitative results on the CSG benchmark in Fig.~\ref{fig:jasper_vis}. The visualizations confirm that our method generates shadows with accurate shapes and realistic density.

\begin{table*}[htbp]
\centering
\footnotesize
\caption{Comparison on Image Harmonization on DESOBAv2 Dataset.}
% \vspace{-0.2cm}
\renewcommand{\arraystretch}{1.1}
\setlength{\tabcolsep}{3pt}
\begin{tabular}{l cccccc cccccc}
\toprule
\multirow{2}{*}{\textbf{Method}} & \multicolumn{6}{c}{\textbf{BOS Test Images}} & \multicolumn{6}{c}{\textbf{BOS-free Test Images}} \\
 \cmidrule(lr){2-7} \cmidrule(lr){8-13}
& GR $\downarrow$ & LR $\downarrow$ & GS $\uparrow$ & LS $\uparrow$ & GB $\downarrow$ & LB $\downarrow$ & GR $\downarrow$ & LR $\downarrow$ & GS $\uparrow$ & LS $\uparrow$ & GB $\downarrow$ & LB $\downarrow$ \\
\midrule
  ShadowGAN~\citenum{zhang2019shadowgan} & 7.511 & 67.464 & 0.961 & 0.197 & 0.446 & 0.890 & 17.325 & 76.508 & 0.901 & 0.060 & 0.425 & 0.842 \\
 Mask-SG~\citenum{hu2019mask}  & 8.997 & 79.418 & 0.951 & 0.180 & 0.500 & 1.000 & 19.338 & 94.327 & 0.906 & 0.044 & 0.500 & 1.000 \\
 AR-SG~\citenum{liu2020arshadowgan} & 7.335 & 58.037 & 0.961 & 0.241 & 0.383 & 0.761&16.067 & 63.713 & 0.908 & 0.104 & 0.349 & 0.682 \\
SGRNet~\citenum{hong2022shadow} & 7.184 & 68.255 & 0.964 & 0.206 & 0.301 &0.596& 15.596 & 60.350 & 0.909 & 0.100 & 0.271 & 0.534 \\
DMASNet~\citenum{tao2024shadow} & 8.256 & 59.380 & 0.961 & 0.228 & 0.276 &0.547& 18.725 & 86.694 & 0.913 & 0.055 & 0.297 & 0.574 \\
SGDiffusion~\citenum{liu2024shadow} & 6.098 & 53.611 & \textbf{0.971} & 0.370 & 0.245 &0.487& 15.110 & 55.874 & 0.913 & 0.117 & 0.233 & 0.452 \\
SGDGP~\citenum{zhao2025shadow} & \textbf{5.896} & 46.713 & 0.966 & 0.374 & \textbf{0.213} &0.423 & 13.809 & 55.616 & \textbf{0.917} & 0.166 & \textbf{0.197} & 0.384 \\
Ours & 5.900 & \textbf{44.753} & 0.961 & \textbf{0.415} & 0.239 & \textbf{0.415} & \textbf{12.979} & \textbf{52.543} & 0.912 & \textbf{0.201} & 0.213 & \textbf{0.358} \\
% Ours & \textbf{6.94} & \textbf{2.933} & 0.906 & \textbf{0.996} & \textbf{0.299} & \textbf{0.298} & \textbf{12.878} & \textbf{9.305} & 0.876 & \textbf{0.980} & \textbf{0.198} & \textbf{0.194} \\
\bottomrule
\end{tabular}
\label{tab:DESOBAv2_comparison}
\vspace{-0.2cm}
\end{table*}

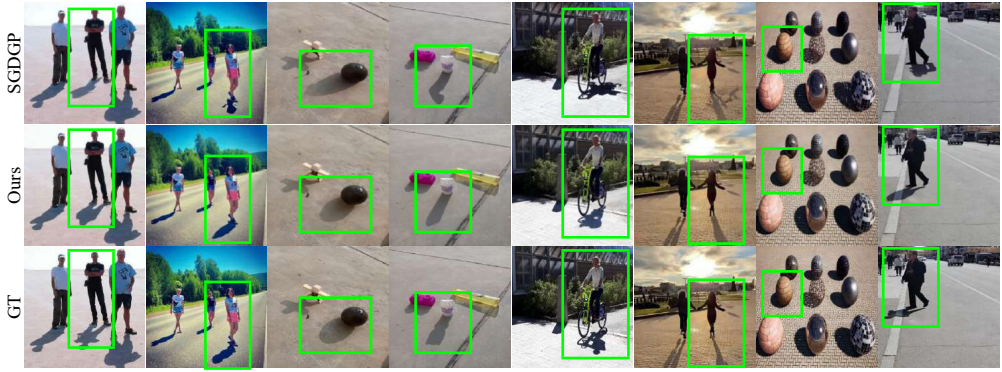
\begin{figure}[!htbp]
\begin{center}
\begin{overpic}[width=\textwidth]{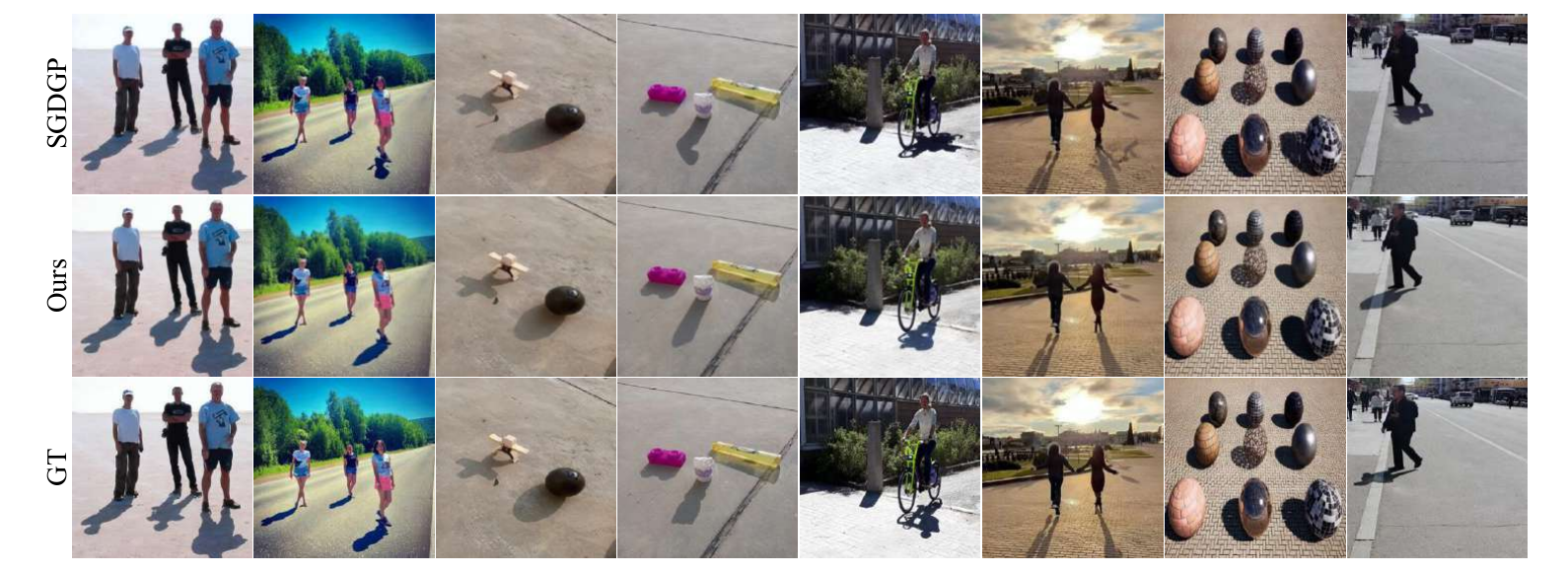}
% box 1
  \put(9,3){\color{green}\linethickness{1pt}\framebox(4,9){}}
  \put(9,14.5){\color{green}\linethickness{1pt}\framebox(4,9){}}
  \put(9,26){\color{green}\linethickness{1pt}\framebox(4,9){}}

% box 2
  \put(22,1){\color{green}\linethickness{1pt}\framebox(4,8){}}
  \put(22,13){\color{green}\linethickness{1pt}\framebox(4,8){}}
  \put(22,25){\color{green}\linethickness{1pt}\framebox(4,8){}}

% box 3
  \put(31,2.5){\color{green}\linethickness{1pt}\framebox(6.5,5){}}
  \put(31,14){\color{green}\linethickness{1pt}\framebox(6.5,5){}}
  \put(31,26){\color{green}\linethickness{1pt}\framebox(6.5,5){}}
  
% box 4
  \put(42,2.5){\color{green}\linethickness{1pt}\framebox(5,5.5){}}
  \put(42,14){\color{green}\linethickness{1pt}\framebox(5,5.5){}}
  \put(42,26){\color{green}\linethickness{1pt}\framebox(5,5.5){}}

% box 5
  \put(56,2){\color{green}\linethickness{1pt}\framebox(6,10){}}
  \put(56,13.5){\color{green}\linethickness{1pt}\framebox(6,10){}}
  \put(56,25){\color{green}\linethickness{1pt}\framebox(6,10){}}

% box 6
  \put(68,1.2){\color{green}\linethickness{1pt}\framebox(5,8){}}
  \put(68,12.9){\color{green}\linethickness{1pt}\framebox(5,8){}}
  \put(68,24.5){\color{green}\linethickness{1pt}\framebox(5,8){}}

% box 7
  \put(75,6){\color{green}\linethickness{1pt}\framebox(3.5,4){}}
  \put(75,17.7){\color{green}\linethickness{1pt}\framebox(3.5,4){}}
  \put(75,29.3){\color{green}\linethickness{1pt}\framebox(3.5,4){}}
  
% box 8
  \put(86.5,5){\color{green}\linethickness{1pt}\framebox(5,7.2){}}
  \put(86.5,16.6){\color{green}\linethickness{1pt}\framebox(5,7.2){}}
  \put(86.5,28.3){\color{green}\linethickness{1pt}\framebox(5,7.2){}}
 
\end{overpic}
\end{center}
\vspace{-0.3cm}
\caption{Qualitative comparison with SGDGP~\cite{zhao2025shadow} . Our method achieves higher accuracy in shadow regions by aligning more closely with object geometry and light direction.}
\label{fig:compare_SGDGP}
\vspace{-0.2cm}
\end{figure}

% \subsection{Generalization to Harmonization}
% \label{sec:harmonization}
\noindent\textbf{Generalization to Harmonization}.
% In this section, we extend our method to the task of image harmonization, demonstrating its robustness and generalization to outdoor and realistic images. The core architecture of our method remains unchanged. We introduce an additional light estimation network
% % \footnote{The architecture of the light estimation network is provided in the supplementary material.} 
% to infer lighting conditions directly from the composited image, removing the need for explicit lighting input.
% Since our LGI maps are fully differentiable, they enable self-supervised light estimation. Specifically, we use the shadow mask to supervise the predicted lighting. Further details are provided in the supplementary material.
%
We compare our method with SOTA approaches
% ShadowGAN~\cite{zhang2019shadowgan}, Mask-SG~\cite{hu2019mask}, AR-SG~\cite{liu2020arshadowgan}, SGRNet~\cite{hong2022shadow}, DMASNet~\cite{tao2024shadow}, SGDiffusion~\cite{liu2024shadow} and SGDGP~\cite{zhao2025shadow} 
on the DESOBAv2~\cite{liu2023desobav2}, with quantitative results in Tab.~\ref{tab:DESOBAv2_comparison} and qualitative comparison in Fig.~\ref{fig:compare_SGDGP}. Our method delivers overall performance comparable to the top-performing approach, SGDGP, while achieving higher accuracy in shadow regions. Although SGDGP also leverages geometry priors, it restricts them to 2D rotated bounding boxes and shadow templates. Qualitative comparisons reveal that the main challenge lies in generating shadows that align with object geometry and light direction. Our method addresses this challenge more effectively, as the LGI-based framework directly embeds light–geometry interactions in a space-aligned representation, providing stronger geometric and illumination cues. This leads to improved robustness in outdoor and realistic scenarios.

\subsection{Ablation Study} 
\label{sec:ablation}

\begin{table*}[htbp]
\centering
\footnotesize
\caption{Analysis of Method Components on ShadRel Dataset}
\vspace{-0.2cm}
\setlength{\tabcolsep}{5pt}
\begin{tabular}{ll cccc cc ccc}
\toprule
 \multicolumn{2}{l}\textbf{Components} & \multicolumn{4}{c}{\textbf{Overall}} & \multicolumn{3}{c}{\textbf{Shadow region}} & \multicolumn{2}{c}{\textbf{Object region}} \\
 \cmidrule(lr){1-2} \cmidrule(lr){3-6} \cmidrule(lr){7-9} \cmidrule(lr){10-11}
W-L1&LGI& RMSE $\downarrow$ & SSIM $\uparrow$ & BER $\downarrow$ & IOU $\uparrow$ & RMSE $\downarrow$ & SSIM  $\uparrow$ & BER $\downarrow$ & RMSE $\downarrow$ & SSIM $\uparrow$ \\
\midrule
&& 0.0408 & 0.7109 & 0.1012 & 0.7193 & 0.1236 & 0.5868 & 0.1923 & 0.0330 & 0.6679 \\
\cmark & & 0.0391 & 0.7148 & 0.0940 & 0.7353 & 0.1154 & 0.5974 & 0.1784 & 0.0317 & 0.6680\\
\cmark &\cmark & 0.0334 & 0.7227 & 0.0588 & 0.8096 &0.0898 & 0.6195 & 0.1103 & 0.0282 & 0.6875  \\
\midrule
\multicolumn{2}{l}{-LGI+Depth}& 0.0388 & 0.7148 &0.0932 & 0.7344 &0.1166 & 0.5938 & 0.1765 & 0.0315 & 0.6719\\
\multicolumn{2}{l}{+Depth}& 0.0339 & 0.7188 & 0.0719 & 0.7921 & 0.0942 & 0.6139 & 0.1364 & 0.0283 & 0.6836\\
\multicolumn{2}{l}{LGI Ch3}& 0.0351 & 0.7179 & 0.0670 & 0.7824 & 0.0932 & 0.6095 & 0.1343 & 0.0290 & 0.6807\\
% \multicolumn{2}{l}{-LGI+DAv2}& 0.0384 & 0.7169 & 0.0904 & 0.7491 & 0.1070 &0.5954 & 0.1631 &0.0314 &0.6719\\
\multicolumn{2}{l}{w/ DAv2}& 0.0334 & 0.7188 & 0.0602 & 0.8148 & 0.0901 & 0.6212 & 0.1269 & 0.0283 & 0.6875 \\
\multicolumn{2}{l}{w/ GT depth} & 0.0326 & 0.7266 & 0.0558 & 0.8107 & 0.0894 & 0.6195 & 0.1070 & 0.0280 &0.6875 \\
\bottomrule
\end{tabular}
\label{tab:ablation}
\vspace{-0.1cm}
\end{table*}

We analyze component contributions in Tab.~\ref{tab:ablation}, using LBM  with standard L1 loss as the baseline. Each added module yields consistent performance gains \footnote{A visual comparison with and without the LGI module is provided in the Appendix~\ref{sec:LGI_study}.}.
Comparing LGI embeddings with direct depth embeddings shows that depth alone (`-LGI+Depth') brings only marginal improvements, while combining both (`+Depth') slightly degrades performance, likely due to noise and misalignment in depth estimation. 
Using only the third channel of LGI (`LGI Ch3') also reduces performance, confirming the necessity of proposed channels. Finally, replacing the original depth estimator with DepthAnythingV2~\cite{yang2024depth} (`w/ DAv2') and ground truth depth produces minimal variation, demonstrating robustness to the choice of depth backbone.
% Results obtained by directly using DepthAnything-predicted depth are denoted as -LGI+DAv2, while those using LGI with DepthAnything depth are denoted as DAv2 LGI. Although direct use of DepthAnything depth yields slightly higher performance, the gap between DAv2 LGI and our full model remains small, indicating that our approach is robust to the choice of depth estimator.
% as indicated by the `w/ DAv2' in Tab.~\ref{tab:ablation}. The results show minimal performance variation, demonstrating the robustness of our method to the choice of depth estimator.
% suggest LGI offers a more structured and task-relevant representation of light-geometry interaction, enabling more effective guidance for shadow and relighting. 

% \begin{figure}[htbp]
% \begin{center}
% \includegraphics[width=\textwidth]{image/wo_LGI.pdf}
% % \fbox{\rule[-.5cm]{0cm}{4cm} \rule[-.5cm]{4cm}{0cm}}
% \end{center}
% \caption{Qualitative comparison with and without LGI, illustrating that LGI enables geometry-aligned shadows and relighting effects.}
% \label{fig:compare_LGI}
% \end{figure}

% Additionally, Fig.\ref{fig:compare_LGI} presents a visual comparison with and without the LGI module. The results demonstrate that LGI enhances the sensitivity to scene geometry, enabling more realistic and geometry-aligned shadow generation.
%
Our model requires \(4.82\,\mathrm{GB}\) of parameter memory and \(3.12\,\mathrm{TFLOPs}\), increasing these costs over the baseline by only \($0.0004\%$\) and \($0.0011\%$\), respectively, underscoring its efficiency. 
The method further extends naturally to multiple objects and light sources, with additional details, visualizations, and an analysis of the impact of the sample-point count $N$ provided in the Appendix~\ref{sec:samplepoint}.
%We further study the effect of the sample-point count \(N\). Because its compute impact is negligible at our reporting precision, Tab.~\ref{tab:m_study} reports accuracy for \(N\in\{8,16,32\}\). While \(N{=}32\) provides only marginal gains, we adopt \(N{=}16\) by default.
%
% Our method naturally extends to multiple objects and light sources, with details and visualizations provided in the supplementary material.

% \begin{table*}[htbp]
% \centering
% \footnotesize
% \caption{Study on Sample Point Count}
% \renewcommand{\arraystretch}{1.1}
% \setlength{\tabcolsep}{5pt}
% \begin{tabular}{l cccc cc ccc}
% \toprule
% \textbf{N} & \multicolumn{4}{c}{\textbf{Overall}} & \multicolumn{3}{c}{\textbf{Shadow region}} & \multicolumn{2}{c}{\textbf{Object region}} \\
%  \cmidrule(lr){2-5} \cmidrule(lr){6-8} \cmidrule(lr){9-10} 
% & RMSE $\downarrow$ & SSIM $\uparrow$ & BER $\downarrow$ & IOU $\uparrow$ & RMSE $\downarrow$ & SSIM  $\uparrow$ & BER $\downarrow$ & RMSE $\downarrow$ & SSIM $\uparrow$  \\
% \midrule
% 8 &  0.0344 & 0.7188 & 0.0612 & 0.7954 & 0.0991 & 0.6125 & 0.1141 & 0.0283 & 0.6875 \\
% 16 &  0.0334 & 0.7227 & 0.0588 & 0.8096 &0.0898 & 0.6195 & 0.1103 & 0.0282 & 0.6875 \\
% 32 & 0.0334 & 0.7231 & 0.0586 & 0.8099 & 0.0869 & 0.6198 & 0.1096 &0.0283 & 0.6875 \\
% \bottomrule
% \end{tabular}
% \label{tab:m_study}
% \end{table*}
\section{Conclusion}
\label{sec:conclusion}
We presented a unified, physically inspired framework utilizing Light-Geometry Interaction maps for joint shadow generation and object-level relighting. By approximating scene geometry from predicted depth maps, our method generates coherent and physically plausible shadows and relighting effects without requiring full 3D reconstruction. Additionally, we introduced a high-quality synthetic dataset specifically designed for joint shadow generation and relighting tasks, featuring diverse object categories, multiple material types, and challenging scenarios involving object-environment interreflection. Extensive experiments confirm that our approach achieves SOTA performance, effectively balancing visual consistency, physical accuracy, efficiency, and generalizability.

\section*{Acknowledgments}
This work was conducted during an internship at Amazon.
Hongdong Li is co-employed by Australian National University (full-time, primary) and Amazon IML (as a part-time Amazon Scholar).  Hongdong Li is also partially supported by an ARC Discovery Grant DP220100800.

\bibliography{iclr2025_conference}

@book{pharr2023physically,
  title={Physically based rendering: From theory to implementation},
  author={Pharr, Matt and Jakob, Wenzel and Humphreys, Greg},
  year={2023},
  publisher={MIT Press}
}

@inproceedings{keller1997instant,
  title={Instant radiosity},
  author={Keller, Alexander},
  booktitle={Proceedings of the 24th annual conference on Computer graphics and interactive techniques},
  pages={49--56},
  year={1997}
}

@InProceedings{Liu_2024_CVPR,
    author    = {Liu, Qingyang and You, Junqi and Wang, Jianting and Tao, Xinhao and Zhang, Bo and Niu, Li},
    title     = {Shadow Generation for Composite Image Using Diffusion Model},
    booktitle = {Proceedings of the IEEE/CVF Conference on Computer Vision and Pattern Recognition (CVPR)},
    month     = {June},
    year      = {2024},
    pages     = {8121-8130}
}

@inproceedings{zhao2025shadow,
  title={Shadow Generation Using Diffusion Model with Geometry Prior},
  author={Zhao, Haonan and Liu, Qingyang and Tao, Xinhao and Niu, Li and Zhai, Guangtao},
  booktitle={Proceedings of the Computer Vision and Pattern Recognition Conference},
  pages={7603--7612},
  year={2025}
}

@article{tasar2024controllable,
  title={Controllable shadow generation with single-step diffusion models from synthetic data},
  author={Tasar, Onur and Chadebec, Cl{\'e}ment and Aubin, Benjamin},
  journal={arXiv preprint arXiv:2412.11972},
  year={2024}
}

@inproceedings{zhang2025scaling,
    title={Scaling In-the-Wild Training for Diffusion-based Illumination Harmonization and Editing by Imposing Consistent Light Transport},
    author={Lvmin Zhang and Anyi Rao and Maneesh Agrawala},
    booktitle={The Thirteenth International Conference on Learning Representations},
    year={2025},
    url={https://openreview.net/forum?id=u1cQYxRI1H}
}

@inproceedings{kim2024switchlight,
  title={Switchlight: Co-design of physics-driven architecture and pre-training framework for human portrait relighting},
  author={Kim, Hoon and Jang, Minje and Yoon, Wonjun and Lee, Jisoo and Na, Donghyun and Woo, Sanghyun},
  booktitle={Proceedings of the IEEE/CVF Conference on Computer Vision and Pattern Recognition},
  pages={25096--25106},
  year={2024}
}

@article{chadebec2025lbm,
  title={LBM: Latent Bridge Matching for Fast Image-to-Image Translation},
  author={Chadebec, Cl{\'e}ment and Tasar, Onur and Sreetharan, Sanjeev and Aubin, Benjamin},
  journal={arXiv preprint arXiv:2503.07535},
  year={2025}
}

@book{revuz2013continuous,
  title={Continuous martingales and Brownian motion},
  author={Revuz, Daniel and Yor, Marc},
  volume={293},
  year={2013},
  publisher={Springer Science \& Business Media}
}

@article{DBLP:journals/corr/abs-2005-05460,
  author       = {Majed El Helou and
                  Ruofan Zhou and
                  Johan Barthas and
                  Sabine S{\"{u}}sstrunk},
  title        = {{VIDIT:} Virtual Image Dataset for Illumination Transfer},
  journal      = {CoRR},
  volume       = {abs/2005.05460},
  year         = {2020},
  url          = {https://arxiv.org/abs/2005.05460},
  eprinttype    = {arXiv},
  eprint       = {2005.05460},
  bibsource    = {dblp computer science bibliography, https://dblp.org}
}

@inproceedings{sheng2023pixht,
  title={Pixht-lab: Pixel height based light effect generation for image compositing},
  author={Sheng, Yichen and Zhang, Jianming and Philip, Julien and Hold-Geoffroy, Yannick and Sun, Xin and Zhang, He and Ling, Lu and Benes, Bedrich},
  booktitle={Proceedings of the IEEE/CVF Conference on Computer Vision and Pattern Recognition},
  pages={16643--16653},
  year={2023}
}

@inproceedings{winter2024objectdrop,
  title={Objectdrop: Bootstrapping counterfactuals for photorealistic object removal and insertion},
  author={Winter, Daniel and Cohen, Matan and Fruchter, Shlomi and Pritch, Yael and Rav-Acha, Alex and Hoshen, Yedid},
  booktitle={European Conference on Computer Vision},
  pages={112--129},
  year={2024},
  organization={Springer}
}

@article{christensen2018renderman,
  title={Renderman: An advanced path-tracing architecture for movie rendering},
  author={Christensen, Per and Fong, Julian and Shade, Jonathan and Wooten, Wayne and Schubert, Brenden and Kensler, Andrew and Friedman, Stephen and Kilpatrick, Charlie and Ramshaw, Cliff and Bannister, Marc and others},
  journal={ACM Transactions on Graphics (TOG)},
  volume={37},
  number={3},
  pages={1--21},
  year={2018},
  publisher={ACM New York, NY, USA}
}

@inproceedings{lafortune1996rendering,
  title={Rendering participating media with bidirectional path tracing},
  author={Lafortune, Eric P and Willems, Yves D},
  booktitle={Eurographics Workshop on Rendering Techniques},
  pages={91--100},
  year={1996},
  organization={Springer}
}

@inproceedings{wald2001interactive,
  title={Interactive rendering with coherent ray tracing},
  author={Wald, Ingo and Slusallek, Philipp and Benthin, Carsten and Wagner, Markus},
  booktitle={Computer graphics forum},
  volume={20},
  number={3},
  pages={153--165},
  year={2001},
  organization={Wiley Online Library}
}

@incollection{purcell2005ray,
  title={Ray tracing on programmable graphics hardware},
  author={Purcell, Timothy J and Buck, Ian and Mark, William R and Hanrahan, Pat},
  booktitle={ACM SIGGRAPH 2005 Courses},
  pages={268--es},
  year={2005}
}

@inproceedings{hong2022shadow,
  title={Shadow generation for composite image in real-world scenes},
  author={Hong, Yan and Niu, Li and Zhang, Jianfu},
  booktitle={Proceedings of the AAAI conference on artificial intelligence},
  volume={36},
  number={1},
  pages={914--922},
  year={2022}
}

@inproceedings{liu2020arshadowgan,
  title={Arshadowgan: Shadow generative adversarial network for augmented reality in single light scenes},
  author={Liu, Daquan and Long, Chengjiang and Zhang, Hongpan and Yu, Hanning and Dong, Xinzhi and Xiao, Chunxia},
  booktitle={Proceedings of the IEEE/CVF conference on computer vision and pattern recognition},
  pages={8139--8148},
  year={2020}
}

@article{zhang2019shadowgan,
  title={Shadowgan: Shadow synthesis for virtual objects with conditional adversarial networks},
  author={Zhang, Shuyang and Liang, Runze and Wang, Miao},
  journal={Computational Visual Media},
  volume={5},
  number={1},
  pages={105--115},
  year={2019},
  publisher={Springer}
}

@inproceedings{sheng2021ssn,
  title={Ssn: Soft shadow network for image compositing},
  author={Sheng, Yichen and Zhang, Jianming and Benes, Bedrich},
  booktitle={Proceedings of the IEEE/CVF Conference on Computer Vision and Pattern Recognition},
  pages={4380--4390},
  year={2021}
}

@inproceedings{sheng2022controllable,
  title={Controllable shadow generation using pixel height maps},
  author={Sheng, Yichen and Liu, Yifan and Zhang, Jianming and Yin, Wei and Oztireli, A Cengiz and Zhang, He and Lin, Zhe and Shechtman, Eli and Benes, Bedrich},
  booktitle={European Conference on Computer Vision},
  pages={240--256},
  year={2022},
  organization={Springer}
}

@inproceedings{bhattad2024stylitgan,
  title={Stylitgan: Image-based relighting via latent control},
  author={Bhattad, Anand and Soole, James and Forsyth, David A},
  booktitle={Proceedings of the IEEE/CVF Conference on Computer Vision and Pattern Recognition},
  pages={4231--4240},
  year={2024}
}

@article{wang2020single,
  title={Single image portrait relighting via explicit multiple reflectance channel modeling},
  author={Wang, Zhibo and Yu, Xin and Lu, Ming and Wang, Quan and Qian, Chen and Xu, Feng},
  journal={ACM Transactions on Graphics (ToG)},
  volume={39},
  number={6},
  pages={1--13},
  year={2020},
  publisher={ACM New York, NY, USA}
}

@article{wenger2005performance,
  title={Performance relighting and reflectance transformation with time-multiplexed illumination},
  author={Wenger, Andreas and Gardner, Andrew and Tchou, Chris and Unger, Jonas and Hawkins, Tim and Debevec, Paul},
  journal={ACM Transactions on Graphics (TOG)},
  volume={24},
  number={3},
  pages={756--764},
  year={2005},
  publisher={ACM New York, NY, USA}
}

@article{xing2024retinex,
  title={Retinex-diffusion: On controlling illumination conditions in diffusion models via retinex theory},
  author={Xing, Xiaoyan and Hu, Vincent Tao and Metzen, Jan Hendrik and Groh, Konrad and Karaoglu, Sezer and Gevers, Theo},
  journal={arXiv preprint arXiv:2407.20785},
  year={2024}
}

@inproceedings{liang2025diffusion,
  title={Diffusion Renderer: Neural Inverse and Forward Rendering with Video Diffusion Models},
  author={Liang, Ruofan and Gojcic, Zan and Ling, Huan and Munkberg, Jacob and Hasselgren, Jon and Lin, Chih-Hao and Gao, Jun and Keller, Alexander and Vijaykumar, Nandita and Fidler, Sanja and others},
  booktitle={Proceedings of the Computer Vision and Pattern Recognition Conference},
  pages={26069--26080},
  year={2025}
}

@inproceedings{zeng2024rgb,
author = {Zeng, Zheng and Deschaintre, Valentin and Georgiev, Iliyan and Hold-Geoffroy, Yannick and Hu, Yiwei and Luan, Fujun and Yan, Ling-Qi and Ha\v{s}an, Milo\v{s}},
title = {RGB↔X: Image decomposition and synthesis using material- and lighting-aware diffusion models},
year = {2024},
isbn = {9798400705250},
publisher = {Association for Computing Machinery},
address = {New York, NY, USA},
url = {https://doi.org/10.1145/3641519.3657445},
doi = {10.1145/3641519.3657445},
booktitle = {ACM SIGGRAPH 2024 Conference Papers},
articleno = {75},
numpages = {11},
location = {Denver, CO, USA},
series = {SIGGRAPH '24}
}

@inproceedings{zhang2025zerocomp,
  title={Zerocomp: Zero-shot object compositing from image intrinsics via diffusion},
  author={Zhang, Zitian and Fortier-Chouinard, Fr{\'e}d{\'e}ric and Garon, Mathieu and Bhattad, Anand and Lalonde, Jean-Fran{\c{c}}ois},
  booktitle={2025 IEEE/CVF Winter Conference on Applications of Computer Vision (WACV)},
  pages={483--494},
  year={2025},
  organization={IEEE}
}

@article{fortier2024spotlight,
  title={Spotlight: Shadow-guided object relighting via diffusion},
  author={Fortier-Chouinard, Fr{\'e}d{\'e}ric and Zhang, Zitian and Messier, Louis-Etienne and Garon, Mathieu and Bhattad, Anand and Lalonde, Jean-Fran{\c{c}}ois},
  journal={arXiv preprint arXiv:2411.18665},
  year={2024}
}

@inproceedings{griffiths2022outcast,
  title={OutCast: Outdoor Single-image Relighting with Cast Shadows},
  author={Griffiths, David and Ritschel, Tobias and Philip, Julien},
  booktitle={Computer Graphics Forum},
  volume={41},
  number={2},
  pages={179--193},
  year={2022},
  organization={Wiley Online Library}
}

@inproceedings{kocsis2024lightit,
  title={Lightit: Illumination modeling and control for diffusion models},
  author={Kocsis, Peter and Philip, Julien and Sunkavalli, Kalyan and Nie{\ss}ner, Matthias and Hold-Geoffroy, Yannick},
  booktitle={Proceedings of the IEEE/CVF Conference on Computer Vision and Pattern Recognition},
  pages={9359--9369},
  year={2024}
}

@article{lin2023urbanir,
  title={Urbanir: Large-scale urban scene inverse rendering from a single video},
  author={Lin, Zhi-Hao and Liu, Bohan and Chen, Yi-Ting and Chen, Kuan-Sheng and Forsyth, David and Huang, Jia-Bin and Bhattad, Anand and Wang, Shenlong},
  journal={arXiv preprint arXiv:2306.09349},
  year={2023}
}

@article{zhao2024illuminerf,
  title={Illuminerf: 3d relighting without inverse rendering},
  author={Zhao, Xiaoming and Srinivasan, Pratul and Verbin, Dor and Park, Keunhong and Martin Brualla, Ricardo and Henzler, Philipp},
  journal={Advances in Neural Information Processing Systems},
  volume={37},
  pages={42593--42617},
  year={2024}
}

@inproceedings{rombach2022high,
  title={High-resolution image synthesis with latent diffusion models},
  author={Rombach, Robin and Blattmann, Andreas and Lorenz, Dominik and Esser, Patrick and Ommer, Bj{\"o}rn},
  booktitle={Proceedings of the IEEE/CVF conference on computer vision and pattern recognition},
  pages={10684--10695},
  year={2022}
}

@inproceedings{hu2019mask,
  title={Mask-shadowgan: Learning to remove shadows from unpaired data},
  author={Hu, Xiaowei and Jiang, Yitong and Fu, Chi-Wing and Heng, Pheng-Ann},
  booktitle={Proceedings of the IEEE/CVF international conference on computer vision},
  pages={2472--2481},
  year={2019}
}

@inproceedings{tao2024shadow,
  title={Shadow generation with decomposed mask prediction and attentive shadow filling},
  author={Tao, Xinhao and Cao, Junyan and Hong, Yan and Niu, Li},
  booktitle={Proceedings of the AAAI Conference on Artificial Intelligence},
  volume={38},
  number={6},
  pages={5198--5206},
  year={2024}
}

@article{liu2023desobav2,
  title={Desobav2: Towards large-scale real-world dataset for shadow generation},
  author={Liu, Qingyang and Wang, Jianting and Niu, Li},
  journal={arXiv preprint arXiv:2308.09972},
  year={2023}
}

@inproceedings{liu2024shadow,
  title={Shadow generation for composite image using diffusion model},
  author={Liu, Qingyang and You, Junqi and Wang, Jianting and Tao, Xinhao and Zhang, Bo and Niu, Li},
  booktitle={Proceedings of the IEEE/CVF Conference on Computer Vision and Pattern Recognition},
  pages={8121--8130},
  year={2024}
}

@software{colour_science_046,
  author       = {colour-science package},
  title        = {{Colour Science for Python}},
  url          = {https://pypi.org/project/colour-science/}
}

@inproceedings{burley2012_principled_bsdf,
  author    = {Brent Burley},
  title     = {{Physically Based Shading at Disney}},
  booktitle = {{ACM SIGGRAPH 2012 Course Notes}},
  year      = {2012},
  address   = {Los Angeles, CA},
  publisher = {{ACM}}
}

@software{blender_cycles_44,
author={Cycles},
title={{Blender Cycles Render Engine}},
url={https://github.com/blender/cycles},
}

@article{loshchilov2017decoupled,
  title={Decoupled weight decay regularization},
  author={Loshchilov, Ilya and Hutter, Frank},
  journal={arXiv preprint arXiv:1711.05101},
  year={2017}
}

@article{yang2024depth,
  title={Depth anything v2},
  author={Yang, Lihe and Kang, Bingyi and Huang, Zilong and Zhao, Zhen and Xu, Xiaogang and Feng, Jiashi and Zhao, Hengshuang},
  journal={Advances in Neural Information Processing Systems},
  volume={37},
  pages={21875--21911},
  year={2024}
}

@inproceedings{zhang2018lpips,
  title={The unreasonable effectiveness of deep features as a perceptual metric},
  author={Zhang, Richard and Isola, Phillip and Efros, Alexei A and Shechtman, Eli and Wang, Oliver},
  booktitle={Proceedings of the IEEE conference on computer vision and pattern recognition},
  pages={586--595},
  year={2018}
}

@incollection{debevec2008rendering,
  title={Rendering synthetic objects into real scenes: Bridging traditional and image-based graphics with global illumination and high dynamic range photography},
  author={Debevec, Paul},
  booktitle={Acm siggraph 2008 classes},
  pages={1--10},
  year={2008}
}

@inproceedings{ward1994radiance,
  title={The RADIANCE lighting simulation and rendering system},
  author={Ward, Gregory J},
  booktitle={Proceedings of the 21st annual conference on Computer graphics and interactive techniques},
  pages={459--472},
  year={1994}
}

@article{podell2023sdxl,
  title={Sdxl: Improving latent diffusion models for high-resolution image synthesis},
  author={Podell, Dustin and English, Zion and Lacey, Kyle and Blattmann, Andreas and Dockhorn, Tim and M{\"u}ller, Jonas and Penna, Joe and Rombach, Robin},
  journal={arXiv preprint arXiv:2307.01952},
  year={2023}
}

@inproceedings{zeng2024dilightnet,
  title={DiLightNet: Fine-grained lighting control for diffusion-based image generation},
  author={Zeng, Chong and Dong, Yue and Peers, Pieter and Kong, Youkang and Wu, Hongzhi and Tong, Xin},
  booktitle={ACM SIGGRAPH 2024 Conference Papers},
  pages={1--12},
  year={2024}
}

@inproceedings{zhu2022learning,
  title={Learning-based inverse rendering of complex indoor scenes with differentiable monte carlo raytracing},
  author={Zhu, Jingsen and Luan, Fujun and Huo, Yuchi and Lin, Zihao and Zhong, Zhihua and Xi, Dianbing and Wang, Rui and Bao, Hujun and Zheng, Jiaxiang and Tang, Rui},
  booktitle={SIGGRAPH Asia 2022 Conference Papers},
  pages={1--8},
  year={2022}
}
\bibliographystyle{iclr2025_conference}

\newpage
\appendix
\section{Extend to shadow-centered Image Harmonization}
\begin{figure}[htbp]
\centering
\includegraphics[width=0.95\linewidth]{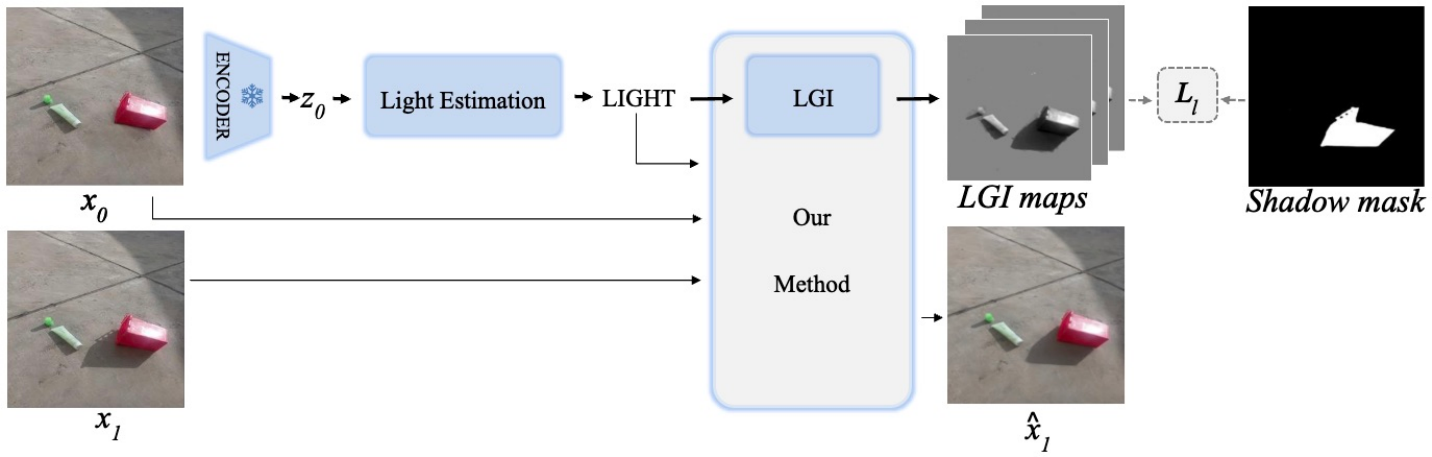}
\caption{ Overview of the image harmonization pipeline. The light estimation network predicts a full set of lighting parameters, with the estimated lighting direction used to compute LGI maps supervised by ground-truth shadow masks. These estimates are integrated into our light-aware shadow generation and relighting framework to produce consistent and realistic harmonization of the inserted object. Since the LGI module is fully differentiable, it enables end-to-end training of the light estimation network.
}
\label{fig:harmonization_overview}
\end{figure}

In this section, we extend our method to realistic image harmonization, where a foreground object is composited onto a new background. Most harmonization datasets do not provide ground-truth lighting information; however, some datasets include shadow masks. In our setting, the model infers lighting conditions directly from the composited image. As shown in Fig.~\ref{fig:harmonization_overview}, the light estimation network predicts a complete set of lighting parameters, including color, radius, distance, intensity, and direction. These parameters are integrated into our light-aware shadow generation and relighting framework to ensure consistent and realistic harmonization of the inserted object. Since LGI is fully differentiable, it enables end-to-end training of the light estimation network. Furthermore, when shadow mask is available, we formulate a loss function that leverages LGI to supervise the estimation of lighting direction.

The light estimation network consists of four convolutional layers with progressively increasing channel dimensions. The resulting feature representation is then processed by a two-layer multilayer perceptron (MLP), producing an 8-dimensional vector encoding the full set of lighting parameters. 
During training, we first freeze the light-aware shadow generation and relighting framework for 5 epochs to warm up the light estimation module, and then continue with end-to-end training where all parameters are updated.

\begin{figure}[htbp]
\centering
    \begin{subfigure}{0.19\textwidth}
        \includegraphics[width=\linewidth]{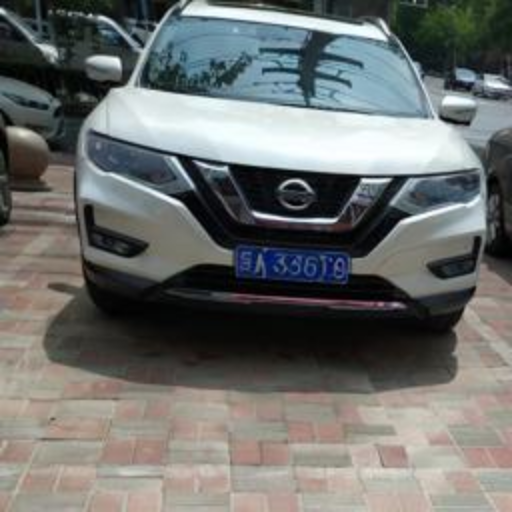}
        \caption{Shadowed Image}
    \end{subfigure}
    \begin{subfigure}{0.19\textwidth}
        \includegraphics[width=\linewidth]{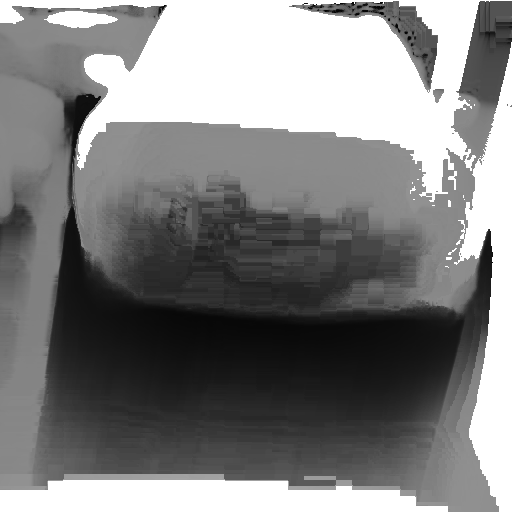}
        \caption{$\vc^m_1$}
        \label{fig:forward_light}
    \end{subfigure}
    \begin{subfigure}{0.19\textwidth}
        \includegraphics[width=\linewidth]{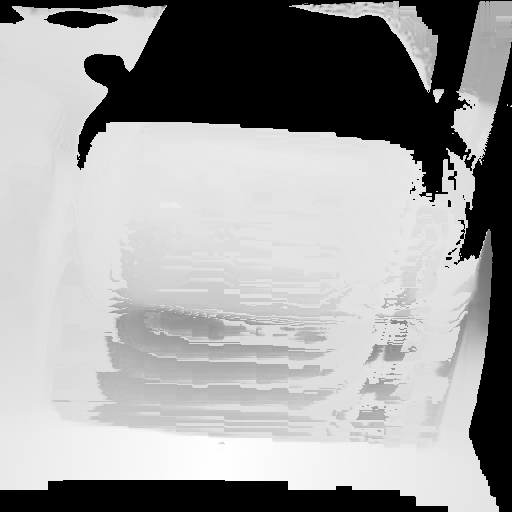}
        \caption{$\vc^m_2$}
    \end{subfigure}
    \begin{subfigure}{0.19\textwidth}
        \includegraphics[width=\linewidth]{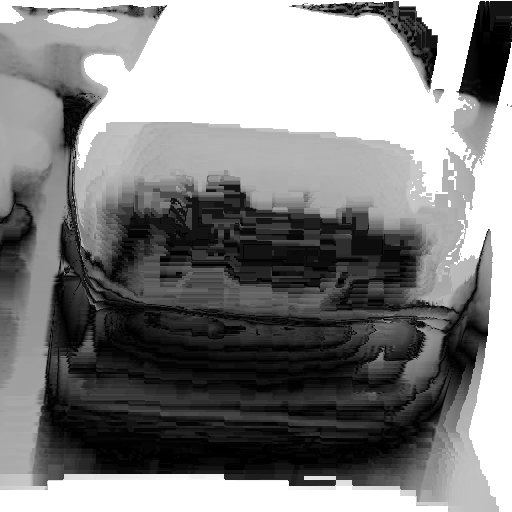}
        \caption{$\vc^m_3$}
    \end{subfigure}
    \begin{subfigure}{0.19\textwidth}
        \includegraphics[width=\linewidth]{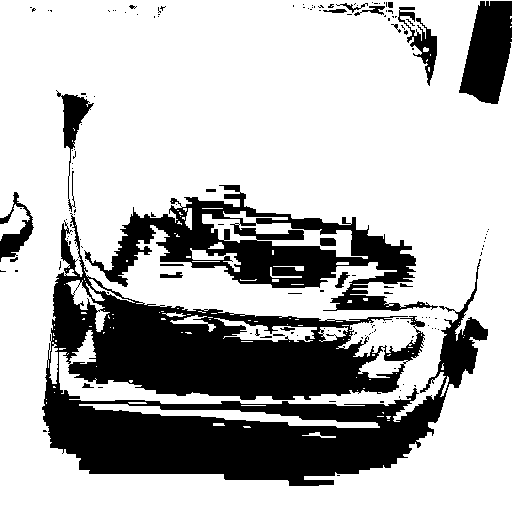}
        \caption{Hard Shadow}
    \end{subfigure}
\caption{
Example of sunlight LGI maps. They establish a correspondence between light direction and shadow shape when scene geometry is known.
}
\label{fig:desoba_embed}
\vspace{-0.4cm}
\end{figure}

For outdoor scenarios, sunlight serves as the primary light source and can be effectively modeled as a point light at an infinite distance, emitting parallel rays. To accommodate this scenario, we introduce a modified variant of our LGI maps specifically adapted for sunlight. Instead of casting rays toward a finite light position, rays are cast along the estimated light direction relative to our normalized camera coordinate system, where the x-axis and y-axis lie within the range $(-0.5, 0.5)$ and the z-axis within $(0, 1]$. According to this coordinate system in Eq.~\ref{eq:lift}, we assume a default intrinsic matrix \( K = \left[\begin{smallmatrix} W & 0 & W/2 \\ 0 & H & H/2 \\ 0 & 0 & 1 \end{smallmatrix}\right] \), where H, W denote the height and width of the input image. An visual example of LGI maps for sunlight is shown in Fig.~\ref{fig:desoba_embed}.

Building on the LGI maps, we obtain the predicted hard shadow mask $\mathcal{M}_p$.
We then define a light-direction estimation loss $\mathcal{L}_{l}$, applied when a ground-truth shadow mask $\mathcal{M}_g$ is available.
Since we only consider the inserted object’s shadow, the loss is computed within a region restricted to the dilated ground-truth shadow mask.
The overall loss combines BCE and IoU terms:
\begin{equation}
\mathcal{L}_{l}(\mathcal{M}_p, \mathcal{M}_g) = \mathcal{L}_{\text{BCE}}(\mathcal{M}_p, \mathcal{M}_g) + \mathcal{L}_{\text{IoU}}(\mathcal{M}_p, \mathcal{M}_g),
\end{equation}
where the BCE loss is defined as
\begin{equation}
\mathcal{L}_{\text{BCE}}(\mathcal{M}_p, \mathcal{M}_g) = - \mathbb{E}\Big[ \mathcal{M}_g \log \mathcal{M}_p + (1 - \mathcal{M}_g) \log (1 - \mathcal{M}_p) \Big],
\end{equation}
and the IoU loss is defined as
\begin{equation}
\mathcal{L}_{\text{IoU}}(\mathcal{M}_p, \mathcal{M}_g)
= 1 - \frac{\mathbb{E}\!\left[ \mathcal{M}_p \mathcal{M}_g \right]}
{\mathbb{E}\!\left[ \mathcal{M}_p + \mathcal{M}_g - \mathcal{M}_p \mathcal{M}_g \right]},
\end{equation}
with $\mathbb{E}$ denoting the expectation over pixels.

% \begin{figure}[!htbp]
% \begin{center}
% \includegraphics[width=\textwidth]{image/compare_SGDGP.pdf}
% \end{center}
% \caption{Qualitative comparison with SGDGP. Compared to their 2D bounding-box and shadow-template geometry prior, our method achieves superior alignment with object geometry.}
% \label{fig:compare_SGDGP}
% \end{figure}

% With the above supervision, we first pre-train the light estimation network for 5 epochs, followed by end-to-end joint training of all parameters.

% We evaluate our method on the DESOBAv2~\cite{liu2023desobav2} dataset. Results in Tab.\ref{tab:DESOBAv2_comparison} show that our method achieves performance comparable to the top method, while obtaining the best results specifically in shadow regions. Additionally, we present a qualitative comparison in Fig.~\ref{fig:compare_SGDGP}. 
% Our approach achieves better alignment with object geometry compared to SGDGP~\cite{zhao2025shadow}, which relies on a 2D shadow-template as its geometry prior. In contrast, our method leverages predicted depth maps as a geometry prior, enabling more precise geometry-aware shadow generation.

\section{Multiple Objects}

\begin{figure}[htbp]
\centering
    \begin{subfigure}{0.32\textwidth}
        \includegraphics[width=\linewidth]{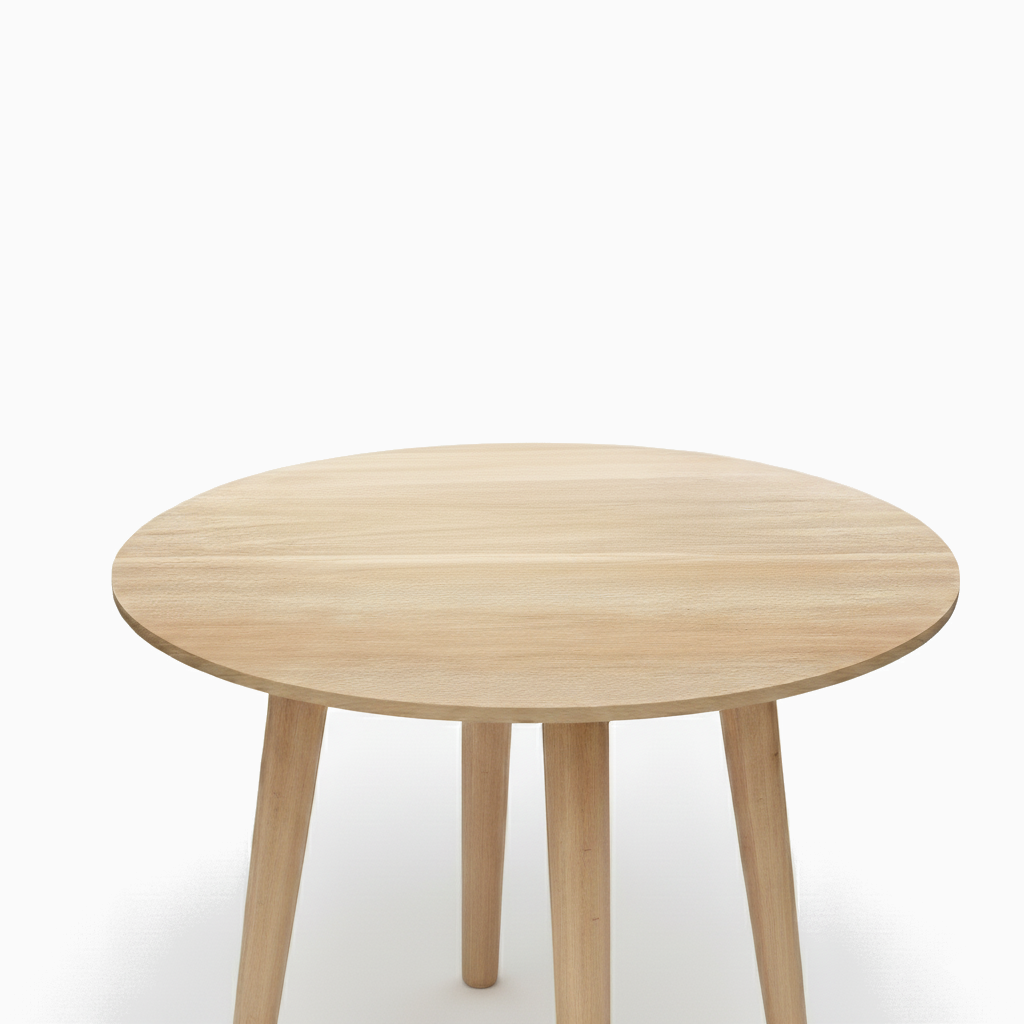}
        \caption{}
    \end{subfigure}
    \begin{subfigure}{0.32\textwidth}
        \includegraphics[width=\linewidth]{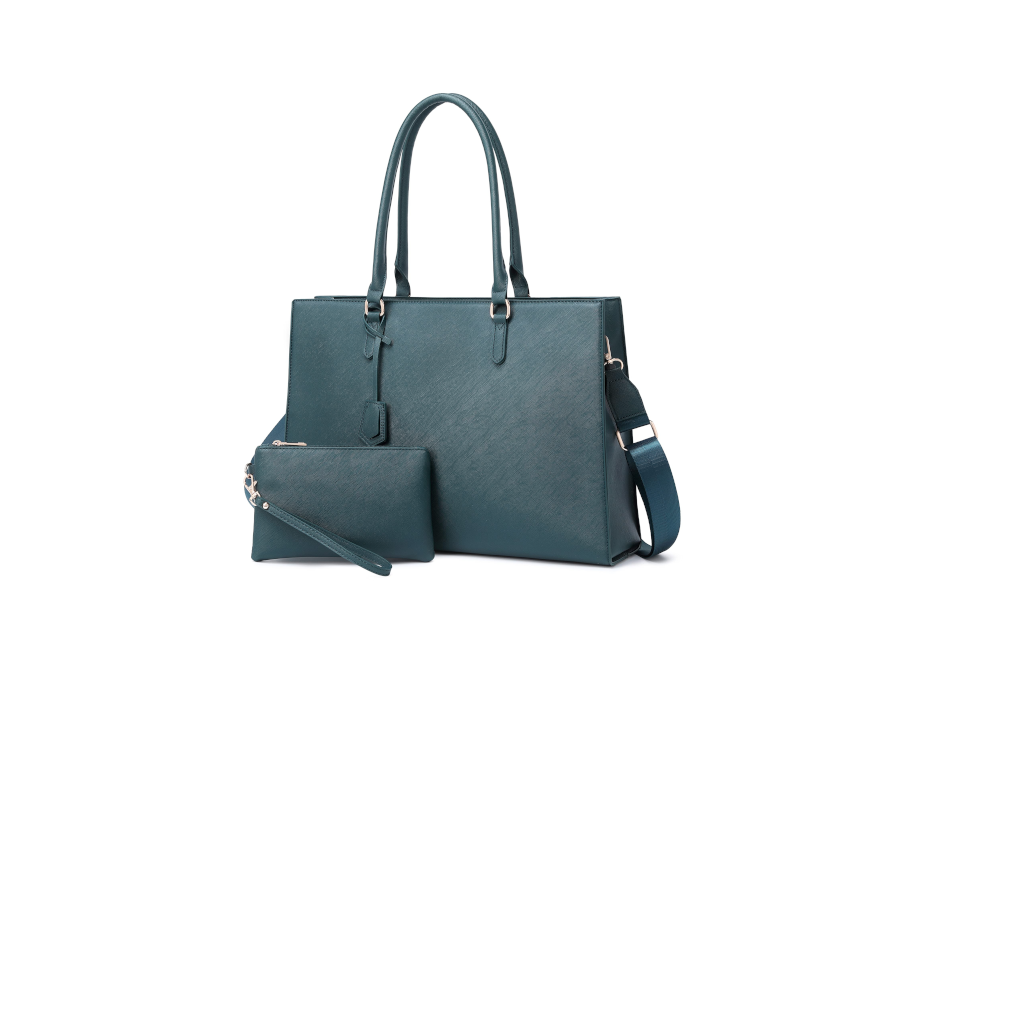}
        \caption{}
    \end{subfigure}
    \begin{subfigure}{0.32\textwidth}
        \includegraphics[width=\linewidth]{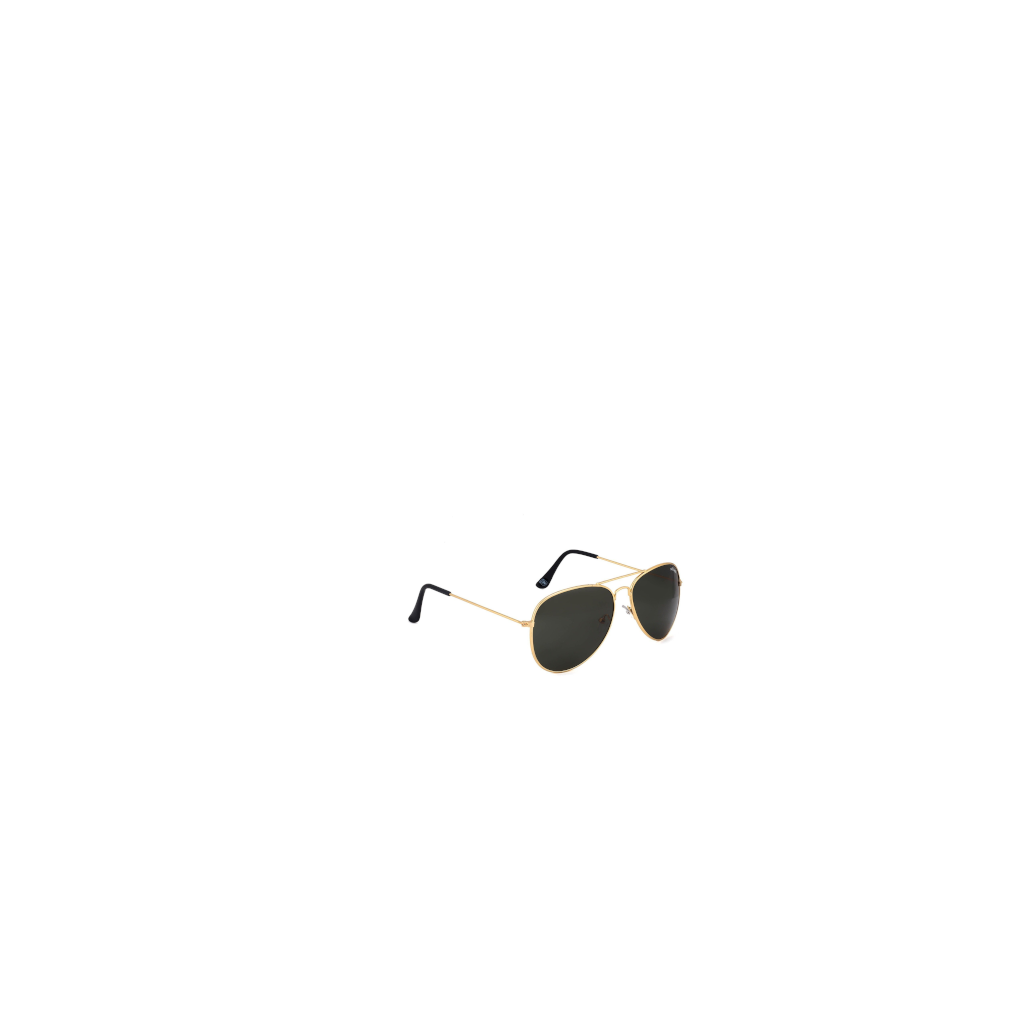}
        \caption{}
    \end{subfigure} \\
    \begin{subfigure}{0.32\textwidth}
        \includegraphics[width=\linewidth]{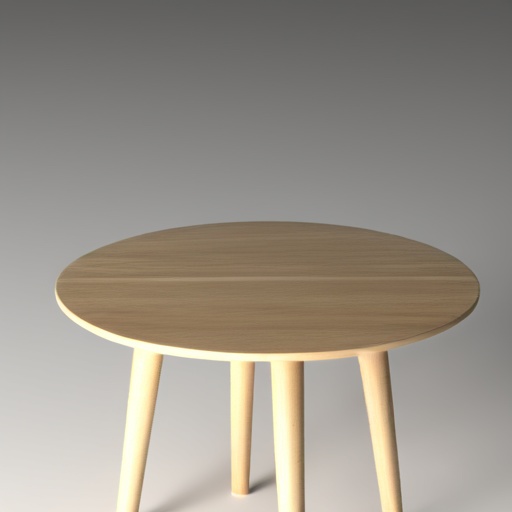}
        \caption{}
    \end{subfigure}
    \begin{subfigure}{0.32\textwidth}
        \includegraphics[width=\linewidth]{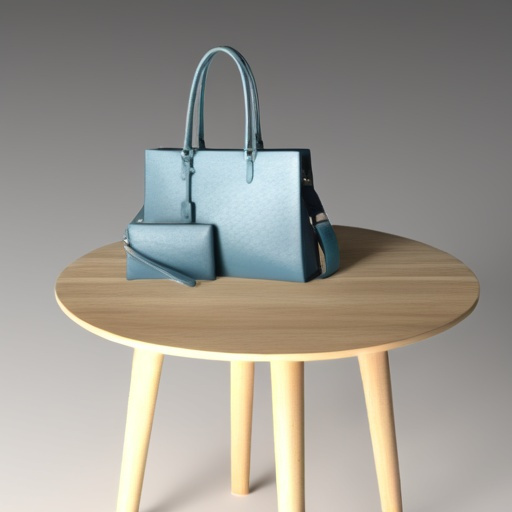}
        \caption{}
    \end{subfigure}
    \begin{subfigure}{0.32\textwidth}
        \includegraphics[width=\linewidth]{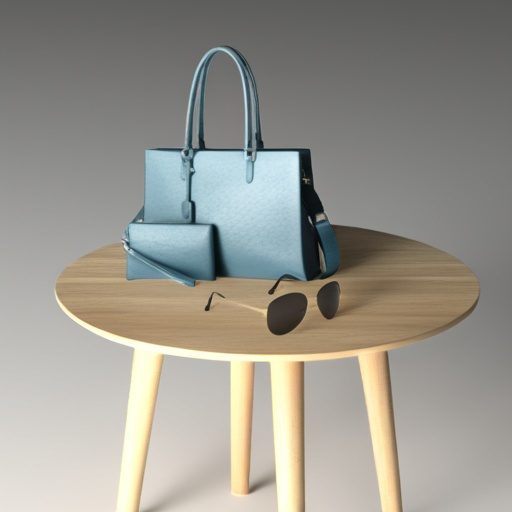}
        \caption{}
    \end{subfigure}
\caption{Examples of multiple object insertion. (a–c) Inserted objects. (d–f) Scenes with inserted objects after shadow generation and relighting. Our method produces realistic, texture-aware shadows on the table and preserves faithful relighting across wood, leather, metal, and glass materials.
}
\label{fig:mul_obj}
\end{figure}

Our method extends naturally to scene editing via object-level insertion. We further demonstrate multi-object insertion by applying the method sequentially, inserting objects one at a time. As shown in Fig.~\ref{fig:mul_obj}, our approach generalizes well to the multi-object setting: it casts realistic shadows onto previously inserted objects, ensures that the generated shadows align with the scene geometry, and maintains relighting consistent with the materials of each object.

\section{Multiple Light Sources}

\begin{figure}[htbp]
\centering
\resizebox{\textwidth}{!}{%
  \begin{tabular}{cccc}
    \includegraphics[width=0.26\textwidth]{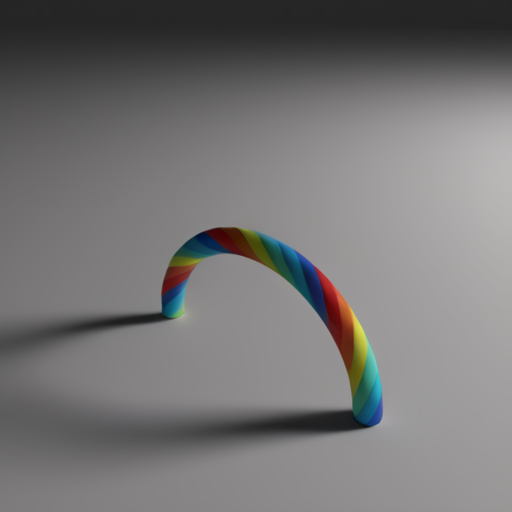}
    \includegraphics[width=0.195\textwidth]{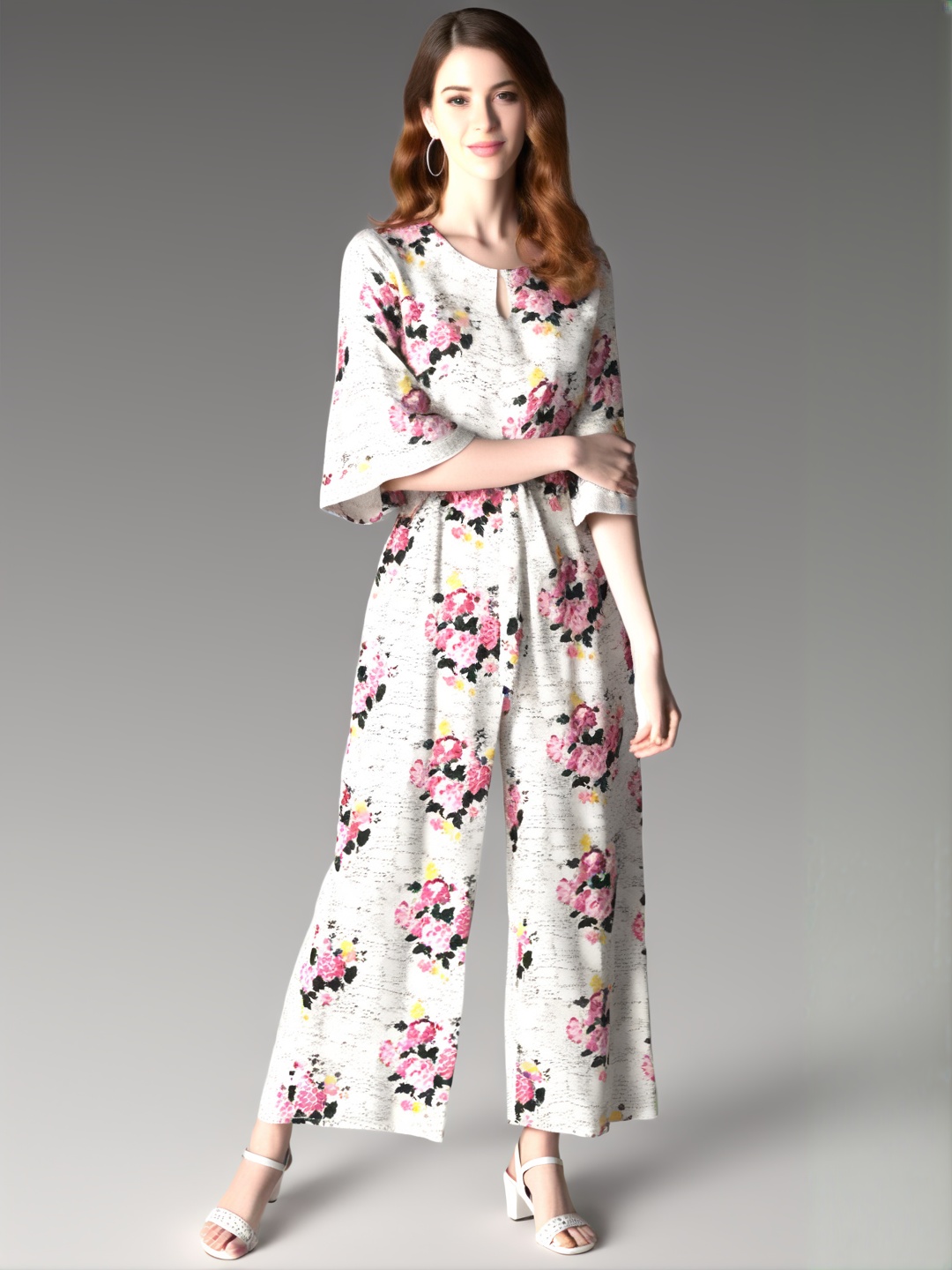}
    \includegraphics[width=0.26\textwidth]{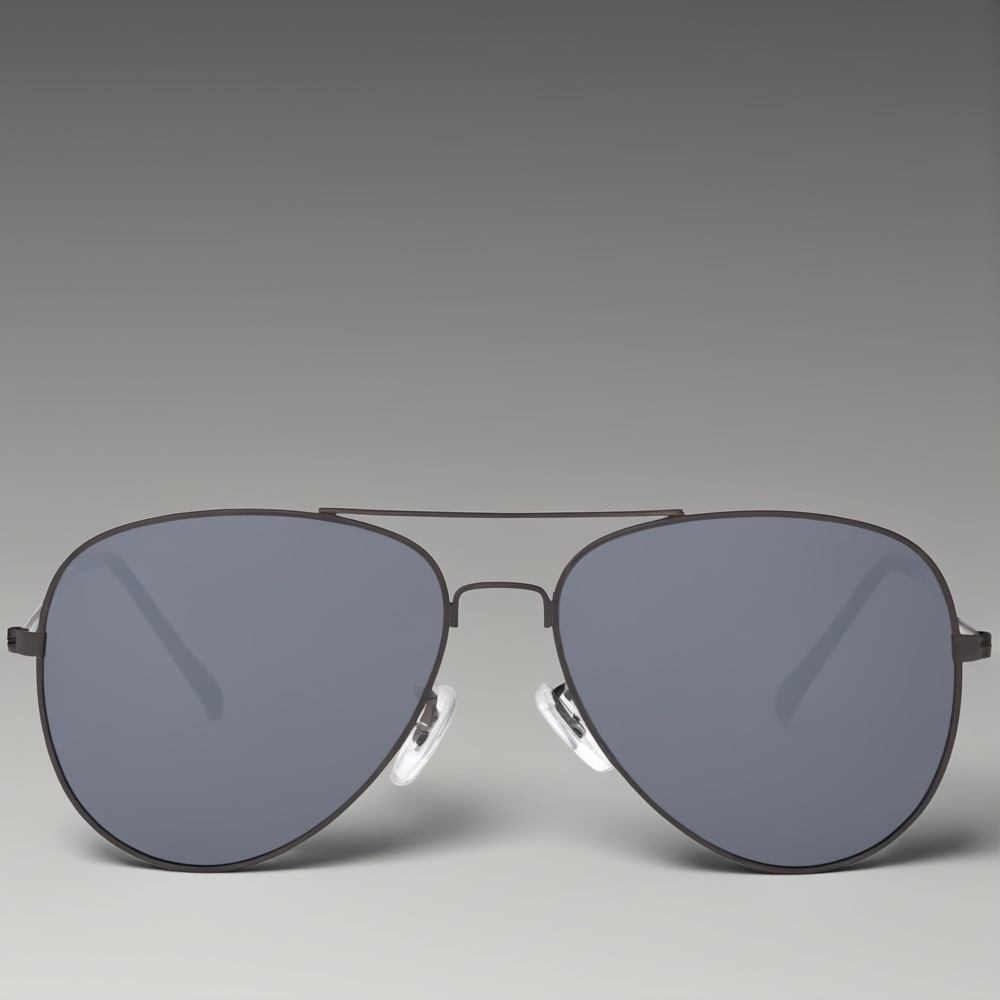}
    \includegraphics[width=0.26\textwidth]{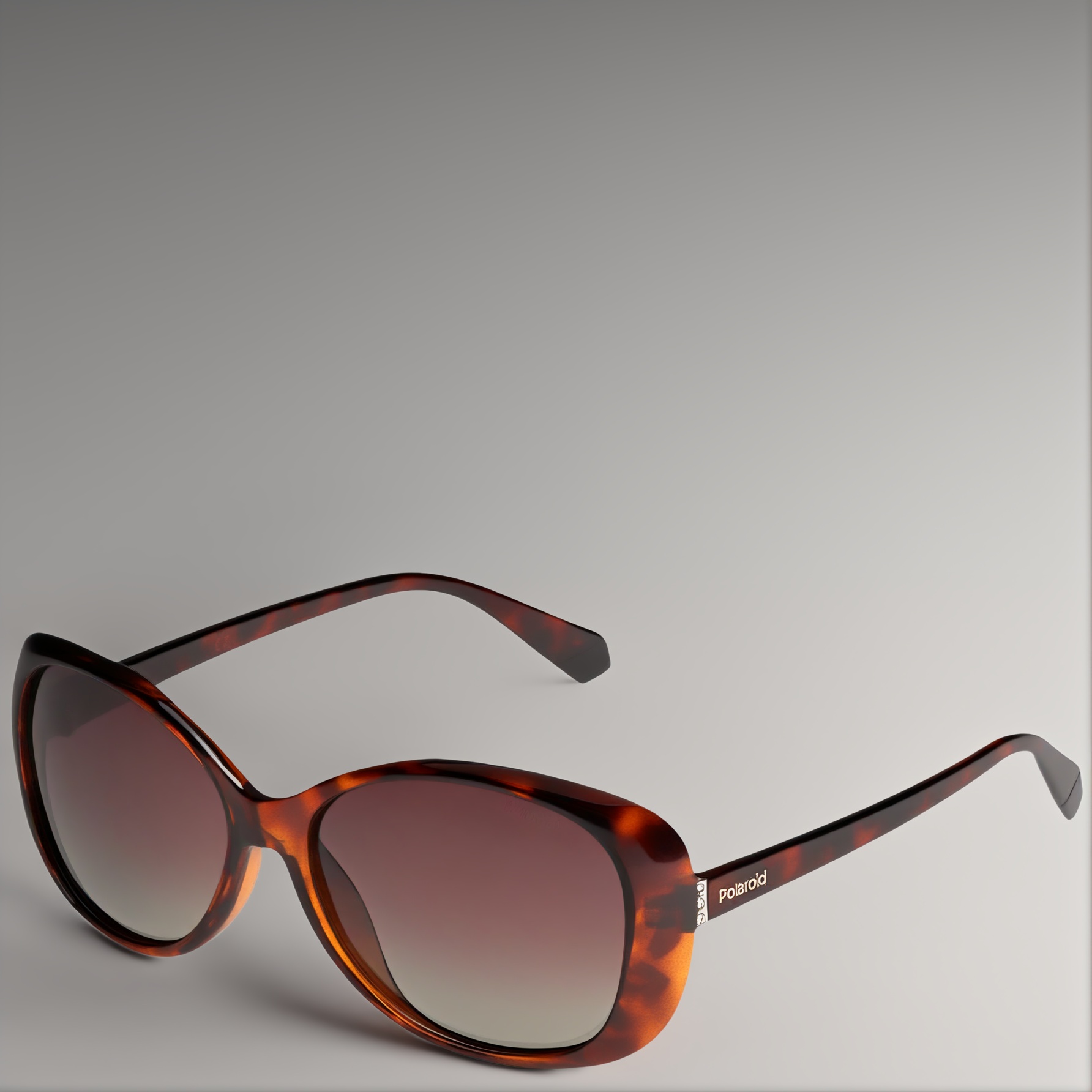} \\
    \includegraphics[width=0.26\textwidth]{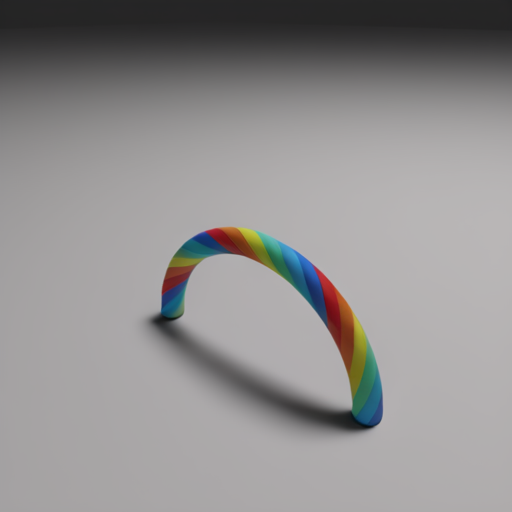}
    \includegraphics[width=0.195\textwidth]{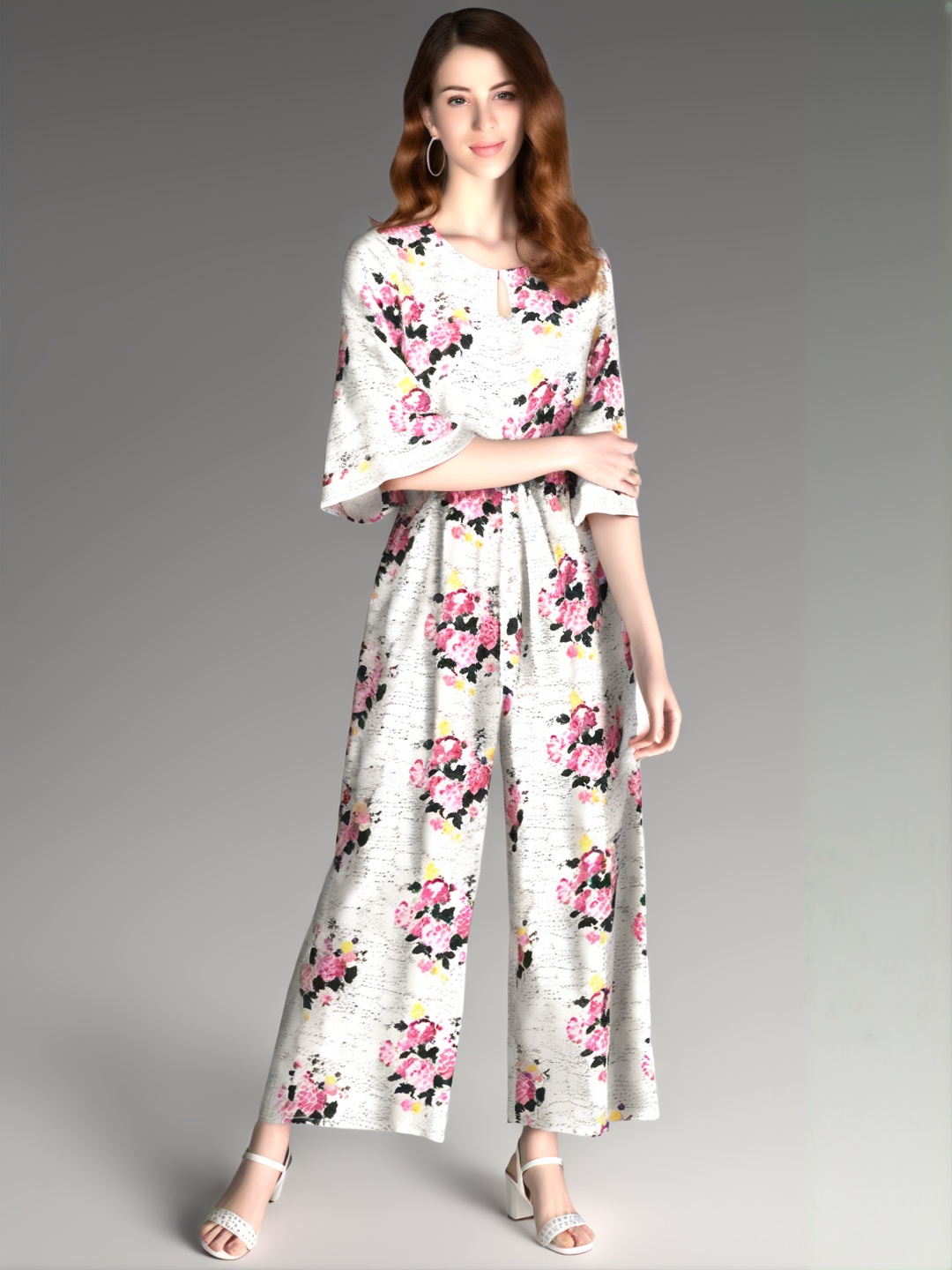}
    \includegraphics[width=0.26\textwidth]{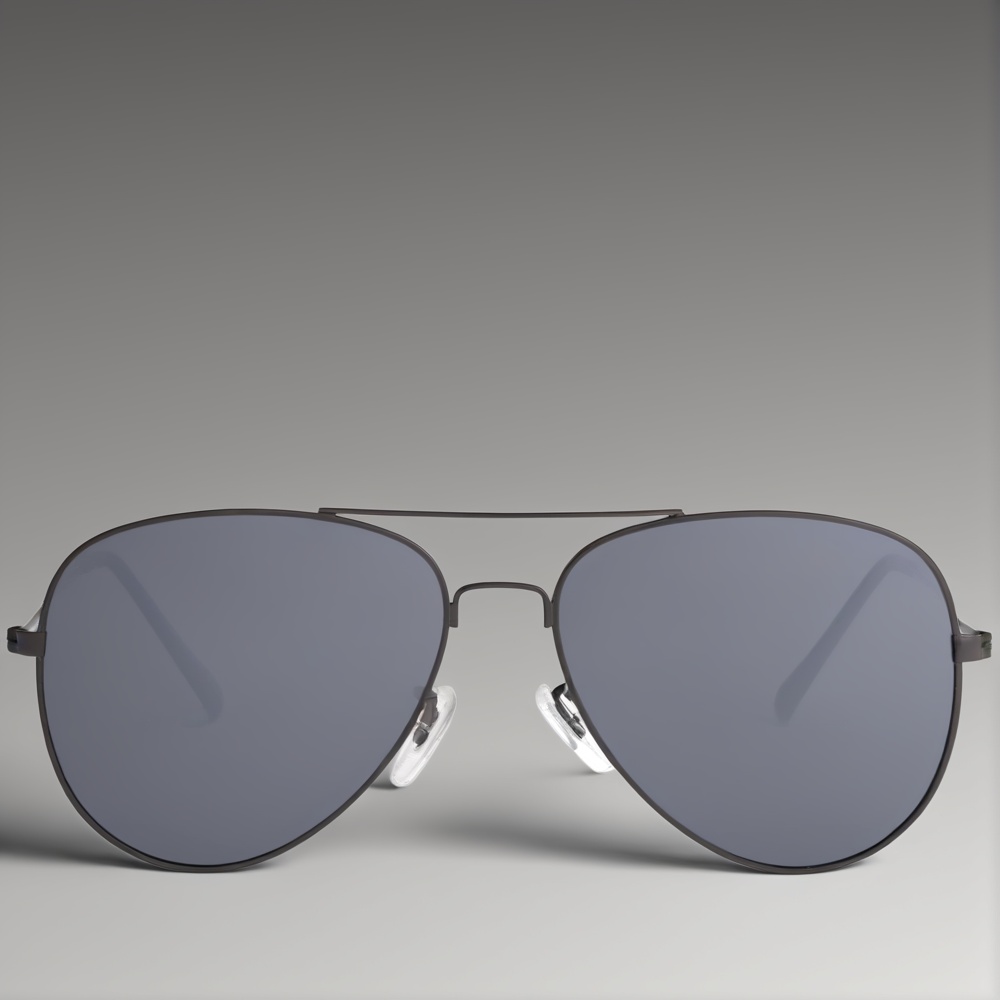}
    \includegraphics[width=0.26\textwidth]{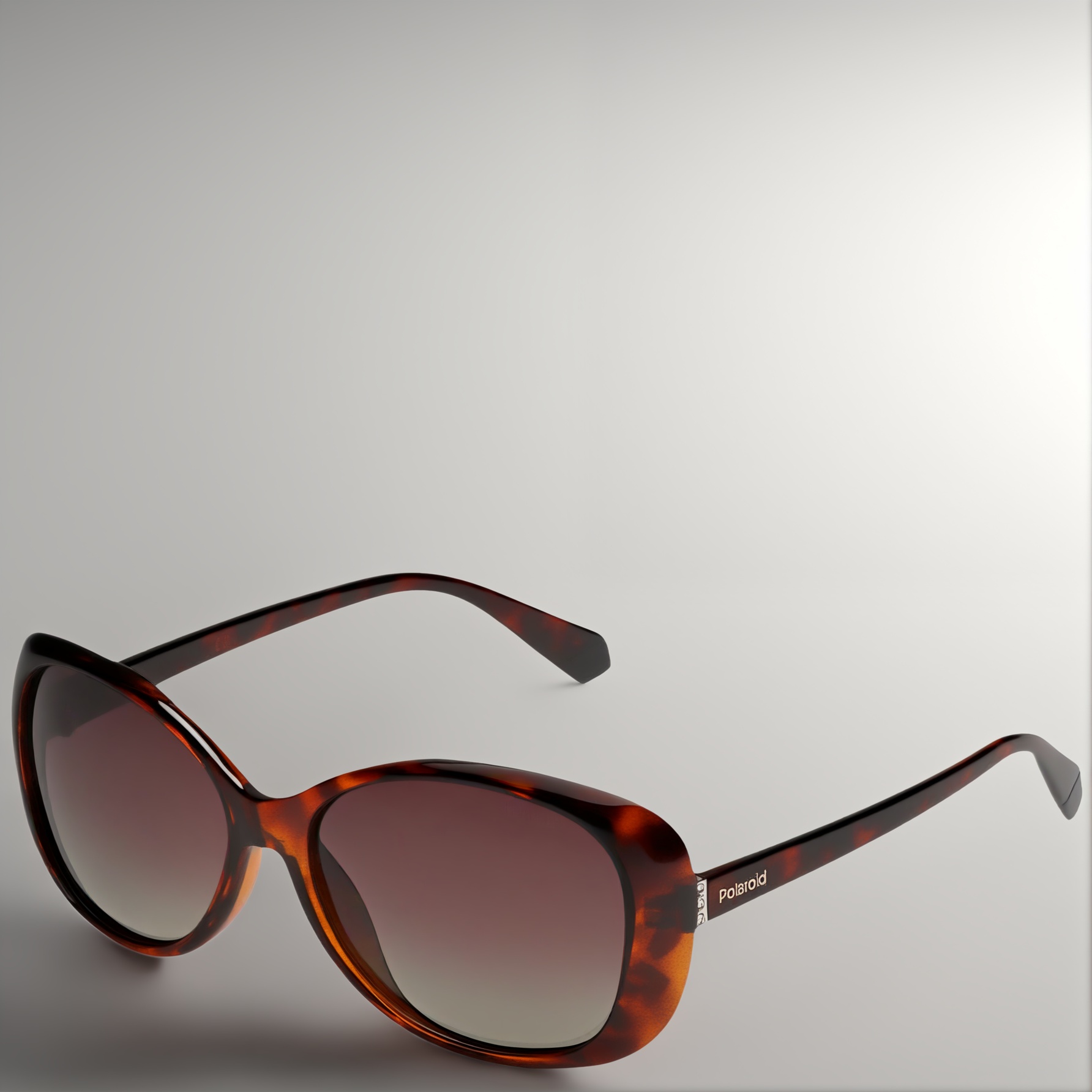} \\
    \includegraphics[width=0.26\textwidth]{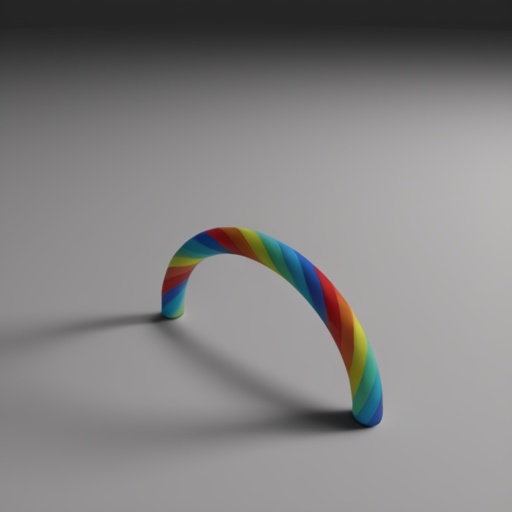}
    \includegraphics[width=0.195\textwidth]{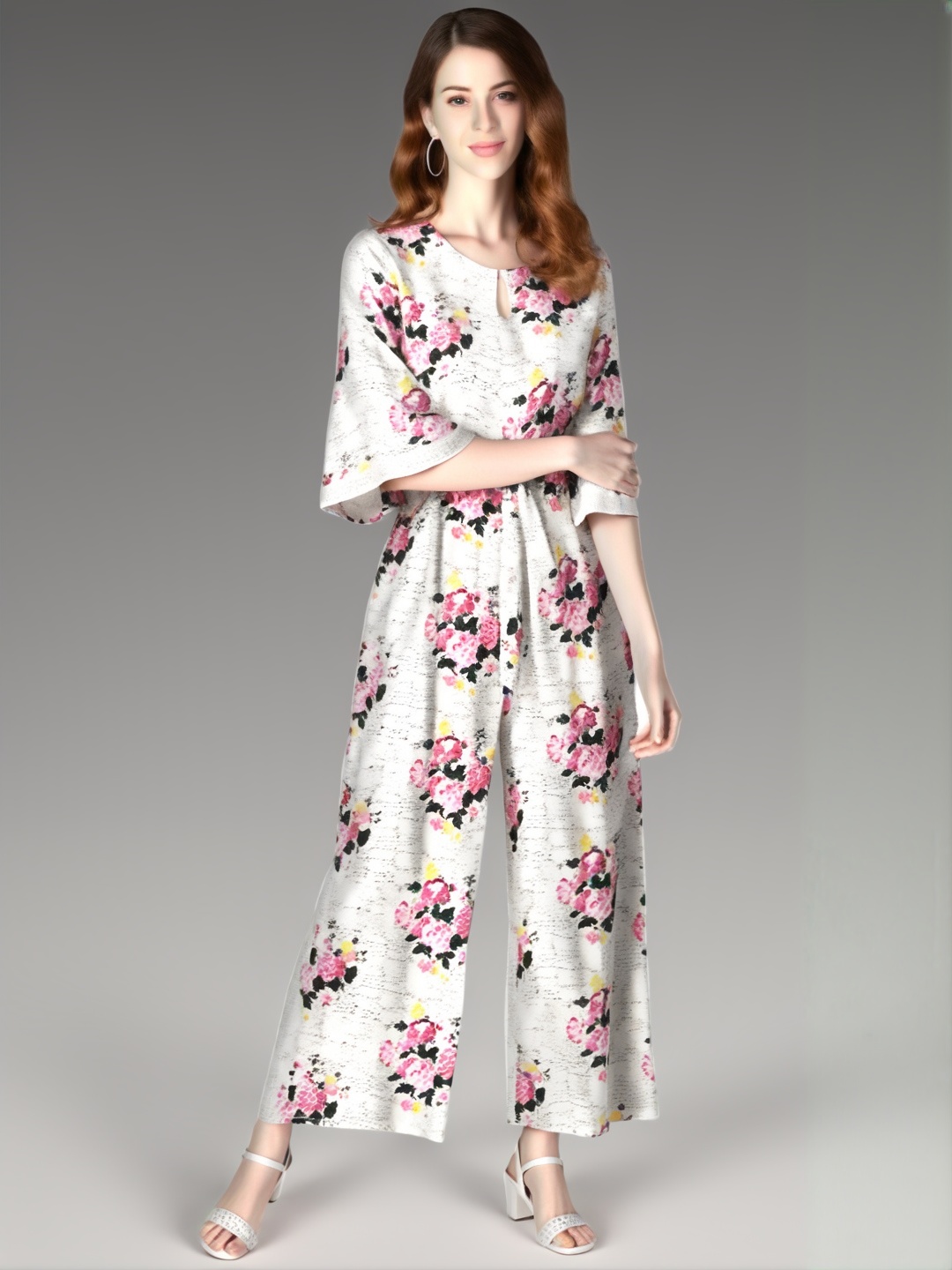}
    \includegraphics[width=0.26\textwidth]{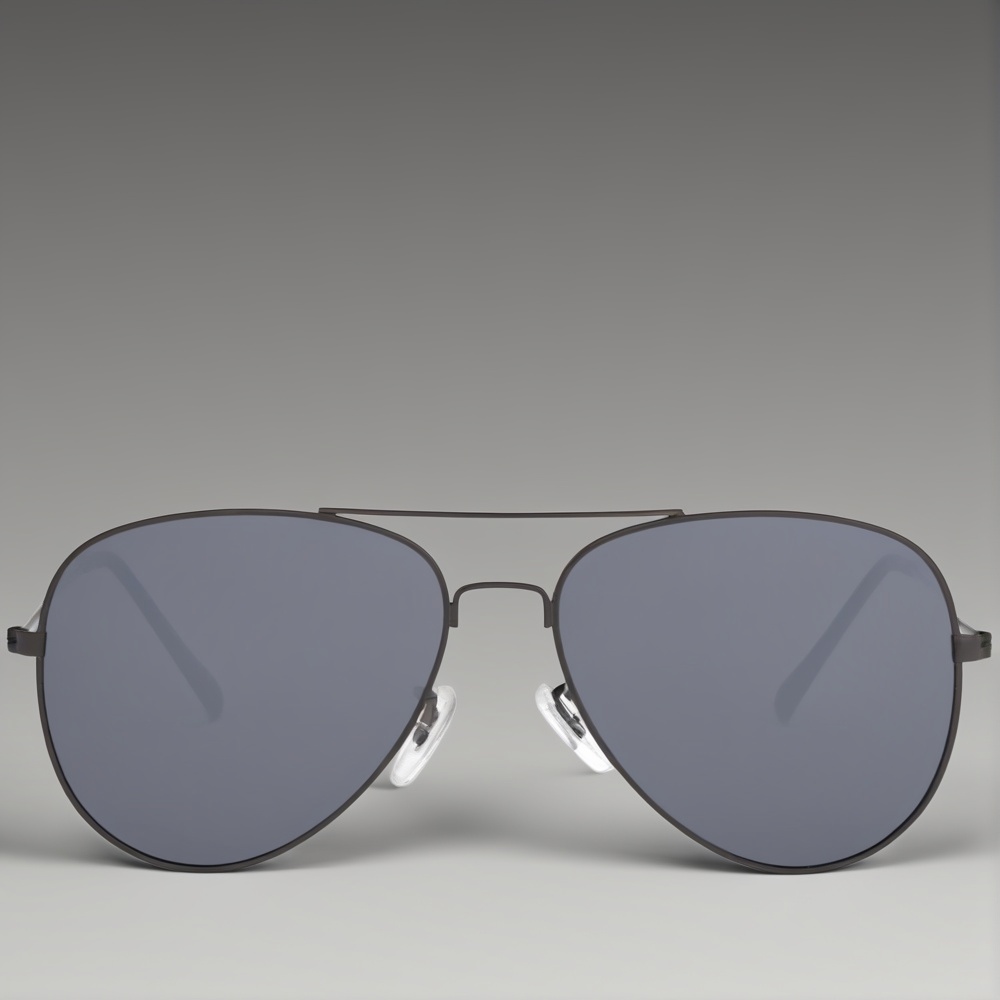}
    \includegraphics[width=0.26\textwidth]{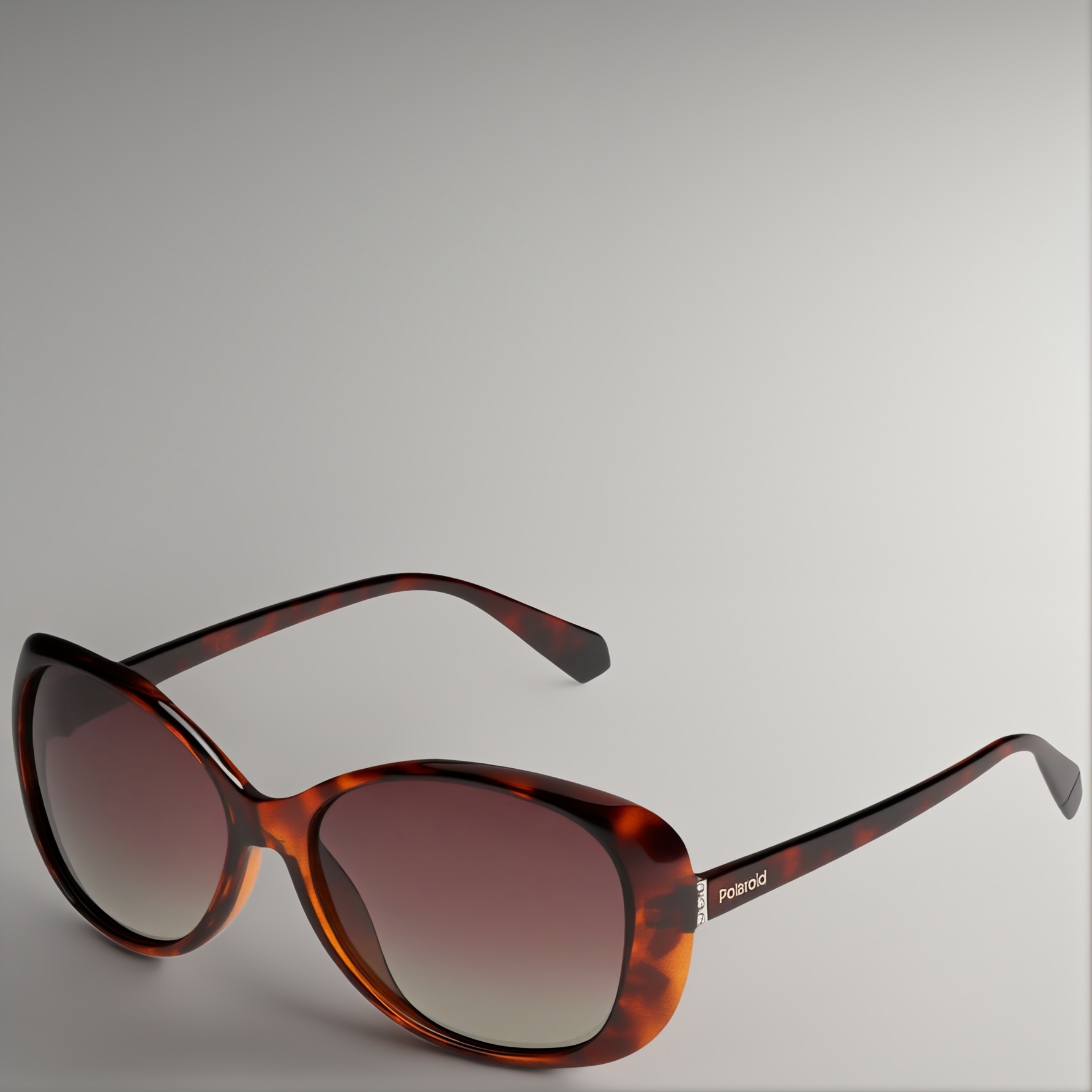}
    \end{tabular}%
}
\caption{Examples of scenes with various light sources (top–middle) rendered using our method, and fused two-light-source scenes (bottom)}
\label{fig:mul_right}
\end{figure}

While our primary focus is on single-light scenarios, the method naturally extends to multiple light sources by exploiting the linearity of radiance ~\cite{debevec2008rendering,ward1994radiance}.
Specifically, we accumulate per-light contributions through additive composition: 
\begin{equation}
x_1 = \sum_{l=1}^{L} x_1^{(l)},
\end{equation}
where $x_1^{(l)}$ denotes the relight result under the l-th light, and $L$ is the total number of light sources.
This straightforward accumulation requires no additional modification and still captures complex phenomena such as overlapping shadows and varying intensities, as shown in Fig.~\ref{fig:mul_right}.

% \section{Qualitative Results on CSG Benchmark}

% \begin{figure}[htbp]
% \begin{center}
% \includegraphics[width=\textwidth]{image/jasper_vis.pdf}
% % \fbox{\rule[-.5cm]{0cm}{4cm} \rule[-.5cm]{4cm}{0cm}}
% \end{center}
% \caption{Visual example of clean-background shadow generation, demonstrating that our light-geometry interaction approach provides accurate control over shadow shapes.}
% \label{fig:jasper_vis}
% \end{figure}

% We present qualitative results on the CSG benchmark in Fig.~\ref{fig:jasper_vis}. The visualizations confirm that our method generates shadows with accurate shapes and realistic density.
\section{Study on Sample-point Count}
\label{sec:samplepoint}
\begin{table*}[htbp]
\centering
\footnotesize
\caption{Study on Sample Point Count}
% \vspace{-0.2cm}
\renewcommand{\arraystretch}{1.1}
\setlength{\tabcolsep}{5pt}
\begin{tabular}{l cccc cc ccc}
\toprule
\textbf{N} & \multicolumn{4}{c}{\textbf{Overall}} & \multicolumn{3}{c}{\textbf{Shadow region}} & \multicolumn{2}{c}{\textbf{Object region}} \\
 \cmidrule(lr){2-5} \cmidrule(lr){6-8} \cmidrule(lr){9-10} 
& RMSE $\downarrow$ & SSIM $\uparrow$ & BER $\downarrow$ & IOU $\uparrow$ & RMSE $\downarrow$ & SSIM  $\uparrow$ & BER $\downarrow$ & RMSE $\downarrow$ & SSIM $\uparrow$  \\
\midrule
8 &  0.0344 & 0.7188 & 0.0612 & 0.7954 & 0.0991 & 0.6125 & 0.1141 & 0.0283 & 0.6875 \\
16 &  0.0334 & 0.7227 & 0.0588 & 0.8096 &0.0898 & 0.6195 & 0.1103 & 0.0282 & 0.6875 \\
32 & 0.0334 & 0.7231 & 0.0586 & 0.8099 & 0.0869 & 0.6198 & 0.1096 &0.0283 & 0.6875 \\
\bottomrule
\end{tabular}
\label{tab:m_study}
% \vspace{-0.1cm}
\end{table*}

We further study the effect of the sample-point count \(N\). Because its computational impact is negligible at our reporting precision, Tab.~\ref{tab:m_study} reports accuracy for \(N\in\{8,16,32\}\). While \(N{=}32\) provides only marginal gains, we adopt \(N{=}16\) by default.

\section{Comparison to LBM with Depth Input}
We additionally compare with LBM equipped with depth input in Tab.~\ref{tab:compare_lbm_depth}. Our method consistently outperforms this depth-augmented LBM across all regions and metrics, showing that the gains of LGI stem from our proposed design rather than merely from using an additional depth modality.

\begin{table*}[htbp]
\centering
\footnotesize
\caption{Comparison to LBM with Depth Input}
\vspace{-0.2cm}
\renewcommand{\arraystretch}{1.1}
\setlength{\tabcolsep}{5pt}
\begin{tabular}{l cccc cc ccc}
\toprule
 & \multicolumn{4}{c}{\textbf{Overall}} & \multicolumn{3}{c}{\textbf{Shadow region}} & \multicolumn{2}{c}{\textbf{Object region}} \\
 \cmidrule(lr){2-5} \cmidrule(lr){6-8} \cmidrule(lr){9-10} 
& RMSE $\downarrow$ & SSIM $\uparrow$ & BER $\downarrow$ & IOU $\uparrow$ & RMSE $\downarrow$ & SSIM  $\uparrow$ & BER $\downarrow$ & RMSE $\downarrow$ & SSIM $\uparrow$  \\
\midrule
LBM+depth  &0.0395 & 0.7122& 0.1003& 0.7284 &0.1208 & 0.5923 & 0.1901 &0.0321 & 0.6692\\
Ours      &0.0334 & 0.7227& 0.0588 & 0.8096 &0.0898 & 0.6195 & 0.1103 &0.0282 & 0.6875 \\
\bottomrule
\end{tabular}
\label{tab:compare_lbm_depth}
\vspace{-0.4cm}
\end{table*}

\section{More Visualization under Different Lights}
\begin{figure}[htbp]
\begin{center}
\includegraphics[width=\textwidth]{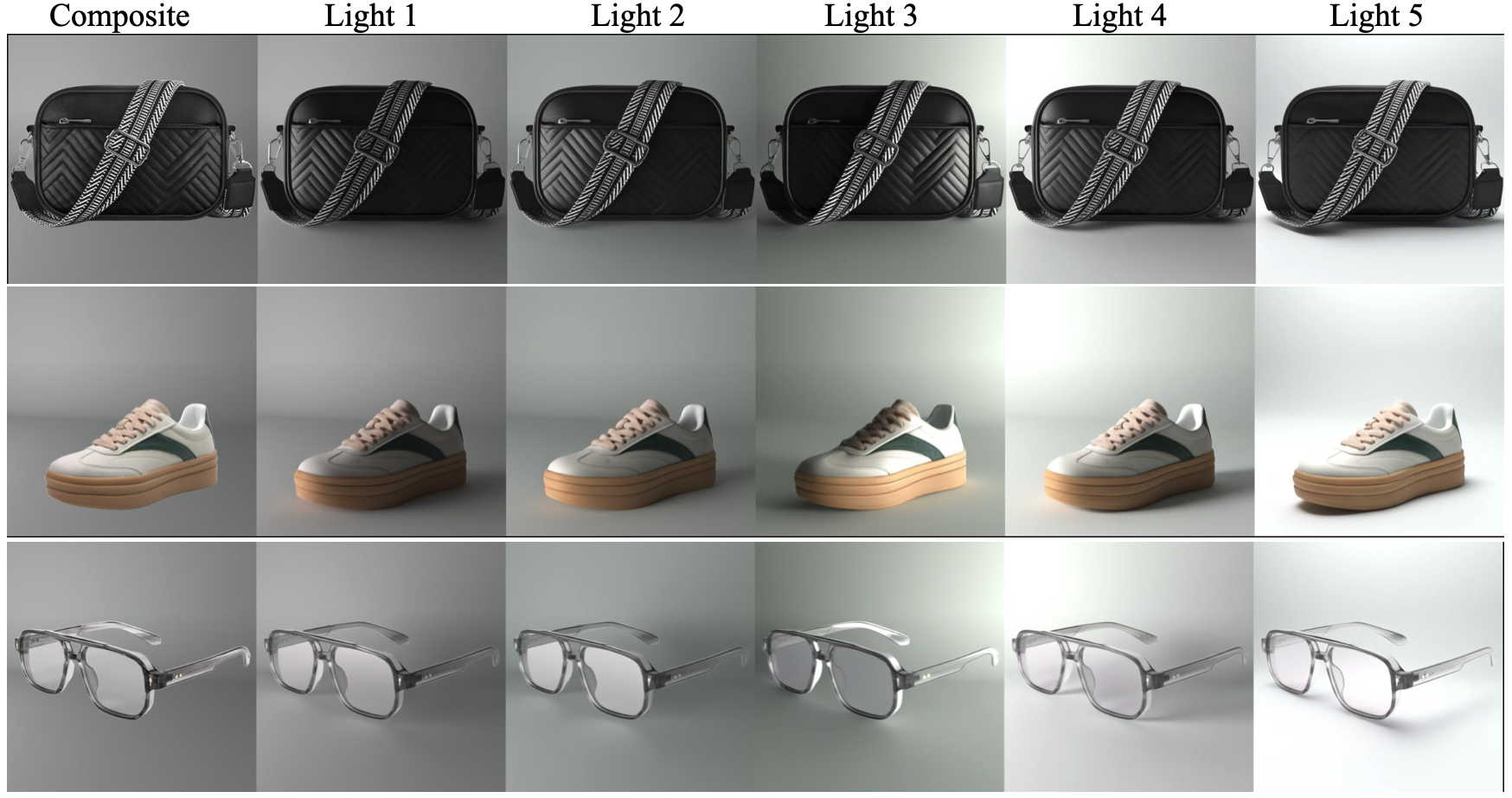}
\end{center}
\caption{Examples produced by our method on real-world object images. }
\label{fig:real_5_lights}
\end{figure}

In this section, we present additional real-world examples of shadow generation and relighting under varying light sources (Fig.~\ref{fig:real_5_lights}). Despite being trained exclusively on synthetic data, our model generalizes effectively to real images and successfully captures indirect lighting effects, such as reflections. More visualizations are included in the supplementary video.

\section{Visual Comparison with and without the LGI Module}
\label{sec:LGI_study}

\begin{figure}[htbp]
\begin{center}
\includegraphics[width=\textwidth]{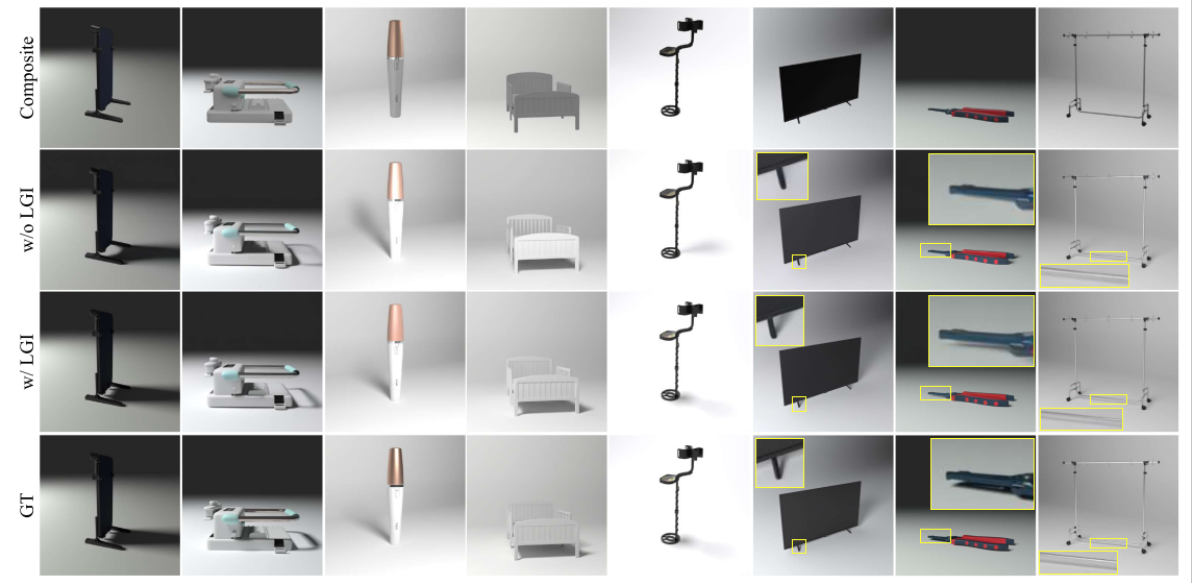}
% \fbox{\rule[-.5cm]{0cm}{4cm} \rule[-.5cm]{4cm}{0cm}}
\end{center}
% \vspace{-0.2cm}
\caption{Qualitative comparison with and without LGI, illustrating that LGI enables geometry-aligned shadows and relighting effects.}
\label{fig:compare_LGI}
% \vspace{-0.4cm}
\end{figure}

Fig.\ref{fig:compare_LGI} compares results with and without the LGI module. Incorporating LGI increases sensitivity to scene geometry, yielding shadows that are more realistic and better aligned with geometric structure.

\section{Complex Background Examples}

\begin{figure}[H] %[htbp]
\begin{center}
\includegraphics[width=0.95\textwidth]{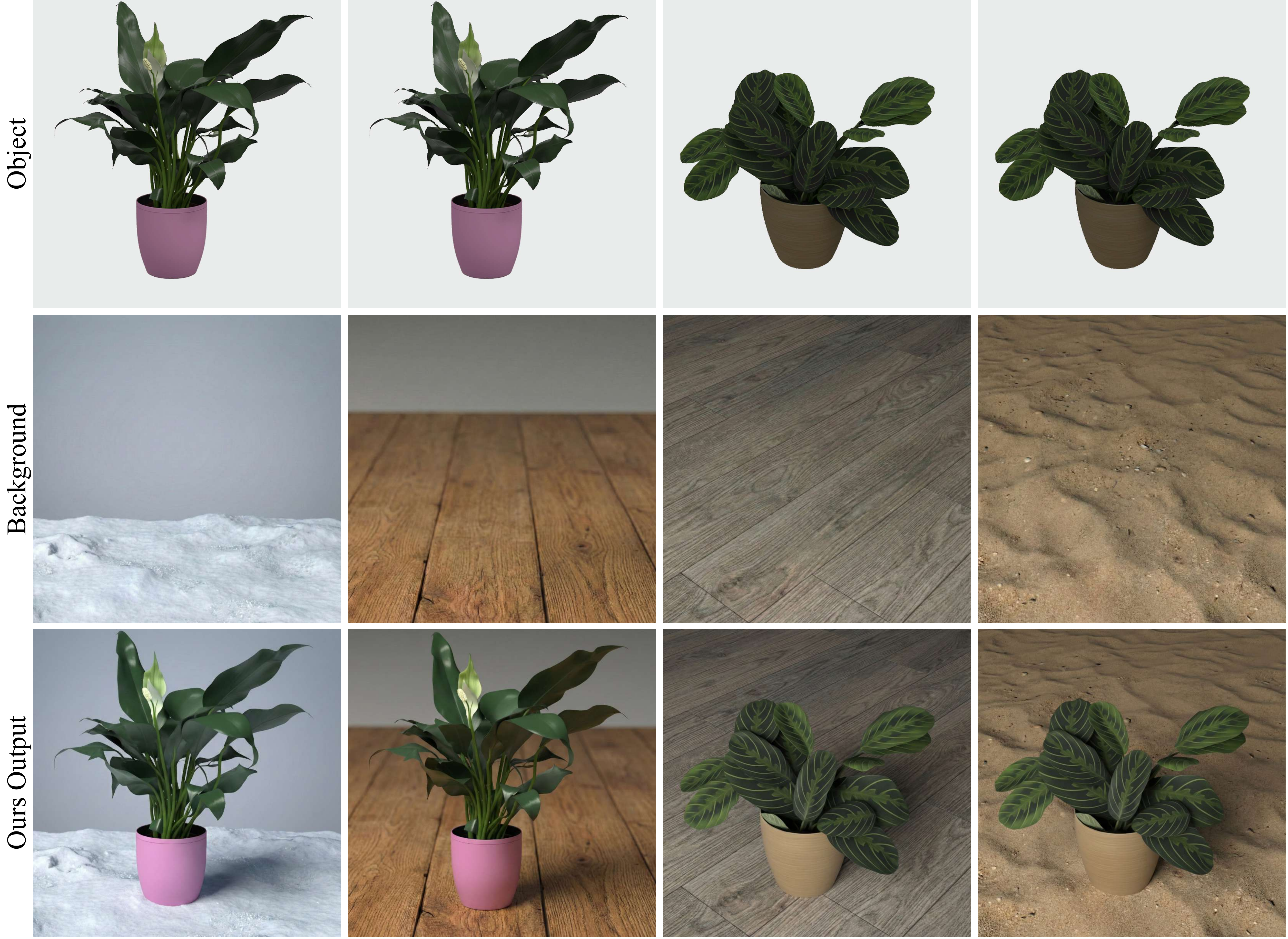}
\end{center}
\vspace{-0.2cm}
\caption{Qualitative results on complex backgrounds. Our method remains robust on both non-planar and textured backgrounds.}
\label{fig:bg_shadrel}
% \vspace{-0.2cm}
\end{figure}

In this section, we shift our focus to complex backgrounds. While the proposed ShadRel dataset is primarily designed to study complex object–background inter-reflections and therefore uses relatively simple backgrounds (e.g., a floor or a floor–wall configuration), the objects themselves span a wide range of materials, geometries, and textures. As a result, complex geometry and texture are still exercised through self-occlusion effects.
Our real-image visualizations in Fig.~\ref{fig:mul_obj} show that the method effectively handles textured backgrounds (e.g., desks, cluttered layouts, and object-rich scenes). Additional qualitative results are provided in Fig.~\ref{fig:bg_shadrel}, where we observe that our method remains robust on non-planar and textured backgrounds.

\begin{figure}[H] %[htbp]
\begin{center}
\includegraphics[width=0.22\textwidth]{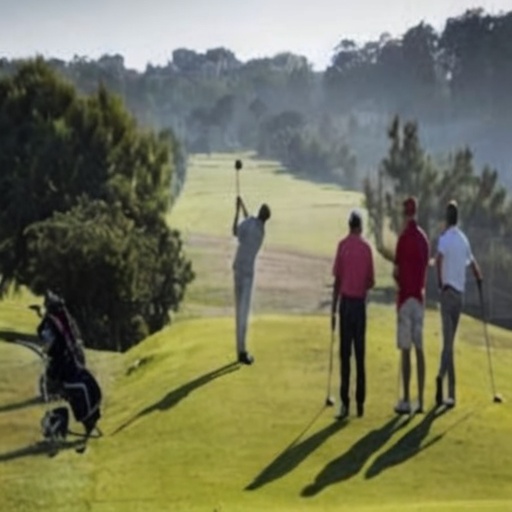}
\includegraphics[width=0.22\textwidth]{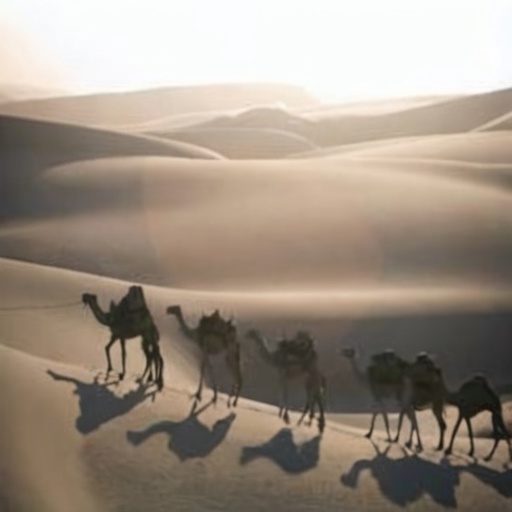}
\includegraphics[width=0.22\textwidth]{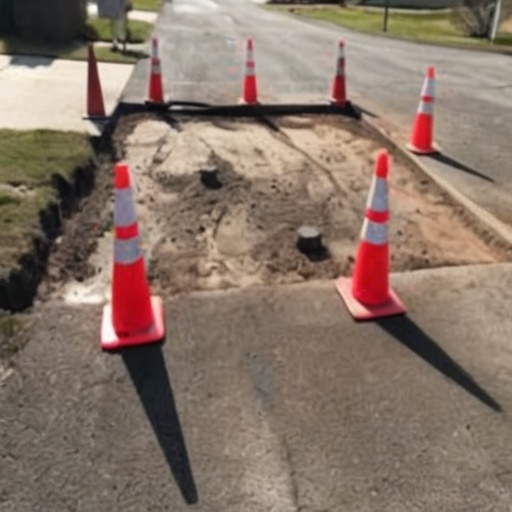}
\includegraphics[width=0.22\textwidth]{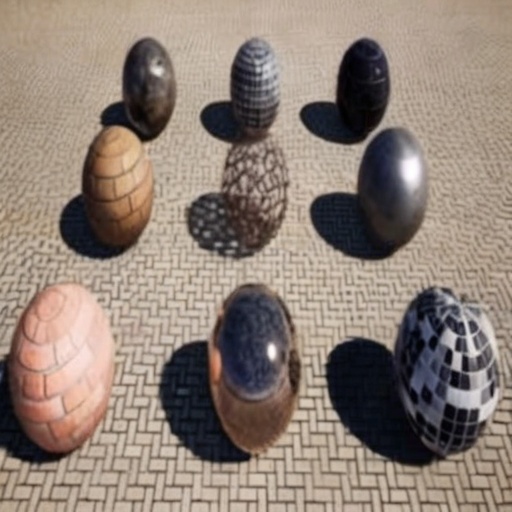}
\includegraphics[width=0.22\textwidth]{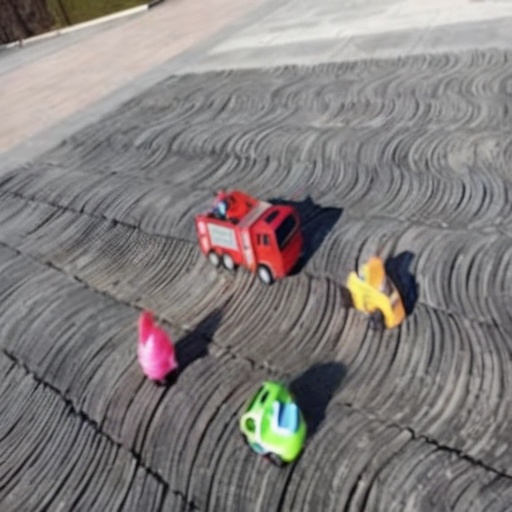}
\includegraphics[width=0.22\textwidth]{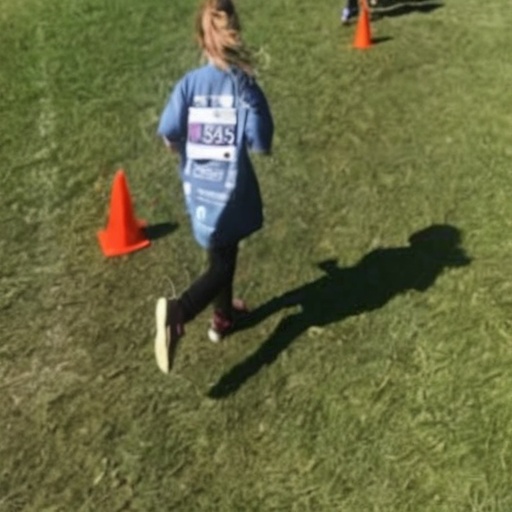}
\includegraphics[width=0.22\textwidth]{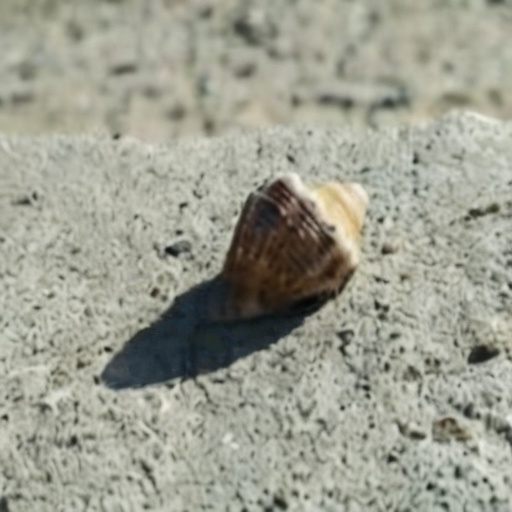}
\includegraphics[width=0.22\textwidth]{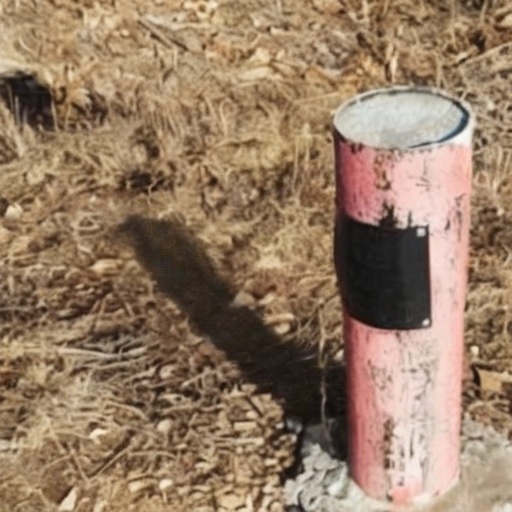}
\end{center}
\vspace{-0.2cm}
\caption{Complex backgrounds in the DESOBAv2 dataset, including non-planar geometry and rich textures and patterns. Our qualitative results on this dataset show that the proposed method produces reasonable predictions even under these challenging background conditions.}
\label{fig:bg_DESOBAv2}
\vspace{-0.2cm}
\end{figure}

\begin{figure}[H] %[htbp]
\begin{center}
\includegraphics[width=\textwidth]{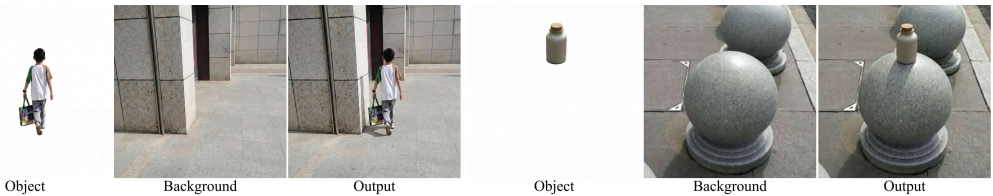}
\end{center}
\caption{Visual results on challenging in-the-wild backgrounds. Our model maintains lighting consistency and shadow realism across diverse surface geometries and textures.}
\label{fig:bg_outdoor_vis}
% \vspace{-0.5cm}
\end{figure}

Furthermore, the DESOBAv2 dataset is a real-world benchmark with images captured on golf courses, tennis courts, beaches, gardens, and other outdoor environments, where the backgrounds are non-planar and contain rich textures and patterns. 
The qualitative results on these challenging cases in Fig.~\ref{fig:bg_DESOBAv2} demonstrate that our method is robust to such complex backgrounds. 
We further demonstrate its generalization capabilities on in-the-wild images in Fig.~\ref{fig:bg_outdoor_vis}, showing that our method generates reasonable, lighting-consistent shadows even on non-planar and textured backgrounds.

\section{Study on Complex Objects}

\begin{figure}[H] %[htbp]
\begin{center}
\includegraphics[width=\textwidth]{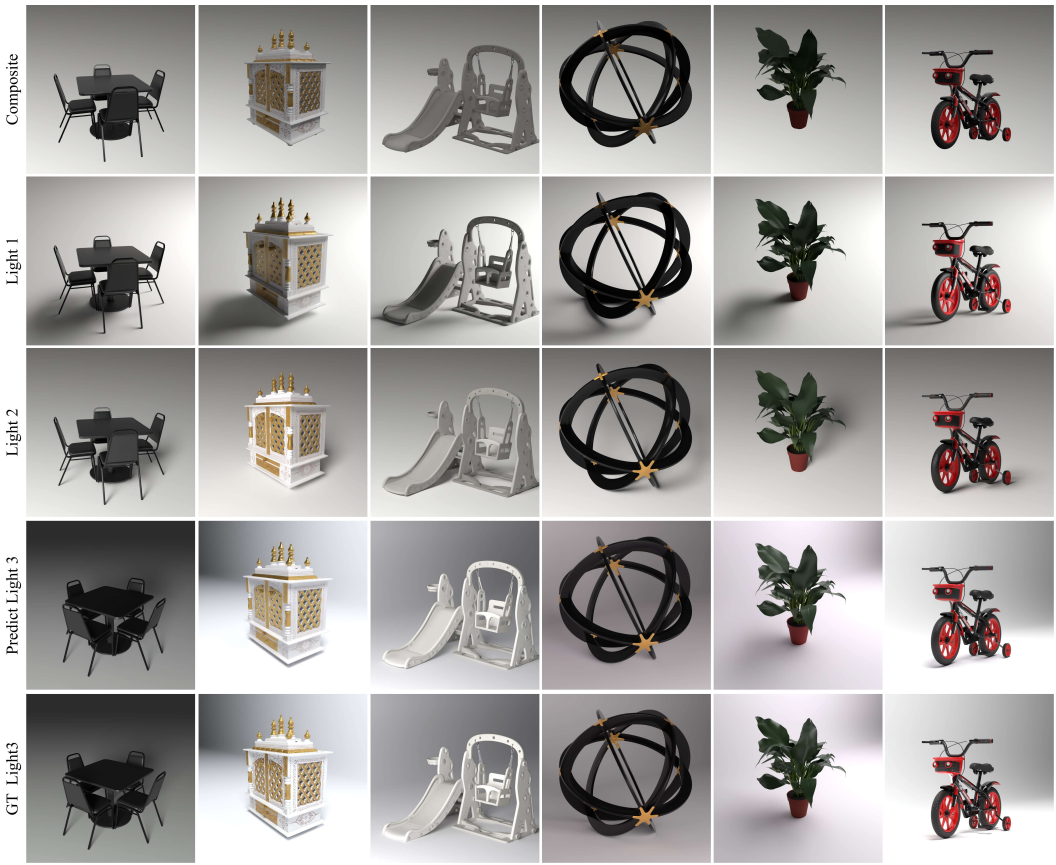}
\end{center}
\caption{Qualitative results on complex objects. Our ShadRel dataset includes complex objects, and our method remains robust on these challenging cases. 
As the dataset provides Ground Truth (GT) for only 5 discrete lighting conditions, the Light 3 shows a direct comparison against the available GT. Light 1 and Light 2 represent unseen illumination directions (sampled from the 24 distinct lights in the supplementary video), demonstrating the model's generalization capabilities where GT is unavailable.}
\label{fig:fg_comp}
% \vspace{-0.5cm}
\end{figure}

To quantitatively evaluate our method on complex objects, we construct a 200-sample subset consisting solely of challenging complex objects and evaluate our method on this subset. The quantitative results in Tab.~\ref{tab:complex_study} show our method outperforms LBM by a large margin, together with the additional qualitative examples in Fig.~\ref{fig:fg_comp}, demonstrate that our approach remains robust beyond simple shapes. We further provide additional qualitative results on this subset and on real images in the supplementary video.
Moreover, the DESOBAv2 dataset also includes complex objects, such as humans with bicycles in Fig.~\ref{fig:compare_SGDGP}; our results on this dataset likewise show that the method preserves high-quality predictions in these cases, further supporting its robustness to complex geometries.

\begin{table*}[htbp]
\centering
\footnotesize
\caption{Study on Complex Objects}
% \vspace{-0.2cm}
\renewcommand{\arraystretch}{1.0}
\setlength{\tabcolsep}{5pt}
\begin{tabular}{l cccc cc ccc}
\toprule
 & \multicolumn{4}{c}{\textbf{Overall}} & \multicolumn{3}{c}{\textbf{Shadow region}} & \multicolumn{2}{c}{\textbf{Object region}} \\
 \cmidrule(lr){2-5} \cmidrule(lr){6-8} \cmidrule(lr){9-10} 
& RMSE $\downarrow$ & SSIM $\uparrow$ & BER $\downarrow$ & IOU $\uparrow$ & RMSE $\downarrow$ & SSIM  $\uparrow$ & BER $\downarrow$ & RMSE $\downarrow$ & SSIM $\uparrow$  \\
\midrule
LBM & 0.0431 & 0.7112 & 0.0954  &0.0691 & 0.1575 & 0.5609 & 0.1758 & 0.0342 & 0.6743 \\
Ours & 0.0341 & 0.7196 & 0.0658 &0.7940 & 0.0929  &0.6114 & 0.1241 & 0.0286 & 0.6828 \\
\bottomrule
\end{tabular}
\label{tab:complex_study}
\vspace{-0.4cm}
\end{table*}

\section{Visualization on Complex Interaction}
\begin{figure}[H] %[htbp]
\begin{center}
\includegraphics[width=\textwidth]{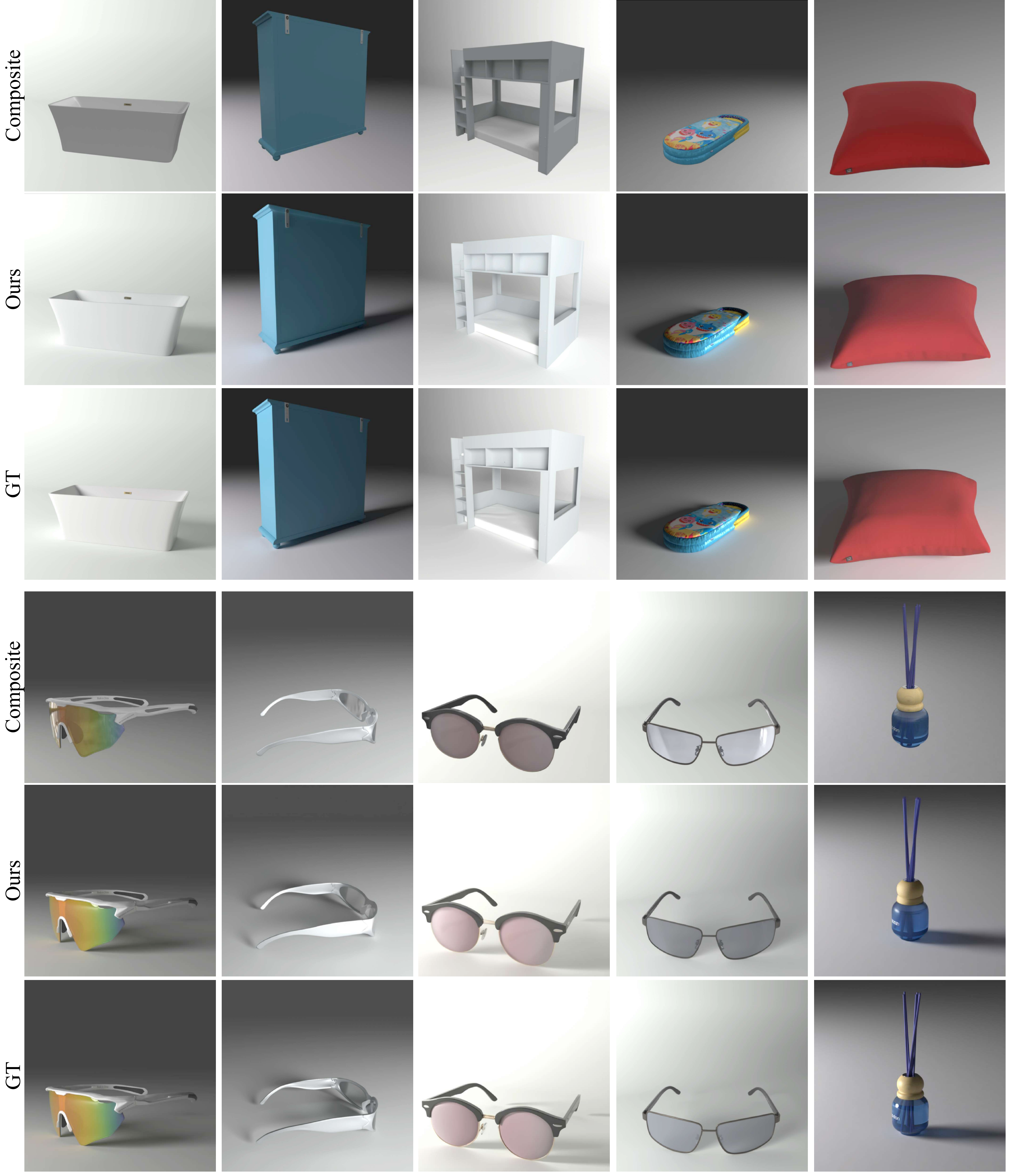}
\end{center}
\vspace{-0.2cm}
\caption{Qualitative results on complex object-background interaction. ShadRel dataset contains objects with diverse materials, including reflective ones such as glass and porcelain, which introduce challenging shadows from transparent objects, with both primary and secondary reflections. Even in these difficult scenarios, our method produces high-fidelity, physically consistent predictions.}
\label{fig:inter}
\vspace{-0.5cm}
\end{figure}

In this section, we present additional qualitative results on complex object–background interaction effects (Fig.~\ref{fig:inter}), including cast shadows from transparent objects and shadows modulated by inter-reflections and secondary reflections. When inserting an object into a scene, not only is the object relit by the background, but the background is also altered by the object (e.g., through shadows and reflections), and these changes in turn further influence the object. Such higher-order object–background interactions remain an open challenge.
Our ShadRel dataset is specifically constructed to capture these challenging phenomena, featuring strong reflections from materials such as glass and weaker reflections from fabrics—scenarios that are close to real applications but rarely covered by existing models and datasets. Even under these demanding conditions, our method produces high-fidelity, physically consistent predictions that align well with the ground truth, highlighting both the unique difficulty of ShadRel and the effectiveness and robustness of our approach.

\section{Dataset Composition and Examples}

\begin{figure}[htbp]
\begin{center}
\includegraphics[width=\textwidth]{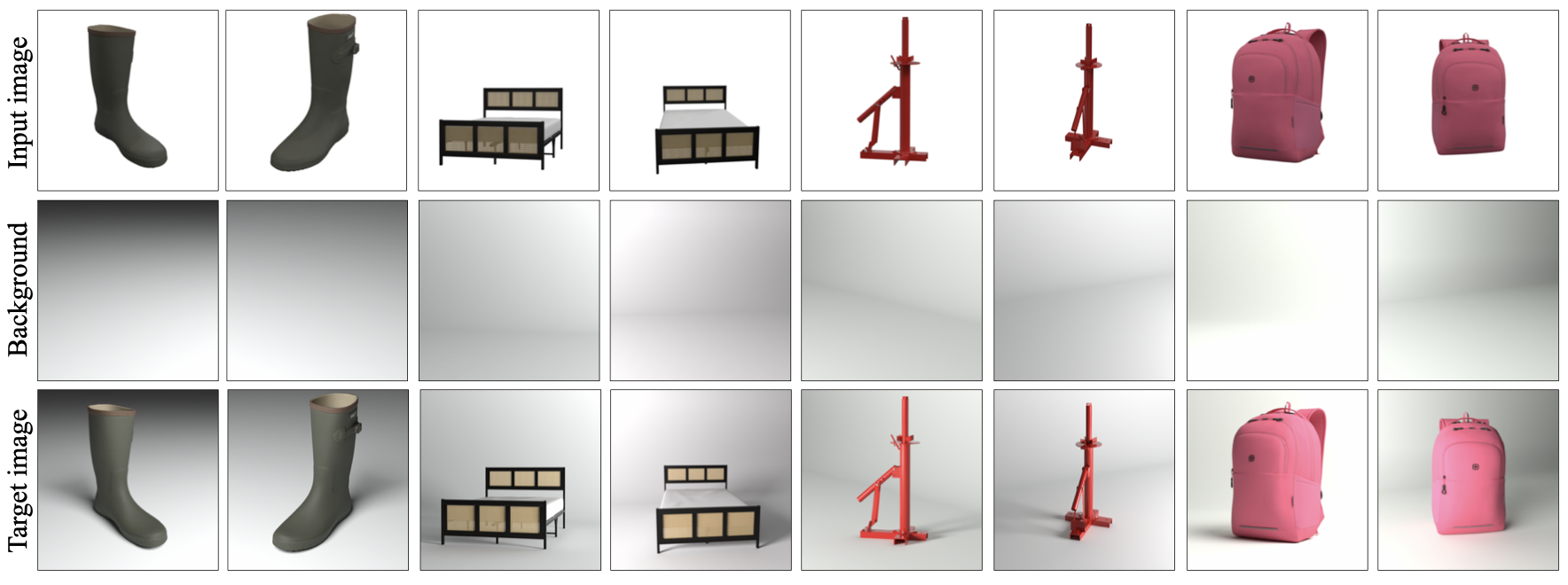}
\end{center}
% \vspace{-0.4cm}
\caption{Examples from our dataset. Each sample includes (top) input images of objects from different camera poses, (middle) background images with floor and optionally a wall under point-light illumination, and (bottom) target images combining the objects and backgrounds under the same point light.}
\label{fig:dataset_img}
\end{figure}

In this section, we present example images from our dataset. Each sample contains 4 random camera poses, and for each pose we generate 5 point-light–background configurations. As shown in Fig.~\ref{fig:dataset_img}, the dataset consists of input images of objects, background images rendered under varying lighting conditions (floor and occasionally a wall), and target images where the same objects and backgrounds are illuminated with different lighting effects.

\section{Limitations}
\begin{figure}[H] %[htbp]
\begin{center}
\includegraphics[width=\textwidth]{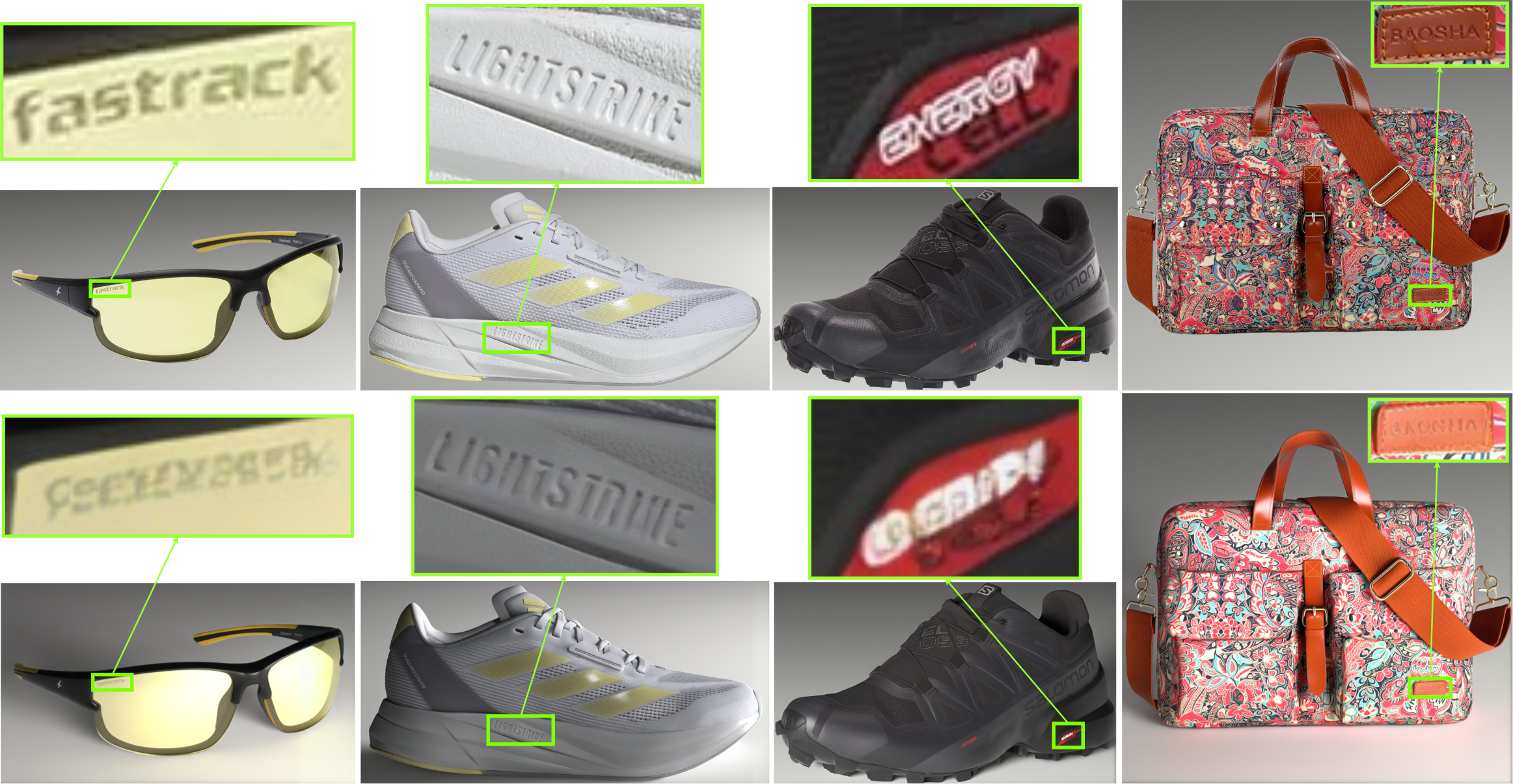}
\end{center}
\caption{Examples illustrating the limitations of our method in reproducing fine-grained details. (Top: Ground Truth; Bottom: Our results.) While our approach can handle certain structural details, tiny elements, such as Logos, may appear inconsistent, distorted, or incomplete.}
\label{fig:limitation}
\end{figure}

Since our approach is initialized from a diffusion model, it inevitably inherits some of the well-known limitations of this class of generative models. Most notably, the generated content may not remain fully faithful to the input specification, particularly in regions requiring high-fidelity reproduction of fine-grained details. As illustrated in Fig.~\ref{fig:limitation}, elements such as logos can appear inconsistent, distorted, or incomplete.

\end{document}